%% file: main.tex
\documentclass[10pt,twocolumn,letterpaper]{article}

\usepackage{cvpr}              
\input{preamble}

\definecolor{cvprblue}{rgb}{0.21,0.49,0.74}
\usepackage[pagebackref,breaklinks,colorlinks,allcolors=cvprblue]{hyperref}
\usepackage{amsmath}
\usepackage[accsupp]{axessibility}  
\usepackage{amssymb}
\usepackage{mathtools}
\usepackage{amsthm}
\usepackage[capitalize,noabbrev]{cleveref}
\input{utils/math_commands}

\input{utils/my_commands}
\input{utils/my_colors}
\makeatletter
\def\blfootnote{\xdef\@thefnmark{}\@footnotetext}
\makeatother
\usepackage[most]{tcolorbox}
\usepackage{caption} 
\def\paperID{} 
\def\confName{CVPR}
\def\confYear{2026}

\title{
Organizing Unstructured Image Collections using Natural Language
}

\author{
Mingxuan Liu$^{1}$~~~
Zhun Zhong$^{4}$\textsuperscript{\dag}~~~
Jun Li$^{6}$~~~
Gianni Franchi$^{3}$~~~
Subhankar Roy$^{5}$~~~
Elisa Ricci$^{1,2}$\\
$^{1}$ University of Trento~~
$^{2}$ Fondazione Bruno Kessler~
$^{3}$ ENSTA Paris, Institut Polytechnique de Paris\\
$^{4}$ Hefei University of Technology~~
$^{5}$ University of Bergamo~~
$^{6}$ Technical University of Munich\\
}

\begin{document}

\twocolumn[{%
    \renewcommand\twocolumn[1][]{#1}%
    \maketitle
    \begin{center}
    \centering
    \vspace{-.8mm}
    \includegraphics[width=\textwidth]{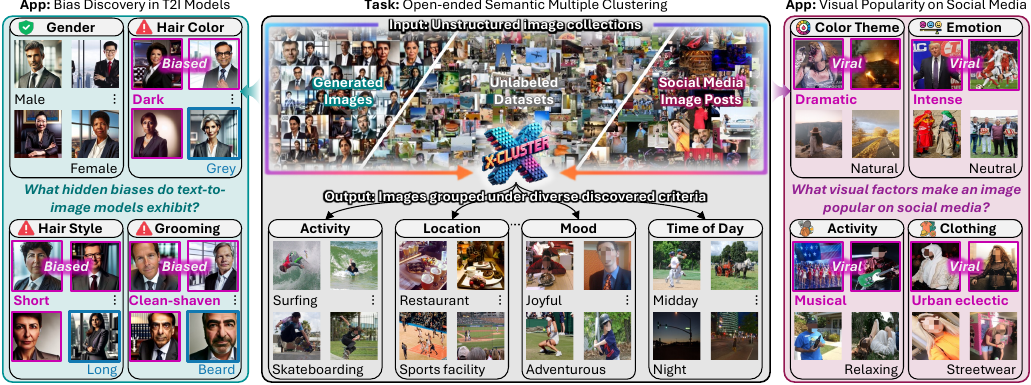}
\captionof{figure}{
\textbf{Organizing unstructured image collections with \methodshortbold.}
{\sethlcolor{midcolor}\hl{Mid:}} We propose \methodshort{}, which takes any unstructured image collection as input, automatically discovers multiple criteria (\eg, \texttt{Activity} or \texttt{Location}) that meaningfully group the data, and outputs images organized into semantic clusters under each criterion, \textit{without} any prior knowledge.
We demonstrate \methodshort{} as a versatile tool for real-world analysis:
\textit{i)} {\sethlcolor{leftcolor}\hl{Left}} - uncovers surprising \textit{novel biases} (\eg, ``Hair color'' or ``Hair style'') when applied to text-to-image (T2I) model outputs, and \textit{ii)} {\sethlcolor{rightcolor}\hl{Right}} - reveals \textit{visual factors} (\eg, ``Dramatic'' color or ``Intensive'' emotion) that drive social media posts virality.
}
    \lblfig{teaser}
    \end{center}
}]

\input{sec/0_abstract}
\footnotetext{Corresponding author.}

\input{sec/1_introduction_v2}

\input{sec/2_related_work}
\input{sec/3_task}

\input{sec/4_method}

\input{sec/5_experiment}

\input{sec/6_application}

\input{sec/7_conclusion}

\clearpage
\section*{Acknowledgments}
This work was supported by the EU Horizon projects ELIAS (No.~101120237) and ELLIOT (No.~101214398). We greatly acknowledge CINECA and the ISCRA initiative for providing high-performance computing resources. We also extend our gratitude to Feng Xue for his valuable suggestions on plot creation. M.L. warmly thanks Margherita Potrich for her unwavering support.

{
    \small
    \bibliographystyle{ieeenat_fullname}
    \bibliography{main}
}

\input{sec/X_appendix}


\end{document}

%% file: preamble.tex
\usepackage{gradient-text}
\usepackage{pifont}
\usepackage{epsfig}
\usepackage{wrapfig}
\usepackage{lipsum}
\usepackage{glossaries}
\usepackage{colortbl}
\usepackage{bbding}
\usepackage{tikz}
\usepackage{comment}
\usepackage{color}
\usepackage{xcolor}
\usepackage{xspace}
\usepackage{nicefrac}
\usepackage{bbm}
\usepackage{bm}
\usepackage{tabularx,verbatim}
\usepackage{multirow}
\usepackage{caption}
\usepackage{xfrac}
\usepackage[accsupp]{axessibility}
\usepackage{changepage}  
\usepackage{url}            
\usepackage{amsfonts}       
\usepackage{subcaption}
\usepackage{dblfloatfix}
\usepackage{textcomp}
\usepackage{setspace}
\usepackage{listings}
\usepackage{mwe} 
\usepackage{makecell}
\usepackage{soul}
\usepackage[scaled=0.85]{DejaVuSansMono}
\usepackage{epigraph}
\usepackage{adjustbox}
\usepackage{titletoc}
\usepackage{dirtytalk}
\usepackage{makecell}
\usepackage[most]{tcolorbox} 

\newtcbox{\navitag}{enhanced,nobeforeafter,tcbox raise base,boxrule=0.4pt,top=0mm,bottom=0mm,
  right=0mm,left=4mm,arc=2pt,boxsep=2pt,before upper={\vphantom{dlg}},
colframe=colabcolor!50!black,coltext=colabcolor!25!black,colback=colabcolor!10!white,
  overlay={\begin{tcbclipinterior}\fill[colabcolor!80] (frame.south west)
    rectangle node[text=white,font=\sffamily\bfseries\tiny,rotate=90] {NAV} ([xshift=4mm]frame.north west);\end{tcbclipinterior}}}

\definecolor{mapcolorRGB}{RGB}{101,114,156}
\colorlet{mapcolor}{mapcolorRGB}

\definecolor{dakrpurpletag}{RGB}{108,72,197}
\definecolor{lightpurpletag}{RGB}{205,193,255}
\newtcbox{\captag}{
  enhanced,
  nobeforeafter,
  tcbox raise base,
  boxrule=0.5pt,
  top=0mm,
  bottom=0mm,
  right=0mm,
  left=0mm,
  arc=2pt,
  boxsep=1.5pt,
  before upper={\vphantom{dlg}},
  colframe=mapcolor!50!black,
  coltext=mapcolor!25!black,
  colback=lightpurpletag
}

\definecolor{lightgraytag}{RGB}{242,242,242}
\newtcbox{\alttag}{
  enhanced,
  nobeforeafter,
  tcbox raise base,
  boxrule=0.5pt,
  top=0mm,
  bottom=0mm,
  right=0mm,
  left=0mm,
  arc=2pt,
  boxsep=1.5pt,
  before upper={\vphantom{dlg}},
  colframe=mapcolor!50!black,
  coltext=mapcolor!25!black,
  colback=lightgraytag 
}
\definecolor{lightgreentag}{RGB}{237,255,213}
\newtcbox{\besttag}{
  enhanced,
  nobeforeafter,
  tcbox raise base,
  boxrule=0.5pt,
  top=0mm,
  bottom=0mm,
  right=0mm,
  left=0mm,
  arc=2pt,
  boxsep=.5pt,
  before upper={\vphantom{dlg}},
  colframe=mapcolor!50!black,
  coltext=mapcolor!25!black,
  colback=lightgreentag 
}

\definecolor{midcolor}{RGB}{229,229,229}
\definecolor{leftcolor}{RGB}{216,235,236}
\definecolor{rightcolor}{RGB}{239,220,229}

%% file: utils/math_commands.tex
\usepackage{amsmath,amsfonts,bm}

\def\eqref#1{equation~\ref{#1}}

\def\1{\bm{1}}

\DeclareMathAlphabet{\mathsfit}{\encodingdefault}{\sfdefault}{m}{sl}
\SetMathAlphabet{\mathsfit}{bold}{\encodingdefault}{\sfdefault}{bx}{n}

%% file: utils/my_commands.tex
\newcommand{\lblsec}[1]{\label{sec:#1}}
\newcommand{\lblfig}[1]{\label{fig:#1}}
\newcommand{\lbltab}[1]{\label{tbl:#1}}

\newcommand{\lbleq}[1]{\label{eq:#1}}

\newcommand{\refsec}[1]{\cref{sec:#1}}
\newcommand{\reffig}[1]{\cref{fig:#1}}
\newcommand{\reftab}[1]{\cref{tbl:#1}}

\newcommand{\refapp}[1]{Supp.~\ref*{sec:#1}}

\newcommand{\myparagraph}[1]{\vspace{0.06cm}\noindent\textbf{#1}}
\newcommand{\lesspace}{\vspace{-0.4cm}}

\newcommand{\action}{Action-3c\xspace}
\newcommand{\fruit}{Fruit-2c\xspace}
\newcommand{\card}{Card-2c\xspace}
\newcommand{\clevr}{Clevr-4c\xspace}
\newcommand{\ourcoco}{COCO-4c\xspace}
\newcommand{\ourfood}{Food-4c\xspace}

\newcommand{\blip}{BLIP\xspace}
\newcommand{\llava}{LLaVA-NeXT\xspace}
\newcommand{\multillava}{LLaVA-NeXT-Interleave\xspace}
\newcommand{\llamathree}{Llama-3\xspace}
\newcommand{\llamathreeone}{Llama-3.1\xspace}

\newcommand{\gptfour}{GPT-4-turbo\xspace}
\newcommand{\gptfouromni}{GPT-4o\xspace}
\newcommand{\gptfourv}{GPT-4V\xspace}

\newcommand{\ictc}{IC$\mid$TC\xspace}
\newcommand{\ssdllm}{SSD-LLM\xspace}
\newcommand{\mmap}{MMaP\xspace}
\newcommand{\msub}{MSub\xspace}

\newcommand{\cacc}{CAcc\xspace}
\newcommand{\cacctitle}{Clustering Accuracy\xspace}
\newcommand{\cacclong}{clustering accuracy\xspace}

\newcommand{\sacc}{SAcc\xspace}
\newcommand{\sacctitle}{Semantic Accuracy\xspace}
\newcommand{\sacclong}{semantic accuracy\xspace}

\newcommand{\tpr}{TPR\xspace}
\newcommand{\tprtitle}{True Positive Rate\xspace}

\newcommand{\hmean}{HM\xspace}
\newcommand{\hmeantitle}{Harmonic Mean\xspace}

\newcommand{\llm}{LLM\xspace}
\newcommand{\llmtitle}{Large Language Model\xspace}
\newcommand{\llmlong}{large language model\xspace}

\newcommand{\mllm}{MLLM\xspace}
\newcommand{\mllmtitle}{Multimodal Large Language Model\xspace}
\newcommand{\mllmlong}{multimodal large language model\xspace}

\newcommand{\vqa}{VQA\xspace}

\newcommand{\vqalong}{visual question answering\xspace}

\newcommand{\dalle}{DALL·E3\xspace}

\newcommand{\taskshort}{OpenSMC\xspace}
\newcommand{\tasktitle}{Open-ended Semantic Multiple Clustering\xspace}
\newcommand{\tasklong}{open-ended semantic multiple clustering\xspace}

\newcommand{\tcmcshort}{TCMC\xspace}

\newcommand{\tcmctitle}{Text Criterion conditioned Multiple Clustering\xspace}

\newcommand{\methodshort}{{$\mathcal{X}$-Cluster}\xspace}
\newcommand{\methodshortbold}{{\textbf{$\bm{\mathcal{X}}$-Cluster}}\xspace}

\newcommand{\methodfirsttitle}{{\textbf{$\bm{\mathcal{X}}$-Cluster}: e\textbf{X}ploratory \textbf{Cluster}ing}\xspace}

\newcommand{\symbsys}{\mathcal{H}}

\newcommand{\data}{\mathcal{D}}

\newcommand{\gtrules}{\mathcal{Y}}
\newcommand{\manyrules}{\mathcal{R}}
\newcommand{\rulespool}{\widetilde{\manyrules}}

\newcommand{\manynames}{\mathcal{S}}
\newcommand{\onename}{s}

\newcommand{\onerule}{R}
\newcommand{\substructure}{\mathcal{O}}
\newcommand{\cluster}{\mathcal{C}}

\newcommand{\bx}{\mathbf{x}}

\crefname{section}{\S}{\S\S}
\crefname{subsection}{\S}{\S\S}
\crefformat{table}{Tab.~#2#1#3}
\crefformat{figure}{Fig.~#2#1#3}
\crefformat{equation}{Eq.~#2#1#3}
\newcommand{\rowNumber}[1]{}

\newcommand{\tablestyle}[2]{\setlength{\tabcolsep}{#1}\renewcommand{\arraystretch}{#2}\centering\footnotesize}

%% file: utils/my_colors.tex
\definecolor{codegreen}{rgb}{0,0.6,0}
\definecolor{codegray}{rgb}{0.5,0.5,0.5}
\definecolor{codepink}{RGB}{252, 142, 172}
\definecolor{codepurple}{rgb}{0.58,0,0.82}
\definecolor{backcolour}{RGB}{245,245,245}
\definecolor{benchmarkgreen}{RGB}{0.00000  0.69020  0.31373}

\definecolor{viralpink}{rgb}{0.83137 , 0.06667,  0.34902}
\definecolor{majorblue}{rgb}{0.20784 , 0.40392 , 1.00000}
\definecolor{benchmarkgreen}{rgb}{0.00000  0.69020  0.31373}

\definecolor{criteriagrey}{rgb}{0.56863, 0.56863, 0.56863}
\definecolor{royalblue(web)}{rgb}{0.25, 0.41, 0.88}
\definecolor{blue-violet}{rgb}{0.54, 0.17, 0.89}
\definecolor{brightmaroon}{rgb}{0.76, 0.13, 0.28}
\definecolor{darkmagenta}{rgb}{0.55, 0.0, 0.55}
\definecolor{bleudefrance}{rgb}{0.19, 0.55, 0.91}
\definecolor{palatinateblue}{rgb}{0.15, 0.23, 0.89}
\definecolor{royalblue(web)}{rgb}{0.25, 0.41, 0.88}
\definecolor{whitesmoke}{rgb}{0.96, 0.96, 0.96}
\definecolor{thulianpink}{rgb}{0.87, 0.44, 0.63}
\definecolor{amber(sae/ece)}{rgb}{1.0, 0.49, 0.0}
\definecolor{darkblue}{rgb}{0.0, 0.0, 0.55}
\definecolor{alizarin}{rgb}{0.82, 0.1, 0.26}
\definecolor{asparagus}{rgb}{0.53, 0.66, 0.42}
\definecolor{darkspringgreen}{rgb}{0.09, 0.45, 0.27}
\definecolor{columbiablue}{rgb}{0.61, 0.87, 1.0}
\definecolor{wildblueyonder}{rgb}{0.64, 0.68, 0.82}
\definecolor{trolleygrey}{rgb}{0.5, 0.5, 0.5}
\definecolor{paleaqua}{rgb}{0.74, 0.83, 0.9}
\definecolor{bubblegum}{rgb}{0.99, 0.76, 0.8}
\definecolor{coralred}{rgb}{1.0, 0.25, 0.25}
\definecolor{green(ryb)}{rgb}{0.4, 0.69, 0.2}
\definecolor{flame}{rgb}{0.89, 0.35, 0.13}
\definecolor{bittersweet}{rgb}{1.0, 0.44, 0.37}
\definecolor{cherrypink}{rgb}{0.86,0.29,0.29}
\definecolor{forestgreen}{rgb}{0.13,0.55,0.13}
\definecolor{champagne}{rgb}{0.74, 0.83, 0.9}
\definecolor{champagne}{rgb}{0.97, 0.91, 0.81}
\definecolor{codegreen}{rgb}{0,0.6,0}
\definecolor{codegray}{rgb}{0.5,0.5,0.5}
\definecolor{codepurple}{rgb}{0.58,0,0.82}
\definecolor{backcolour}{rgb}{0.96,0.96,0.94}
\definecolor{verygrey}{rgb}{0.9, 0.9, 0.9}

\definecolor{darksalmon}{rgb}{0.91, 0.59, 0.48}
\definecolor{emerald}{rgb}{0.31, 0.78, 0.47}
\definecolor{green(pigment)}{rgb}{0.0, 0.65, 0.31}
\definecolor{amaranth}{rgb}{0.9, 0.17, 0.31}
\definecolor{iris}{rgb}{0.35, 0.31, 0.81}
\definecolor{uu}{rgb}{0.95, 0.51, 0.51}
\definecolor{spirodiscoball}{rgb}{0.06, 0.75, 0.99}

\definecolor{highlightRowColor}{rgb}{0.95, 0.95, 1}
\definecolor{firstBest}{rgb}{0.9, 1, 0.9}
\definecolor{newFirstBest}{rgb}{0.76078, 94902, 0.78431}
\definecolor{secondBest}{rgb}{1, 0.95, 0.95}

\definecolor{imgGreen}{rgb}{0.48235,  0.66275,  0.48627}
\definecolor{tagBlue}{rgb}{0.31765,  0.51765,  0.69804}
\definecolor{capPink}{rgb}{0.83529,  0.32157,  0.46275}
\definecolor{annoGrey}{rgb}{0.55098,  0.55098,  0.55098}

\definecolor{trendingPink}{RGB}{212,17,89}
\definecolor{trendingBlue}{RGB}{53,103,255}

\definecolor{higher}{RGB}{16,119,51}
\definecolor{lower}{RGB}{220,50,32}

\definecolor{correctpred}{RGB}{0,153,77}
\definecolor{makesensepred}{RGB}{209, 70, 153}
\definecolor{wrongpred}{RGB}{250, 0, 0}
\definecolor{superpred}{RGB}{0, 127, 255}
\definecolor{warningred}{RGB}{1,0,0}

\definecolor{rebuttal}{RGB}{220,50,32}



%% file: sec/0_abstract.tex
\begin{abstract}
In this work, we introduce and study the novel task of \tasktitle{} (\taskshort). Given a large, unstructured image collection, the goal is to automatically discover several, diverse \textit{semantic} clustering criteria (\eg, \texttt{Activity} or \texttt{Location}) from the images, and subsequently organize them according to the discovered criteria, without requiring any human input. Our framework, \methodfirsttitle{}, treats text as a reasoning proxy: it concurrently scans the entire image collection, proposes candidate criteria in natural language, and groups images into meaningful clusters per criterion. This radically differs from previous works, which either assume predefined clustering criteria or fixed cluster counts. To evaluate \methodshort, we create two new benchmarks, COCO-4C and Food-4C, each annotated with four distinct grouping criteria and corresponding cluster labels. Experiments show that \methodshort{} can effectively reveal meaningful partitions on several datasets. Finally, we use \methodshort{} to achieve various real-world applications, including uncovering hidden biases in text-to-image (T2I) generative models and analyzing image virality on social media.
Project page: {\color{trendingPink}https://oatmealliu.github.io/xcluster.html}
\end{abstract}

%% file: sec/1_introduction_v2.tex
\vspace{-3.0mm}
\section{Introduction}
\lblsec{introduction}

When organizing a large collection of unlabelled images, a natural question arises: \textit{how should we group them?} One could imagine many possible criteria, \ie, based on \texttt{Activity}, \texttt{Location}, or even \texttt{Color}. Yet, it is often unclear which criterion, if any, best describes the dataset, or whether multiple valid grouping principles coexist.
As a result, the seemingly simple task of \textit{clustering images} becomes challenging, 
as it is 
influenced by both the visual appearance of the data and their underlying semantics.
However, tackling this open-ended unsupervised task of \textit{automatically uncovering diverse and interpretable substructures within large image collections} is pivotal for many applications, such as social media recommendation~\citep{cheng2024retrieval} and dataset auditing~\citep{bianchi2023easily}.

Existing clustering approaches still heavily rely on an iterative, human-in-the-loop interpretation and refinement process. Typically, we begin by setting a few hyperparameters (\eg, the number of grouping criteria or clusters) for Deep Clustering (DC) methods~\citep{caron2018deep,van2020scan} to get a single partition, or for Multiple Clustering (MC) methods~\citep{yu2024multiple} to produce several partitions showing different views of the data. We then inspect sample images from each cluster, hoping they correspond to meaningful categories (\eg, ``Surfing'' or ``Skateboarding'') and, ideally, that all clusters follow a coherent criterion (\eg, \texttt{Activity}). When such patterns fail to emerge, we tweak the hyperparameters and try again until the clusters finally make sense to us. This labor-intensive trial-and-error loop exists because \textit{i)} the resulting clusters are not directly interpretable, being represented only as index assignments.
\textit{ii)} both DC and MC methods converge to solutions shaped by model inductive biases and hyperparameter settings, rather than the data’s intrinsic semantics.

To enhance controllability and interpretability, recent studies have introduced Text-Conditioned Multiple Clustering (\tcmcshort)~\citep{yao2024multi,yao2025customized}. 
TCMC approaches employ Multimodal Large Language Models (\mllm{s})~\citep{openai2024gpt4o,liu2024llava} to generate semantic clusters based on user-defined criteria and assign images accordingly, producing human-understandable cluster labels. However, these approaches assume that users already know meaningful ways to organize the dataset. As datasets grow in size and complexity, defining such criteria becomes increasingly unrealistic.
Moreover, by relying on static preset criteria, this paradigm may overlook previously unknown grouping dimensions that organically emerge from ever-evolving data.

In this paper, we introduce the task of \textit{\tasktitle{}} (\taskshort{}),  the \textit{first} task that aims to automatically generate \textit{open-ended} and interpretable groupings of large, unstructured image collections \textit{without} any human priors.
Specifically, the goal of \taskshort is to \textit{discover} clustering criteria directly from the data and uncover their corresponding semantic clusters to organize images accordingly. This task is particularly challenging because \textit{i)} it requires \textit{concurrent} reasoning over all images to identify valid clustering criteria, and \textit{ii)} it assumes \textit{no} access to user knowledge about either the clustering criteria or the number of clusters. \reftab{diff_table} summarizes the key differences between \taskshort and other paradigms.

\input{tables/diff_settings}

To address \taskshort{}, we make the following contributions. First, we introduce \methodfirsttitle, a novel training-free two-stage framework powered by \mllm{s}~\citep{liu2024llava,li2024llava,li2023blip} and \llm{s}~\citep{meta2024llama31}. \methodshort consists of two consecutive modules: the {Criteria Proposer} and the {Semantic Grouper}. The Criteria Proposer employs a \llm{} to holistically reason over the entire image collection through textual representations to discover potential clustering criteria. For each discovered criterion, the Semantic Grouper then organizes images into distinct semantic substructures based on their criterion-related visual content.
As shown in \reffig{teaser}(mid), our \methodshort automatically discovers clustering criteria (\eg, \texttt{Activity}, \texttt{Location}) and uncovers their corresponding semantic clusters (\eg, ``Surfing'', ``Skateboarding'' under \texttt{Activity}), all expressed in human-interpretable natural language.

As our second contribution, we introduce two realistic and large-scale benchmarks, COCO-4c and Food-4c, each annotated with ground-truth data for up to \textit{four} clustering criteria. Using them, we comprehensively evaluate the effectiveness of our method in both criteria discovery and semantic grouping.
As our third contribution, we demonstrate the versatility of \methodshort by applying it across diverse applications.
When applied to occupation portrait images generated by text-to-image (T2I) generative models (\reffig{teaser}{(left)}), it uncovers novel occupational biases, such as \dalle{}~\citep{betker2023improving} associating CEOs with ``dark'' and ``short'' hair, beyond well-known biases (\eg, \texttt{Gender}).
When applied to social media image posts, \methodshort finds that images featuring ``dramatic'' colors, ``intense'' emotions, or `urban eclectic'' clothing styles tend to attract greater popularity online.
These findings show that \methodshort is a practical tool for understanding large-scale unstructured visual data, enabling the discovery of novel, unexpected patterns.

%% file: tables/diff_settings.tex
\begin{table}[!t]
\centering
\caption{
    \textbf{Comparison of different clustering paradigms.}  Unlike DC, MC, and \tcmcshort settings, the proposed \taskshort task does not assume any prior knowledge and offers interpretable results.
}
\lbltab{diff_table}
\footnotesize 

\begin{tabular}{llcccc}
\toprule

&& DC & MC & TCMC & \taskshort
\\
\toprule

\multirow{3}{*}{\rotatebox[origin=c]{90}{Prior}}
& Knowledge \# Criteria 
& \multicolumn{1}{c}{\textcolor{darksalmon}{\XSolidBrush}}
& \multicolumn{1}{c}{\textcolor{green(pigment)}{\Checkmark}}
& \multicolumn{1}{c}{\textcolor{darksalmon}{\XSolidBrush}}
& \multicolumn{1}{c}{\textcolor{darksalmon}{\XSolidBrush}}
\\

& Text Criteria 
& \multicolumn{1}{c}{\textcolor{darksalmon}{\XSolidBrush}}
& \multicolumn{1}{c}{\textcolor{darksalmon}{\XSolidBrush}}
& \multicolumn{1}{c}{\textcolor{green(pigment)}{\Checkmark}} 
& \multicolumn{1}{c}{\textcolor{darksalmon}{\XSolidBrush}}
\\

& Knowledge \# Clusters 
& \multicolumn{1}{c}{\textcolor{green(pigment)}{\Checkmark}}
& \multicolumn{1}{c}{\textcolor{green(pigment)}{\Checkmark}} 
& \multicolumn{1}{c}{\textcolor{green(pigment)}{\Checkmark}} 
& \multicolumn{1}{c}{\textcolor{darksalmon}{\XSolidBrush}}
\\

\midrule

\multirow{3}{*}{\rotatebox[origin=c]{90}{Output}}
& Multiple Clustering
& \multicolumn{1}{c}{\textcolor{darksalmon}{\XSolidBrush}}
& \multicolumn{1}{c}{\textcolor{green(pigment)}{\Checkmark}} 
& \multicolumn{1}{c}{\textcolor{green(pigment)}{\Checkmark}} 
& \multicolumn{1}{c}{\textcolor{green(pigment)}{\Checkmark}} 
\\

& Interpretable 
& \multicolumn{1}{c}{\textcolor{darksalmon}{\XSolidBrush}}
& \multicolumn{1}{c}{\textcolor{darksalmon}{\XSolidBrush}}
& \multicolumn{1}{c}{\textcolor{green(pigment)}{\Checkmark}} 
& \multicolumn{1}{c}{\textcolor{green(pigment)}{\Checkmark}}
\\

& Open-ended 
& \multicolumn{1}{c}{\textcolor{darksalmon}{\XSolidBrush}}
& \multicolumn{1}{c}{\textcolor{darksalmon}{\XSolidBrush}}
& \multicolumn{1}{c}{\textcolor{darksalmon}{\XSolidBrush}}
& \multicolumn{1}{c}{\textcolor{green(pigment)}{\Checkmark}}
\\

\bottomrule
\vspace{-9.0mm}
\end{tabular}
\end{table}

%% file: sec/2_related_work.tex
\section{Related Work}
\lblsec{relatedwork}
\myparagraph{Image Clustering.}
Deep clustering learns visual features and produces a \emph{single} partition of an unlabeled dataset via self‑supervision~\citep{caron2018deep,caron2020unsupervised,ren2024deep}.  
Multiple clustering extends this idea, seeking \emph{multiple} non‑redundant partitions with data augmentations, diversity losses, or subspace methods~\citep{qi2009principled,hu2018subspace,yao2023augdmc,metaxas2023divclust,yu2024multiple}.
Despite steady progress, both paradigms share key limitations: \textit{i)} their results are shaped by model inductive biases and training algorithms, limiting generalization beyond object-centric data and often misaligning with user intent or data semantics; and \textit{ii)} clusters are produced as numeric indices rather than human-readable names.
In contrast, \methodshort derives both meaningful criteria and cluster names directly from unlabeled data.

\myparagraph{\tcmctitle.}
\tcmcshort lets users steer clustering by specifying the grouping criteria.  
Learning‑based approaches such as \mmap~\citep{yao2024multi} and \msub~\cite{yao2025customized} first use GPT-4~\citep{openai2023gpt4} to generate reference words (\eg, fruits colors like ``Red'' or ``Green'') conditioned on the user-provided criterion (\eg, \textit{Color}-based fruits clustering). They then optimize learnable image embeddings by aligning with these criterion conditioned reference words.
Training‑free methods instead translate images into text. 
\ictc~\citep{kwon2023image} first captions each image with LLaVA~\citep{liu2024visual} conditioned on the user’s criterion, then uses GPT‑4 to refine the captions and assign cluster names for a user‑specified number of clusters. \ssdllm~\citep{luo2024llm} strengthens \ictc{} by augmenting the prompt with the dataset’s primary object labels.
Similar ideas have also been explored in applications such as visual trend discovery~\citep{deng2025visual}, bias analysis~\citep{dunlap2025discovering,girrbach2025large}, and robot failure diagnosis~\citep{gupta2025perception}.
While \methodshort likewise uses text as its reasoning medium, it fundamentally differs from \tcmcshort in two key aspects:
\textit{i)} it \emph{automatically} discovers the grouping criteria rather than relying on a user‑supplied one;
\textit{ii)} it infers both the number and the names of clusters, requiring \textit{no} user-specified parameters.

\myparagraph{Topic Discovery.}  
\taskshort{} is also related to \textit{Topic Discovery}~\citep{blei2003latent,wang2009multi,eklund2022topic} in NLP, which identifies latent themes~\citep{viswanathan2024large,zhang2023clusterllm,zhong2024explaining} or events~\citep{tipirneni2024context,nakshatri2023using} from \textit{text corpora}.  
Our work similarly aims to uncover common themes from large, unstructured data but operates on \textit{images}, which is more challenging since \textit{i)} visual semantics are implicit, unlike text where meaning is explicit, and \textit{ii)} no current vision model can reliably reason over large image sets.

%% file: sec/3_task.tex
\section{\tasktitle}
\lblsec{problem_formulation}

\myparagraph{Task Definition.}
Given a collection of unlabeled images $\data = \{\bx_n\}_{n=1}^{N}$, the goal of \textit{\tasktitle{}} is to build a system, $\symbsys$, that automatically \textit{i)} discovers a set of $L$ grouping criteria $\manyrules = \{\onerule_l\}_{l=1}^{L}$ described in natural language, and \textit{ii)} finds interpretable substructures $\substructure_l$ for each criterion by uncovering semantically meaningful clusters and assigns images to them. Formally, we define an \taskshort system as:
\begin{equation*}
    \symbsys : 
    \data 
    \mapsto 
    \left\{
    \substructure_l=\Bigl\{
    \cluster_{k}^{l} = \bigl(\onename_{k}^{l}, \data_{k}^{l}\bigr)
    \Bigr\}_{k=1}^{K_l} 
    \,\middle|\, \onerule_l
    \right\}_{l=1}^{L},
    \lbleq{system}
\end{equation*}
where each cluster $\cluster_{k}^{l}$ is characterized by a semantic name $s_{k}^{l}$ and a subset of images $\data_{k}^{l} \subset \data$ that share the same semantics. A criterion $R_l$ refers to a \textit{theme} for grouping images, such that all the clusters under $R_l$ should align with the theme. As shown in \reffig{teaser}(top), if $R_l = \texttt{Activity}$, each cluster under this criterion should collect images $\data_{k}^{l}$ that depict an activity, such as $\onename_{k}^{l}=\text{``Surfing''}$. If $R_l = \texttt{Location}$, the same dataset should be organized into clusters like ``Restaurant'', ``Sports facility'', and so on.

An \taskshort system should find $\mathcal{R}$ and $\mathcal{O}_l$ automatically, both expressed in natural language. In contrast to TCMC setting, where criteria $\mathcal{R}$ and the corresponding cluster counts $K$ are \textit{preset} by human operators.

\begin{figure}[!t]
    \centering
    \includegraphics[width=\linewidth]{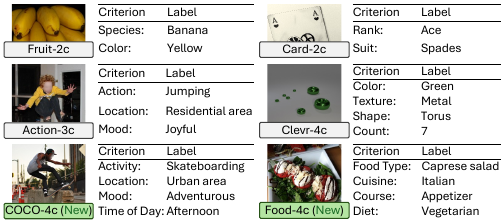}
    \caption{
        \textbf{\taskshort{} benchmarks.} We introduce two {\color{darkspringgreen} \textbf{new}} challenging benchmarks: \textbf{\ourcoco} and \textbf{\ourfood}.
        We show all annotated criteria and the corresponding labels for the example images.
    }
    \lblfig{demo_benchmarks}
    \vspace{-5.0mm}
\end{figure}

\begin{figure*}[!ht]
    \centering
    \includegraphics[width=\textwidth]{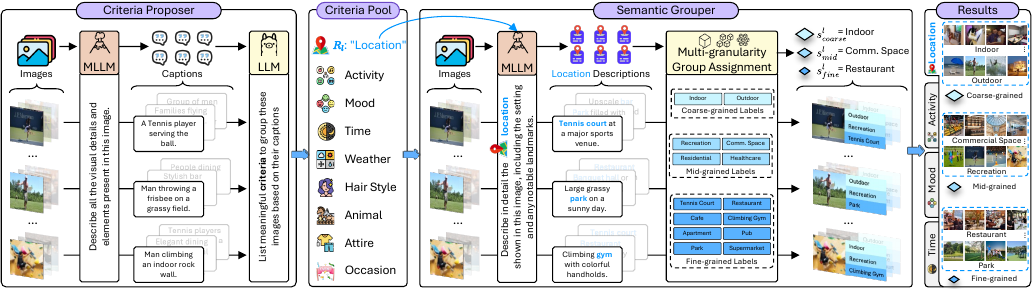};
\caption{
{\textbf{$\bm{\mathcal{X}}$-Cluster}} consists of a \textit{Criteria Proposer} and a \textit{Semantic Grouper}.
\textbf{(left)} Given a set of images, the Proposer discovers and outputs a pool of grouping criteria in natural language.
\textbf{(right)} The Grouper subsequently extracts criterion-specific descriptions from images relevant to each criterion, discovers the underlying semantic clusters, and groups each image at three semantic granularity levels.
\textbf{Results} shows an example, as how an unstructured image collection can be grouped into clusters of different semantic granularity corresponding to criterion ``Location".
See \refapp{app_prompt} for implementation and prompt details.
}
    \vspace{-5.0mm}
    \lblfig{system}
\end{figure*}


\myparagraph{Benchmark.}
Evaluating \taskshort methods requires benchmarks that can be partitioned under multiple criteria. Currently, only a few benchmarks~\citep{yu2024multiple} support the evaluation of \taskshort methods: \fruit{}~\citep{muresan2018fruit}, \card{}~\citep{kaggle2022card}, \action{}~\citep{kwon2023image}, and \clevr{}~\citep{vaze2024no}. As shown in \reffig{demo_benchmarks}, these benchmarks are limited by their object-centric nature with simple backgrounds (\eg \fruit{}), an insufficient number of criteria (\eg up to three in \action{}), and a lack of photorealism due to synthetic generation (\eg \clevr{}).

Given that the data encountered in real-world applications is more complex, we annotate and propose two \textit{new} benchmarks for \taskshort: \ourfood{} and \ourcoco{}. \ourfood{} is sourced from Food-101~\citep{bossard2014food}, which includes 101 \texttt{Food type} (original annotations), along with new annotations for 15 \texttt{Cuisine} types, 5 \texttt{Courses} types, and 4 \texttt{Diet} preferences, totaling \textit{four} clustering criteria. Additionally, we introduced \ourcoco{} using images from COCO-val~\citep{lin2014microsoft}, where we annotated \textit{four} criteria  with varying number of clusters: 64 \texttt{Activity}, 19 \texttt{Location}, 20 \texttt{Mood}, and 
6 \texttt{Time of day}. Examples of these newly constructed benchmarks are shown in \reffig{demo_benchmarks}. Further details, such as cluster names and the annotation pipeline, are provided in \refapp{app_datasets}.

%% file: sec/4_method.tex
\section{Method}
\lblsec{method}

The goal of an \taskshort system is to first discover meaningful grouping criteria (or {themes}) from an unstructured image collection by finding commonalities among the images, and then group them into semantic clusters as per the discovered criteria. This is particularly a challenging task because it requires reasoning over the visual content of all images \textit{simultaneously}. To address \taskshort, we diverge from representation learning-based MC approaches~\cite{yao2024multi,yao2025customized}, as no existing model can yet encode large image sets and reason over them reliably. Instead, we convert the visual content of all images into text and use \textit{text descriptions as a proxy} to discover the grouping criteria and the semantic substructures.

\myparagraph{System Overview:} As illustrated in \reffig{system}, our proposed \methodshort is a two-stage framework that is composed of two modules: \textit{Criteria Proposer} and \textit{Semantic Grouper}. The Criteria Proposer processes the \textit{entire} image set $\data$ to discover diverse common themes among the images and proposes grouping criteria $\manyrules$ in natural language (\eg \texttt{Location}). Once the criteria are proposed, the Semantic Grouper uncovers the substructure $\substructure_l$ of $\data$ by discovering distinct semantic clusters and assigning images to their respective clusters (\eg ``Tennis Court''), adhering to each criterion $\onerule_l \in \manyrules$.
As the \taskshort task operates without user priors to guide semantic granularity, we also design our Semantic Grouper to automatically discovers $\substructure_l$ across multiple granularity levels, from coarse (\eg, ``Outdoor'') to mid (\eg, ``Recreation'') to fine (\eg, ``Tennis Court''), and organizes the images accordingly.
In this work, we explore \textit{three} design choices for both the Proposer and the Grouper. Due to space constraints, only the main variant of \methodshort is illustrated in \reffig{system}, while the illustrations of the alternative variants and additional implementation details, including the exact prompts, are provided in \refapp{app_prompt}.
Next, we describe each variant in detail.

\subsection{Criteria Proposer}
As shown in \reffig{system}{(left)}, the Proposer takes as input a set of input images and generates distinct grouping criteria (or Criteria Pool) in natural language. Its core design principle is the ability to \textit{concurrently reason} across a large set of images. Next, we explore three systematic approaches.

\myparagraph{\captag{Caption-based Proposer (main):}}
To enable reasoning over a large image set for criterion discovery, we first leverage a \mllm{}~\citep{liu2024llava} to generate a comprehensive caption $e_{n}$ for each image, converting its visual content into text representations $e_{n} = \text{MLLM}(\bx_{n})$.
The resulting caption set $\{e_{n}\}_{n=1}^{N}$ serves as a rich and holistic semantic proxy for the image collection $\data$.
Using these compact textual proxies, we then prompt a \llm{}~\citep{meta2024llama31, openai2024gpt4o} to jointly analyze the aggregated visual content and propose multiple valid grouping criteria, denoted as $\rulespool = \text{LLM}(\{e_{n}\}_{n=1}^{N})$. As an example, the LLM could use different cues such as ``Tennis'', ``grassy field'', ``rock wall'' in the captions (see \reffig{system}) and its reasoning capability to discover the criterion ``Location", since they are usually associated to a physical location, albeit locations may not explicitly appear in any caption.

\myparagraph{\alttag{Tag-based Proposer (alternative):}}
Instead of using captions as textual proxies for reasoning, we further explore an approach that relies on image tags. 
Specifically, using the WordNet~\citep{miller1995wordnet} vocabulary as the candidate tag set, we employ an open-vocabulary tagger (\eg, CLIP~\citep{radford2021learning}) to assign ten tags $t_{i,n}$ to each image as
$\{t_{i,n}\}_{i=1}^{10} = \text{Tagger}(\bx_n, \text{WordNet})$. 
These tags act as concise semantic descriptors that summarize the key elements present in each image. 
We then aggregate all assigned tags and prompt a \llm{} to analyze them jointly and propose grouping criteria as
$\rulespool = \text{LLM}(\{\{t_{i,n}\}_{i=1}^{10}\}_{n=1}^{N})$.

\myparagraph{\alttag{Image-based Proposer (alternative):}}
Lastly, we explore an approach that reasons directly over images rather than their textual proxies. 
Since no existing model can reliably encode large image sets at once, we adopt a simple workaround: divide $\data$ into 64-image batches, stitch each batch into an $8 \times 8$ grid as a \textit{single} composite image, and feed the resulting image grids to a \mllm{}. 
The model is prompted to propose grouping criteria from each grid, and we aggregate and deduplicate these subset proposals to obtain the final criteria set $\rulespool$.

\myparagraph{Criteria Refinement:}
The accumulated criteria in $\rulespool$ may contain redundant or noisy entries,
such as semantically overlapping concepts (\eg, ``Outdoor'' \textit{vs.}\ ``Open space'') or irrelevant ones (\eg, ``High resolution'').
To clean them, we input all initially proposed criteria into a \llm{}, prompting it to consolidate similar ones and discard noise. 
This yields a refined criteria set $\manyrules = \text{LLM}(\rulespool)$, which is then stored in a pool for the subsequent substructure discovery stage.

\subsection{Semantic Grouper}

Each discovered criterion $\onerule_{l} \in \manyrules$ serves as a thematic indicator for a distinct semantic substructure $\substructure_{l}$ within the image set $\data$. To uncover these substructures, as shown in \reffig{system}{(right)}, the Grouper takes $\data$ and each criterion $\onerule_{l}$ as inputs, discovers cluster names $\{s_{k}^{l}\}_{k=1}^{K_{l}}$, and groups images $\data_{k}^{l}$ to their corresponding clusters. As a result, the interpretable substructure $\substructure_l = \{\cluster_{k}^{l} = \bigl(\onename_{k}^{l}, \data_{k}^{l}\bigr)\}_{k=1}^{K_l}$ emerges for each $\onerule_{l}$. The core design of the Grouper focuses on \textit{aligning} semantic substructure discovery with the given partitioning criterion. Like the Proposer, we explore three distinct approaches for the Grouper.

Furthermore, as clusters under a given criterion can be formed at varying semantic granularities based on user preferences, we have designed our Grouper to clusters $\data$ at three levels: coarse, middle, and fine-grained. This allows \methodshort to provide insights at different granularities. For example, under the \texttt{Cuisine} criterion, \methodshort can organize images at a coarse continental level (\eg, ``European'' or ``Asian''), a middle regional level (\eg, ``Mediterranean'' or ``Southeast Asian''), or a fine national level (\eg, ``Italian'' or ``Thai''). See \refapp{app_prompt_grouper} for design details.


\myparagraph{\captag{Caption-based Grouper (main):}}
Given a target criterion $\onerule_{l}$, we prompt the \mllm{} to generate criterion-specific captions that focus exclusively on the visual content relevant to $\onerule_{l}$ for each image, as $e_{n}^{l} = \text{MLLM}(\bx_{n}, \onerule_{l})$.
Next, we design a \textit{Multi-granularity Group Assignment (MGA)} module that uses the \llm{} to group images into clusters across multiple semantic granularity levels through a three-step process:
\textit{i) Initial Naming}: The \llm{} assigns a provisional class name to each caption as $s_{n}^{l} = \text{LLM}(e_{n}^{l}, \onerule_{l})$, producing an initial set of names $\manynames_{\text{init}}^{l}$;
\textit{ii) Multi-granularity Cluster Refinement}: The \llm{} refines $\manynames_{\text{init}}^{l}$ into three structured granularity levels: $\left(\manynames_{\text{coarse}}^{l}, \manynames_{\text{mid}}^{l}, \manynames_{\text{fine}}^{l}\right) = \text{LLM}(\manynames_{\text{init}}^{l}, \onerule_l)$, which serve as candidate cluster names;
\textit{iii) Final Assignment}: \llm{} assigns each image $\bx_{n}$ to a cluster by linking its criterion-specific caption to the structured class names at different granularity levels as $\left(s_{\text{coarse}}^{l}, s_{\text{mid}}^{l}, s_{\text{fine}}^{l}\right) = \text{LLM}(e_{n}^{l}, \manynames_{\text{coarse}}^{l}, \manynames_{\text{mid}}^{l}, \manynames_{\text{fine}}^{l})$.
By aggregating these cluster assignments across $\data$ at different levels, we derive multi-granularity semantic substructures.
As we will show in \refsec{assessment_grouper}, the Caption-based Grouper outperforms other alternatives, making it our main method.

\myparagraph{\alttag{Tag-based Grouper (alternative):}}
Given a target criterion $\onerule_{l}$, we prompt the \llm{} to generate a set of common categories (\eg, ``Commercial Space'') related to the criterion $\manynames_{\text{mid}}^{l} = \text{LLM}(\onerule_{l})$ as the mid-grained tags.
Following \citet{liu2024shine}, we further query the \llm{} to infer potential super- and sub-categories (\eg, ``Indoor'' and ``Restaurant'') for each mid-grained tag, thereby obtaining the corresponding coarse- and fine-grained tag sets, $\manynames_{\text{coarse}}^{l}$ and $\manynames_{\text{fine}}^{l}$.
Finally, we employ an open-vocabulary image tagger~\citep{radford2021learning} to assign the most relevant tag at each granularity level to each image,
$\left(s_{\text{coarse}}^{l}, s_{\text{mid}}^{l}, s_{\text{fine}}^{l}\right) = \text{Tagger}(\bx_{n}, \manynames_{\text{coarse}}^{l}, \manynames_{\text{mid}}^{l}, \manynames_{\text{fine}}^{l})$, yielding multi-granularity substructures after aggregation.

\myparagraph{\alttag{Image-based Grouper (alternative):}} 
Given a target criterion $\onerule_{l}$, we first prompt a \llm{} to generate a question $q_l$ tailored to $\onerule_l$, For \eg, for the criterion \texttt{Mood} the generated question is: ``What mood is conveyed by this image? Answer with an abstract, common, and specific category name, respectively''.
We then use $q_l$ to guide a \vqalong{} (\vqa{}) model~\citep{li2023blip} in directly inferring semantic cluster names and assignments for each image at different granularity levels as $\left(s_{\text{coarse}}^{l}, s_{\text{mid}}^{l}, s_{\text{fine}}^{l}\right) = \text{VQA}(\bx_{n}, q_{l})$.

%% file: sec/5_experiment.tex
\section{Experiments}
\label{sec:exp}

\subsection{Experimental Protocol}
\myparagraph{Implementation Details:}
We run with our proposed \methodshort framework using: \textit{i)} CLIP ViT-L/14~\citep{radford2021learning} as the Tagger, \textit{ii)} \llava{-7B}~\citep{liu2024llava} as the \mllm{}, \textit{iii)} \llamathreeone{-8B}~\citep{meta2024llama31} as the \llm{}, and \textit{iv)} \blip{-2 Flan-T5$_{\text{XXL}}$}~\citep{li2023blip} as the \vqa{} model. For the Image-based Proposer we use \multillava{-7B}~\citep{li2024llava} as the \mllm{} due to its strong multi-image reasoning capability. Additionally, we explore a variant of the Image-based Grouper using \llava{-7B} as the \vqa{} model.
We provide further details of \methodshort, including the exact prompt designs, in \refapp{app_prompt}.  


\myparagraph{Evaluation Metric for Criteria Discovery:}
We use \tprtitle{} (\tpr{})~\citep{csurka2024could} to evaluate the criteria discovery performance of different proposers. Specifically, we compute \tpr{} as $\text{TPR} =\frac{\mid\manyrules \cap \gtrules\mid}{|\gtrules|}$, measuring to what extent the predicted set covers the ground-truth criteria $\gtrules$.
It is important to note that the number of grouping criteria is subjective and can be as extensive as one's preferences allow (open-ended), making False Positives hard to define.
Thus, we use \tpr{} as the primary metric. A higher \tpr{} means better coverage of predicted criteria compared to the ground truth.

\myparagraph{Evaluation Metrics for Substructure Uncovering:}  
To assess each criterion-specific substructure uncovered by the Grouper, we evaluate its alignment with the ground-truth substructure along two dimensions: 
\textit{i) Semantic Consistency:} For each image, we compute the semantic similarity between its assigned cluster name and the ground-truth label under the current criterion using Sentence-BERT. The average similarity across the dataset, reported as \sacctitle (\sacc)~\citep{ liu2024democratizing}, measures how well the predicted substructure aligns semantically with the ground truth.  
\textit{ii) Structural Consistency:} We use \cacclong (\cacc)~\citep{han2021autonovel, vaze2022generalized} to measure the degree of structural match between the predicted and ground-truth substructures (clusters) using Hungarian matching algorithm~\citep{kuhn1955hungarian}.

Since the granularity of ground-truth annotations is unknown during \taskshort evaluation, we select the predicted substructure with the highest \cacc for assessment. Unlike \tcmcshort methods~\citep{kwon2023image, yao2025customized} that rely on ground-truth cluster counts for perfect matching, our strategy provides a fair and practical evaluation for open-ended \taskshort systems.

\begin{figure}[!t]
\centering
    \includegraphics[width=\linewidth]{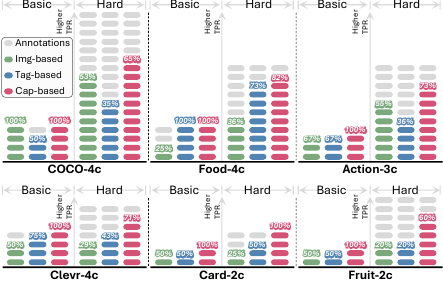}
    \caption{
    \textbf{Comprehensiveness Comparison of Criteria Proposers.}  
    \tpr performance of each proposer is evaluated against Basic and Hard ground-truth criteria, and visualized using a Progress Bar Chart. Each block represents one ground-truth criterion, with 
    {\color{imgGreen}\textbf{Co}}{\color{tagBlue}\textbf{lo}}{\color{capPink}\textbf{red}}
    blocks indicating successfully discovered criteria and
    {\color{annoGrey}\textbf{Gray}}
    blocks representing undiscovered criteria.
    }
    \lblfig{study_proposer_tpr}
\vspace{-5.0mm}
\end{figure}



\subsection{Study of the Criteria Proposer}
\lblsec{assessment_proposer}

We also evaluate the performance of our design for the Proposer module. To properly assess his effectiveness, we realize that  
for complex datasets like \ourcoco, four ground-truth criteria may not cover all valid grouping options. Therefore, we expanded the ground-truth criteria for each of the six benchmarks in \refsec{problem_formulation} using human annotators, resulting in \{10, 4, 11, 7, 17, 11\} distinct criteria for \{\fruit{}, \card{}, \action{}, \clevr{}, \ourcoco{}, \ourfood{}\}. We refer to the original per-image annotated criteria set (see \reffig{demo_benchmarks}) as \textbf{Basic} ground truth and the expanded set as \textbf{Hard} during evaluation. See \refapp{app_benchmark_criteria} for annotations.

\myparagraph{Which Criteria Proposer Performs the Best?}
In \reffig{study_proposer_tpr}, we compare different approaches for the Proposers in terms of the comprehensiveness of the discovered criteria using TPR. From \reffig{study_proposer_tpr}, we observe that our caption-based Proposer discovers the most comprehensive criteria, making it the \textit{closest} to the human-annotated set among all methods. It consistently outperforms other variants in both the Basic and Hard sets across all six benchmarks. Its superior performance is particularly evident under the Hard criteria set, where it surpasses the second-best Tag-based Proposer by +32.2\% \tpr. Intuitively, the Caption-based Proposer works better because captions capture more diverse and nuanced aspects of the image set, which further guides the LLM to comprehensively discover different grouping criteria. Contrarily, the Tag-based Proposer is less effective in complex benchmarks (\eg \ourcoco{} and \action{}) since tags provide less contextual and descriptive information. Similarly, the Image-based Proposer is subpar in terms of performance since it is limited to reasoning over a small subset of images and loses visual details when combining images into a grid.

\begin{figure}[!t]
\centering
    \includegraphics[width=\linewidth]{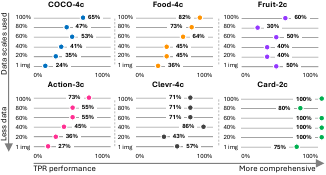}
    \caption{
        \textbf{Impact of Image Quantity on Criteria Discovery.} We evaluate the \tpr performance of the {\color{capPink} Caption-based Proposer} at different image scales against the Hard ground-truth criteria set.  
    }
    \lblfig{study_proposer_datascales}
\vspace{-6.0mm}
\end{figure}

\begin{figure*}[!th]
    \centering
    \includegraphics[width=\textwidth]{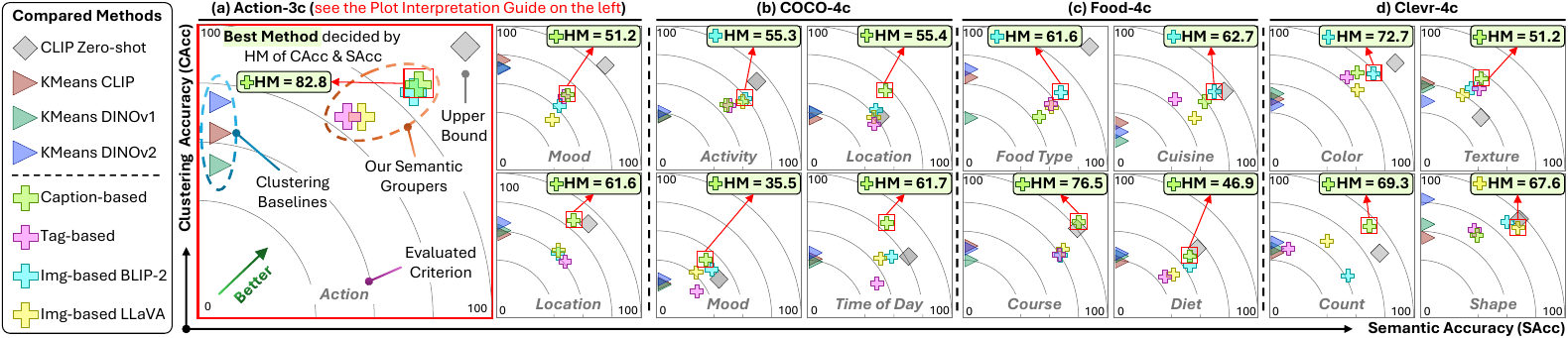}
    \caption{
    \textbf{Comparison of Semantic Groupers.}  
    We report \cacc, \sacc, and their \hmeantitle (\hmean) for different Semantic Groupers (\includegraphics[height=0.85em]{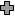}) on the Basic criteria across four benchmarks. CLIP zero-shot classification (\includegraphics[height=0.85em]{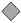}) serves as an oracle, while KMeans (\includegraphics[height=0.85em]{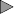}) with strong visual features is used as a \cacc baseline.
    The \besttag{best performer} for each criterion, determined by \hmean, is {highlighted in green}.
    Overall, our \textit{Caption-based Grouper} performs best, ranking first in \textit{10 of 15} evaluated criteria.
    See \refapp{app_expt_expanded_grouper_study} for clustering visualizations.
    }
    \lblfig{study_grouper}
\end{figure*}


\myparagraph{Impact of Image Quantity on Criteria Discovery:}  
\reffig{study_proposer_datascales} shows the \tpr performance of the Caption-based Proposer across different image scales. Interestingly, in \textit{object-centric} benchmarks like \card and \clevr, satisfactory performance is achieved with just a \textit{few} images. In fact, even a \textit{single} image often suffices for reasonable criteria discovery, as object-centric datasets tend to have uniform structures, \ie, seeing one playing card is enough to suggest criteria like \texttt{Suit}. 
However, this does not hold for more complex datasets like \ourcoco, \ourfood, and \action, which feature
diverse and
realistic scenarios. Here, reducing the number of images leads to a clear drop in \tpr performance, as capturing intricate and varied thematic criteria requires a larger image set. Since \methodshort operates \textit{without} prior knowledge of the dataset, we use the \textit{entire} dataset by default to ensure comprehensive 
discovery.  


\subsection{Study of the Semantic Grouper}
\lblsec{assessment_grouper}

\input{tables/comp_sota}

\myparagraph{Which Semantic Grouper Performs the Best?}  
In \reffig{study_grouper}, we evaluate different design choices for the Grouper using \cacc and \sacc for each criterion, determining the best performer based on \hmeantitle (\hmean). To contextualize performance, we establish an oracle using CLIP ViT-L/14 in a zero-shot classification setup, where grouping criteria, cluster names, and the number of clusters are all \textit{known}. We also use KMeans with ground-truth cluster numbers and visual features from CLIP-L/14, DINOv1-B/16~\citep{li2022emergent}, and DINOv2-G/14~\citep{oquab2023dinov2} as \cacc baselines.  

From \reffig{study_grouper}, we observe that the proposed Caption-based Grouper performs best, ranking first in 10 out of 15 tested criteria based on the \hmean across four benchmarks. It achieves an average \cacc of 59.9\%, closely matching the oracle performance of 58.1\%, highlighting the effectiveness of our text-driven approach. For \sacc, the Caption-based Grouper achieves an average of 60.5\%, surpassing its counterparts, but falling short of the oracle 74.2\% which benefits from exact ground-truth class names. This gap is expected due to the open nature of the semantic space, \ie, terms like ``Joyful'', ``Happy'', and ``Cheerful'' often describe the same \texttt{Mood} but lack full semantic equivalence. The \blip{-2} Image-based Grouper ranks second. Its criterion-specific questions improve labeling accuracy, but per-image predictions can introduce noise in clustering.  

\begin{figure}[!t]
    \centering
    \includegraphics[width=\linewidth]{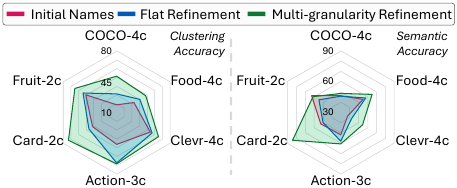}
    \vspace{+.05mm}
    \caption{
    {\textbf{Ablation study} of multi-granularity refinement.}
    }
    \lblfig{study_granularity}
    \vspace{-5.0mm}
\end{figure}

\myparagraph{Necessity of Multi-Granularity Cluster Refinement:}
To evaluate the effectiveness of multi-granularity cluster refinement design, we conduct controlled experiments using our Caption-based Grouper with three cluster naming strategies: \textit{i) Initial Names}, where the initially assigned names are used as the final output; \textit{ii) Flat Refinement}, where the \llm{} refines initial names into a single-level list with uniform granularity; and \textit{iii) Multi-Granularity Refinement}, our proposed approach. As shown in \reffig{study_granularity}, both refinement methods significantly improve clustering accuracy compared to using noisy initial names, highlighting the importance of granularity-consistent cluster names for revealing substructures. Moreover, our multi-granularity refinement outperforms flat refinement by enabling clustering at different levels of detail, providing greater flexibility in aligning with user-preferred grouping granularity.  

\subsection{Comparison with \tcmcshort Methods}
We first perform some experiments to compare our approach
with state of the art TMTC methods: \ictc~\citep{kwon2023image}, \ssdllm~\citep{luo2024llm}, \mmap~\citep{yao2024multi}, and \msub~\citep{yao2025customized}. Results are shown in \reftab{comp_sota}. Unlike our fully automated \methodshort method, which {discovers} criteria through the Proposer and requires \textit{no} pre-set cluster counts, \textit{all \tcmcshort methods used ground-truth text criteria and the number of clusters ($K_l$) as prior input}. 
The primary goal of this experiment is to evaluate dataset grouping performance. 
Our approach outperforms \mmap, \msub, and \ssdllm, while achieving results comparable to \ictc across six benchmarks. This demonstrates that our framework generates high-quality clusters for \taskshort \textit{without} requiring users to define criteria or cluster counts. Implementation details of the compared methods are provided in \refapp{app_implementation_compared}. 

\myparagraph{Further Analysis of} \methodshortbold is provided in the supplementary material: \textit{i)} \refapp{app_viz} presents qualitative results; \textit{ii)} \refapp{app_failure} examines failure cases; \textit{iii)} \refapp{app_discussion_fps} explores how \methodshort handles invalid (hallucinated) criteria; \textit{iv)} \refapp{app_limitations} investigates model biases; \textit{v)} \refapp{app_computational_cost} analyzes computational costs; \textit{vi)} \refapp{app_expt_sensitivity} studies system sensitivity to different \mllm{s} and \llm{s}; \textit{vii)} \refapp{app_multi_granularity_clustering} further investigates the impact of multi-granularity clustering; and \textit{viii)} \refapp{app_fine_grained} explores improvements for handling fine-grained criteria.  

%% file: tables/comp_sota.tex
\begin{table*}[!t]
    \centering
    \caption{
    \textbf{Comparison with \tcmcshort methods.}
    For each benchmark, we report the average \cacc (\%) and \sacc (\%) \textit{across all criteria}.
    We provide CLIP L/14 zero-shot performance as the pseudo upper-bound reference (UB).
    \textbf{Note}: \textbf{\dag}-marked methods used the ground-truth criteria and the number of clusters ($K_l$) as prior input.
    \mmap and \msub do not build semantic clusters.
    See expanded results in \refapp{app_expt_expanded_grouper_sota}.
    }
    
    \lbltab{comp_sota}
    \tablestyle{6.35pt}{.6}

    \begin{tabular}{lcccccccccccccc}

    \toprule
    & \multicolumn{2}{c}{\ourcoco} 
    & \multicolumn{2}{c}{\ourfood} 
    & \multicolumn{2}{c}{\clevr} 
    & \multicolumn{2}{c}{\action} 
    & \multicolumn{2}{c}{\card} 
    & \multicolumn{2}{c}{\fruit} 
    & \multicolumn{2}{c}{Avg}
    \\
    
    & \cacc & \sacc 
    & \cacc & \sacc 
    & \cacc & \sacc 
    & \cacc & \sacc 
    & \cacc & \sacc 
    & \cacc & \sacc 
    & \cacc & \sacc 
    \\
    
    \cmidrule(r){1-1}
    \cmidrule(r){2-3}
    \cmidrule(r){4-5}
    \cmidrule(r){6-7}
    \cmidrule(r){8-9}
    \cmidrule(r){10-11}
    \cmidrule(r){12-13}
    \cmidrule(r){14-15}

    \color{gray} UB 
    & \color{gray} 40.1 & \color{gray} 60.6
    & \color{gray} 64.1 & \color{gray} 80.2 
    & \color{gray} 56.7 & \color{gray} 72.5 
    & \color{gray} 79.8 & \color{gray} 82.3 
    & \color{gray} 41.4 & \color{gray} 66.9 
    & \color{gray} 69.4 & \color{gray} 88.3 
    & \color{gray} 50.2 & \color{gray} 64.4
    \\
        
    \cmidrule(r){1-1}
    \cmidrule(r){2-3}
    \cmidrule(r){4-5}
    \cmidrule(r){6-7}
    \cmidrule(r){8-9}
    \cmidrule(r){10-11}
    \cmidrule(r){12-13}
    \cmidrule(r){14-15}

    \mmap\textbf{\dag}~\citep{yao2024multi}
    & 33.9 & - 
    & 43.8 & - 
    & 62.8 & - 
    & 60.6 & - 
    & 36.9 & - 
    & 51.0 & - 
    & 48.2 & -
    \\

    \msub\textbf{\dag}~\citep{yao2025customized}
    & 36.0 & - 
    & 47.3 & - 
    & \textbf{72.2} & - 
    & 64.3 & - 
    & 39.6 & - 
    & 54.4 & - 
    & 52.3 & -
    \\
    
    \ictc\textbf{\dag}~\citep{kwon2023image}
    & 48.9 & {\bf 53.2}
    & {\bf 50.5} & 61.7 
    & 58.3 & 36.8 
    & {76.4} & 56.3 
    & {\bf 74.8} & 81.2 
    & 63.3 & 55.1 
    & {\bf 62.0} & 57.4
    \\
    											
    \ssdllm\textbf{\dag}~\citep{luo2024llm}
    & 41.6 & 52.1
    & 47.5 & 55.5
    & 54.8 & 37.6
    & \textbf{78.1} & 52.9
    & 67.3  & 76.3
    & 62.0 & 46.8
    & 58.6 & 53.6
    \\
    
    \cmidrule(r){1-1}
    \cmidrule(r){2-3}
    \cmidrule(r){4-5}
    \cmidrule(r){6-7}
    \cmidrule(r){8-9}
    \cmidrule(r){10-11}
    \cmidrule(r){12-13}
    \cmidrule(r){14-15}

    \methodshort (Ours) 
    & {\bf 51.2} & 48.4 
    & 48.1 & {\bf 64.9} 
    & {64.9} & {\bf 54.3} 
    & 68.3 & {\bf 60.6} 
    & 73.3 & {\bf 84.3} 
    & {\bf 65.1} & {\bf 61.1} 
    & 61.8 & {\bf 62.3}
    \\
    
\bottomrule
\vspace{-5.0mm}
\end{tabular}
\end{table*}

%% file: sec/6_application.tex
\begin{figure*}[!ht]
    \centering
    \includegraphics[width=\textwidth]{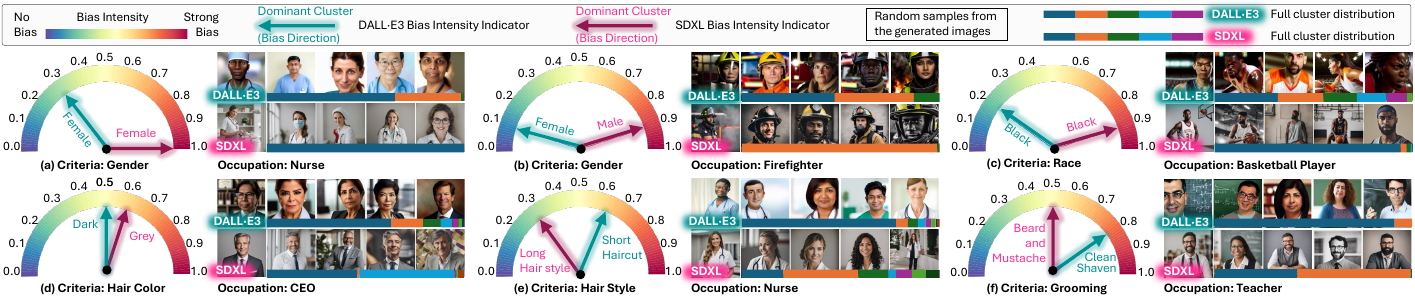}
    \caption{
    \textbf{Bias Discovery in T2I-Generated Images.} Bias intensity, dominant clusters, and example images are shown for few occupations.  
    }
    \lblfig{app_occupation}
    \vspace{-4.5mm}
\end{figure*}

\section{Applications}
\lblsec{applications}
We apply \methodshort to three applications, demonstrating its ability to generate novel, human-interpretable criteria for real-world analysis.
Below, we present results for the first two applications, while additional results for the third application on confirming and mitigating gender bias in 162k CelebA~\citep{liu2015deep} images are provided in \refapp{app_dataset_bias}.

\subsection{Discovering Biases in T2I Diffusion Models}  
\textbf{Do T2I models exhibit biases beyond the widely studied ones, such as gender and racial stereotypes?}~\citep{naik2023social, nicoletti2023bias} To investigate this, we selected \textit{nine} occupations (e.g., Nurse, CEO) from prior studies~\citep{bianchi2023easily, bolukbasi2016man} and generated 100 images per occupation using the prompt ``A portrait photo of a $<$OCCUPATION$>$'' with DALL-E3~\citep{betker2023improving} and SDXL~\citep{podell2023sdxl}, resulting in 1.8k images. Applying \methodshort, we automatically identified 10 grouping criteria (bias dimensions) and their distributions for each occupation. To quantify bias, we measured the normalized entropy of each distribution~\citep{d2024openbias} as bias intensity and identified the dominant cluster (the largest group) as the potential bias direction. We conducted a user study with 54 participants to validate our findings. \methodshort's predicted bias intensity closely matched human ratings with an Absolute Mean Error of 0.1396 (0--1 scale) and aligned with human-identified bias directions 72.3\% of the time.
User study details are provided in \refapp{app_occupation}.

\myparagraph{Findings:} 
As shown in \reffig{app_occupation}, our method identifies both well-known and novel biases in occupational images without relying on predefined categories. For instance, \reffig{app_occupation}{(a–c)} reveals strong gender and racial imbalances in SDXL-generated images for roles like Nurse, Firefighter, and Basketball Player, exceeding official statistics~\citep{blsEmployedPersons}. In contrast, \dalle exhibits improved bias mitigation, likely due to its built-in ``guardrails''~\citep{dalle_guards}. More notably, \reffig{app_occupation}{(d–f)} highlights previously unrecognized bias dimensions. For example, SDXL strongly associates CEOs with ``Grey'' hair, while \dalle favors ``Dark'' hair. Additionally, \dalle shows stronger biases in \texttt{Hair style} and \texttt{Grooming} for occupations like Nurse (\reffig{app_occupation}{(e)}) and Teacher (\reffig{app_occupation}{(f)}). These findings suggest that while industrial T2I models with guardrails may address well-known biases, they may still overlook emerging or less-discussed ones, underscoring the need for broader bias analysis.
For additional findings and experimental details, see \refapp{app_occupation}.

\subsection{Analyzing Social Media Image Popularity}  
\textbf{What makes a photo popular?} To explore this, we apply \methodshort to 4.1k Flickr photos from the SPID dataset~\citep{ortis2019prediction}, where popularity is measured by image view count. \methodshort discovered 10 grouping criteria and organizes photos into semantic clusters under each. Using the grouping results, \reffig{app_popularity} compares the sample popularity distributions of the {\color{trendingPink} Top Trending} and the {\color{trendingBlue} Top Mainstream} clusters across three criteria.

\myparagraph{Findings:} 
As shown in \reffig{app_popularity}, combining \methodshort's grouping with popularity scores provides a direct interpretation of the visual elements that drive trends versus those that define widely uploaded images. Interestingly, we find that trending elements often contrast with mainstream ones, such as {``Musical activities''} \textit{vs.}\ {``Rest and relaxation''} or {``High-intensity expressions''} \textit{vs.}\ {``Neutral emotion''}. These results suggest that attention-grabbing visuals stand out due to novelty or intensity, especially in today's short attention span era~\citep{mcspadden2015you, farid_2024}, underscoring \methodshort as a powerful tool for deep dataset analysis and understanding social behavior. For full findings and additional analysis, see \refapp{app_popularity}.

\begin{figure}[!t]
\centering
    \includegraphics[width=\linewidth]{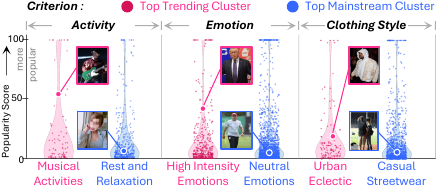}

    \caption{
    \textbf{Social media image popularity analysis}. We show the popularity score distributions for {\color{trendingPink}Top Trending (have \textit{highest} average popularity score)} and {\color{trendingBlue}Top Mainstream (contain \textit{most} images)} clusters,  discovered by \methodshort across three criteria.
    }

    \lblfig{app_popularity}
    \vspace{-5.8mm}
\end{figure}

%% file: sec/7_conclusion.tex
\section{Conclusion}
\lblsec{conclusion}
We introduce the \taskshort task and propose \methodshort, a system that discovers interpretable grouping criteria and substructures in image collections, effectively extracting valuable insights across six datasets and three applications.

%% file: sec/X_appendix.tex

\clearpage
\setcounter{page}{1}
\setcounter{table}{0}
\setcounter{figure}{0}
\maketitlesupplementary

\appendix
\section*{Table of Contents}
\small{
\startcontents[chapters]
\printcontents[chapters]{}{1}{}
}

\section*{Overview}  
This supplementary material provides additional details supporting the implementations, experiments, findings, and discussions in the main paper.  

In \refapp{repro_stat}, we provide a reproducibility statement to ensure the transparency and replicability of our work. Additionally, in \refapp{ethics}, we present an ethics statement to discuss and address the potential ethical implications and concerns associated with the proposed task and methodology, highlighting possible societal impacts and mitigation strategies.

\refapp{addrelatedwork} expands on Related Work, covering relevant tasks and methods. \refapp{app_datasets} describes the benchmarks used in our study, including the construction of the newly proposed \ourcoco and \ourfood datasets, as well as the process for creating hard ground-truth criteria for proposer evaluation. \refapp{app_evaluation_details} details the evaluation metrics used in this work.  

\refapp{app_prompt} presents the prompts and implementation details for \methodshort, covering both Criteria Proposers and Semantic Groupers. \refapp{app_implementation_compared} provides implementation specifics of the compared methods. Additional quantitative results, including evaluations of the Criteria Proposer, Semantic Grouper, and comparisons with other clustering methods, are in \refapp{app_expt_details}.  

For qualitative analysis, \refapp{app_viz} presents further visualizations of predicted clusters, while \refapp{app_failure} examines failure cases. \refapp{app_multi_granularity_clustering} investigates the impact of multi-granularity clustering. \refapp{app_discussion_fps} explores the effect of invalid (hallucinated) criteria on system performance, and \refapp{app_limitations} studies the influence of foundation model hallucinations and biases.  

\refapp{app_computational_cost} analyzes the computational cost and runtime of \methodshort. \refapp{app_expt_sensitivity} and \refapp{app_fine_grained} extend our analysis with studies on system sensitivity and fine-grained image collections. \refapp{app_application} provides additional findings, implementation details, and user study results for the three explored applications. \refapp{app_discussion_system_design} discusses how LLMs enhance image clustering. Finally, \refapp{app_future_work} outlines potential future research directions for our proposed \taskshort task.

The Table of Contents on the next page outlines the main topics in this supplementary material, with hyperlinks for direct navigation to each section.  

\section{Reproducibility Statement}
\lblsec{repro_stat}
Upon publication, we will \textit{open-source} all essential resources for reproducing this work. Specifically, we will provide the full code implementation of \methodshort, along with the exact prompts used in each module. Additionally, we will release our two newly proposed benchmarks, \ourcoco and \ourfood, including annotations for each grouping criterion. Lastly, we will provide the code for our evaluation protocol, experiments, and application studies.

\section{Ethics Statement}
\lblsec{ethics}
We do not anticipate any immediate negative societal impacts from our work. However, we encourage future researchers building on this work to remain vigilant, as we have, about the potential for \methodshort, which integrates \llm{s} and \mllm{s}–particularly their human-like reasoning abilities– to be used both for good and for harm.

The motivation behind our studies on biases in existing datasets and text-to-image (T2I) generative model outputs is to \textit{reveal and address} these biases that objectively exist in the datasets and models. We emphasize that our aim is \textit{to study and mitigate these issues}, and in doing so, we \textit{do not create} any new biases or disturbing content. Specifically, in \refsec{applications}, we use well-established benchmarks, such as CelebA~\citep{liu2015deep}, for our study of dataset bias, and for bias discovery in T2I generative models, we select occupation-related subjects known to be associated with biases from prior studies~\citep{bianchi2023easily, bolukbasi2016man}. However, we acknowledge that our methodology and findings could potentially be misused by malicious actors to promote harmful narratives or discrimination against certain groups. We strongly oppose any such misuse or misrepresentation of our work. Our research is conducted with the aim of advancing technology while prioritizing public welfare and well-being.

For the creation of our two new benchmarks, \ourcoco and \ourfood, we sourced images exclusively from the COCO-val-2017~\citep{lin2014microsoft} and Food-101~\citep{bossard2014food} datasets, strictly adhering to their licensing agreements. Additionally, we utilized voluntary human annotators for proposing valid grouping criteria and creating annotations along these criteria, rather than employing annotators from crowdsourcing platforms. This decision was made to ensure sustainability, fair compensation, and high-quality work, as well as to safeguard the psychological well-being of participants. Similarly, for our user study on T2I model bias evaluation, we recruited voluntary participants via questionnaires to collect human evaluation results. The user study was conducted entirely anonymously, with participants providing informed consent. Our project, including data annotation and the user study involving human subjects, was approved by the Ethical Review Board of our university.

Lastly, we emphasize that our proposed framework, \methodshort, relies on open-source \llm{s} and \mllm{s}, allowing full deployment on local machines. We refrain from using APIs from industrial \llm{s} or \mllm{s}, both to ensure reproducibility and to protect data privacy.

\section{Additional Related Work}
\lblsec{addrelatedwork}
\myparagraph{Topic Discovery.}
The setting of \tasklong (\taskshort) is also related to the field of Topic Discovery~\citep{blei2003latent, wang2009multi, eklund2022topic} in natural language processing, which aims to identify textual themes from large \textit{text corpora} (\eg, documents). Our work shares motivational similarities with topic discovery because both tasks seek to find common, thematic concepts from large volumes of data. In contrast, our work focuses on discovering thematic criteria from large \textit{visual content}. However, indeed, the core challenges of \taskshort and topic discovery are highly similar: they both require systems that can concurrently reason over large volumes of data. Nevertheless, \taskshort is an even more challenging task than topic discovery for two reasons: \textit{i)} semantics are not explicitly expressed in images, whereas they are in text; \textit{ii)} there is currently no vision model that can encode large sets of images and reliably reason over them. Thus, in this work, we translate images to text and use text as a proxy to elicit the large-scale reasoning capability of \llmlong{s}~\citep{meta2024llama31}.

\myparagraph{\mllmtitle.} Recent advancements in \mllmlong{s} (\mllm{s}) have been driven by the availability of large-scale vision-language aligned training data. The typical paradigm~\citep{liu2024visual} involves using a pre-trained \llmlong{} (\llm{})~\citep{meta2024llama3, vicuna2023, jiang2023mistral, meta2024llama31} alongside a pre-trained vision encoder~\citep{radford2021learning}. A projector is learned to align visual inputs with the \llm{} in the embedding space, which enhances visual understanding by utilizing the reasoning capabilities of \llm{s}. Several models have achieved significant success in zero-shot image captioning and visual question answering (VQA), including BLIP-2~\citep{li2023blip}, BLIP-3~\citep{xue2024xgen}, Kosmos-2~\citep{peng2023kosmos}, and the LLaVA series~\citep{liu2024visual, liu2024llava, li2024llava}. In our proposed \methodshort framework, we employ \mllm{} primarily as a \textit{zero-shot} image parser, converting visual information into text and using this text as a proxy to elicit \llm{s} for reasoning over large image collections and discovering grouping criteria. Additionally, we leverage the multi-image reasoning capability of \multillava~\citep{li2024llava} to establish a baseline image-based proposer for the \taskshort task, while utilizing BLIP-2 with customized prompts in a VQA style~\citep{shao2023prompting, zhu2023chatgpt} as the image-based grouper to form semantic clusters linked to specific visual content within the images.

\myparagraph{\llmtitle.} In the era of \llmlong{s} (\llm{s}) advancement~\citep{ouyang2022training}, modern \llm{s}, such as the Llama series~\citep{touvron2023llama, meta2024llama3, meta2024llama31}, Vicuna~\citep{vicuna2023}, Mistral-7B~\citep{jiang2023mistral}, and the GPT series~\citep{brown2020language}, have demonstrated remarkable zero-shot capabilities in tasks involving text analysis, completion, generation, and summarization. With advanced prompting techniques like Chain-of-Thought (CoT)~\citep{wei2022chain}, the reasoning abilities of \llm{s} can be further enhanced. In the proposed \methodshort framework, we design CoT prompts (see \refapp{app_prompt}) to harness the text generation and summarization capabilities of \llamathreeone as a reasoning engine. This aids \methodshort in several key areas: discovering grouping criteria from large sets of image captions, automatically prompting VQA models, generating criterion-specific tags, uncovering cluster semantics, and grouping images based on their captions. Unlike prior works~\citep{zhuge2023mindstorms} that focus on set difference captioning~\citep{dunlap2024describing}, fine-grained concept discovery~\citep{liu2024democratizing}, or video understanding~\citep{wang2024videoagent}, we leverage \llm{s} to tackle the challenging \tasklong task. While \ictc~\citep{kwon2023image} also uses the \llm{} (GPT-4~\citep{openai2023gpt4}) for grouping visual data, our proposed \methodshort differs in two key aspects: \textit{i)} \methodshort does \textit{not} require user-defined grouping criteria or the number of clusters, and \textit{ii)} \methodshort provides \textit{multi-granularity} outputs to meet various user preferences.

\myparagraph{Text-Driven Image Retrieval.} 
Given a query text (\eg, ``sofa'' or ``person wearing a blue T-shirt''), text-driven image retrieval methods~\citep{karthik2024visionbylanguage, liu2023zero, wu2023cap4video} aim to find images from an image collection that are relevant to the query. In other words, in the scenario we are considering, given the image collection and a list of text queries, one can organize images according to the text using text-driven image retrieval techniques. In this context, the query can be considered as a sort of ``cluster name''. However, this differs significantly from the proposed task of \tasklong (\taskshort), because \taskshort requires both discovering the textual criteria and the corresponding textual clusters. Thus, without knowing text queries as prior information, text-driven image retrieval methods are not able to accomplish \taskshort.

\section{Benchmark Details}
\lblsec{app_datasets}

\subsection{Construction of \ourcoco and \ourfood}
\lblsec{app_benchmark_construction}

\input{tables/benchmarks/benchmark_summary_basic}

To create high-quality benchmarks for \ourcoco and \ourfood, we designed a four-step annotation pipeline:

\myparagraph{(1) Criteria Identification:} We first split COCO-val-2017~\citep{lin2014microsoft} and Food-101~\citep{bossard2014food} images into batches of 100. Each batch was stitched into a 10$\times$10 grid to form a single image. These grid images were then distributed to 5 human annotators, who were tasked with identifying grouping criteria. For each dataset, we selected the 4 most frequently occurring criteria, as shown in \reftab{benchmark_summary_basic}, to proceed with per-image annotation.

\myparagraph{(2) Label Candidate Generation:} To facilitate the annotation process, we used GPT-4V~\citep{openai2023gpt4} to generate an initial list of candidate labels for each criterion. Specifically, for each criterion of \ourcoco and \ourfood, GPT-4V was prompted to assign a label that reflected the criterion for each image. This resulted in a list of criterion-specific label candidates for each dataset.

\myparagraph{(3) Image Annotation:} Next, 10 human annotators were tasked with assigning a label from the criterion-specific candidates to each image in \ourcoco and \ourfood for each criterion. The entire annotation process took 25 days to complete.

\myparagraph{(4) Label Merging:} Image annotation is inherently subjective, with annotators potentially assigning different labels for the same criterion. For example, one annotator might label the \texttt{Mood} criterion as 
\say{Happy}, while another might label it as \say{Joyful} or 
\say{Delightful}. 
To resolve such discrepancies, we used majority voting to determine the final label for each image. Specifically, the most frequently assigned label among the 10 annotators was chosen as the final label for each criterion.

Following these steps, we constructed \ourcoco and \ourfood. \textit{Note that we used the official COCO-val-2017~\citep{lin2014microsoft} and Food-101~\citep{bossard2014food} images for our benchmarks and did not collect any new images. We adhered strictly to the licenses of the datasets during their creation.} The exact number of classes is presented in \reftab{benchmark_summary_basic}. Additionally, the annotated class names for each criterion of \ourcoco are provided in \reftab{benchmark_names_coco4c}, and for \ourfood in \reftab{benchmark_names_food4c}.

\input{tables/benchmarks/benchmark_names_coco4c}
\input{tables/benchmarks/benchmark_names_food4c}

\subsection{Details on Hard Grouping Criteria Annotation}
\lblsec{app_benchmark_criteria}
\input{tables/benchmarks/benchmark_criteria}

In \reftab{benchmark_criteria}, we present the additional annotated \textbf{Hard} grouping criteria ground truth alongside the \textbf{Basic} criteria for each benchmark.

While we have established more rigorous and challenging benchmarks such as \ourcoco and \ourfood, which feature up to \textit{four} distinct grouping criteria, these annotated criteria sets do not encompass all potential grouping criteria within the image collections. This is particularly true for more complex and realistic datasets like \ourcoco, \ourfood, and \action. As a result, the performance differences between different criteria proposers on these basic criteria, as shown in \reffig{study_proposer_tpr}, tend to be close to each other, limiting our understanding of each proposer's ability to generate comprehensive grouping criteria.

To address this limitation, we employed human annotators to further identify and propose grouping criteria across the six benchmarks, resulting in a more extensive ground-truth set for each benchmark. This provides a better basis for evaluating the comprehensiveness of the different proposers. We refer to this set of larger annotation criteria as the \textbf{Hard} criteria, in contrast to the \textbf{Basic} criteria, which involve per-image annotations. Note that for the \textbf{Hard} criteria, per-image label annotation is not provided due to the high cost of annotation. The procedure for obtaining the \textbf{Hard} grouping criteria is as follows:

\myparagraph{(1) Criteria Discovery:} We divided each dataset into batches of 100 images, displaying each batch in a 10$\times$10 grid. Five human annotators were assigned to each batch and instructed to identify as many valid grouping criteria as possible. The proposed criteria from each annotator were then combined to form a comprehensive set of grouping criteria for the dataset.

\myparagraph{(2) Criteria Merging:} After collecting the annotated criteria from all five annotators, we aggregated the criteria and manually cleaned the set by merging semantically similar criteria (\eg, \texttt{Location} and \texttt{Place}) and discarding binary grouping criteria, as the inclusion of binary criteria can result in an unmanageable number of grouping criteria for complex datasets.

By following this process, we developed a more comprehensive grouping criteria set as the \textbf{Hard} ground-truth for each benchmark, as shown in \reftab{benchmark_criteria}. This resulted in sets containing 8 criteria for \fruit{}, 4 criteria for card, 11 criteria for \action{}, 7 criteria for \clevr{}, 17 criteria for \ourcoco{}, and 11 criteria for \ourfood{}. These expanded ground-truth sets enable us to more effectively evaluate the capabilities of various criteria discovery methods, providing a clearer understanding of different criteria proposers.

\section{Further Details of Evaluation Protocol}
\lblsec{app_evaluation_details}
\myparagraph{Further Discussion on Clustering Accuracy (CAcc).} Clustering Accuracy (CAcc)~\citep{han2021autonovel} is evaluated by applying the Hungarian algorithm~\citep{kuhn1955hungarian} to determine the optimal assignment between the predicted cluster indices and ground-truth labels. As extensively discussed in the GCD~\citep{vaze2022generalized} literature, if the number of predicted clusters (groups) exceeds the total number of ground-truth classes (groups), the extra clusters (not matched by the Hungarian algorithm) are assigned to a null set, and all instances in those clusters are considered incorrect during evaluation. On the other hand, if the number of predicted clusters is lower than the number of ground-truth classes, the extra classes are assigned to a null set, and all instances with those ground-truth labels are similarly considered incorrect. Thus, CAcc is maximized only when the number of predicted clusters matches the number of ground-truth clusters.

In the \tasktitle (\taskshort) task newly proposed in this work, we do not assume access to the ground-truth number of clusters as prior input. Consequently, our proposed method \methodshort does not rely on the ground-truth number of clusters to achieve an "optimal" CAcc with respect to the testing dataset. All clusters are automatically predicted by the \methodshort system. In stark contrast, in the comparison with criterion-conditioned clustering methods shown in \reftab{comp_sota}, both \ictc~\citep{kwon2023image} and \mmap~\citep{yao2024multi} use the ground-truth number of clusters as prior input.

\section{Further Implementation and Prompt Details}
\lblsec{app_prompt}
In this section, we provide detailed descriptions of the exact prompts used in our framework, along with additional implementation details for the proposed Criteria Proposer in \refapp{app_prompt_proposer} and the Semantic Grouper in \refapp{app_prompt_grouper}.

\begin{figure*}[!ht]
    \centering
    \includegraphics[width=\textwidth]{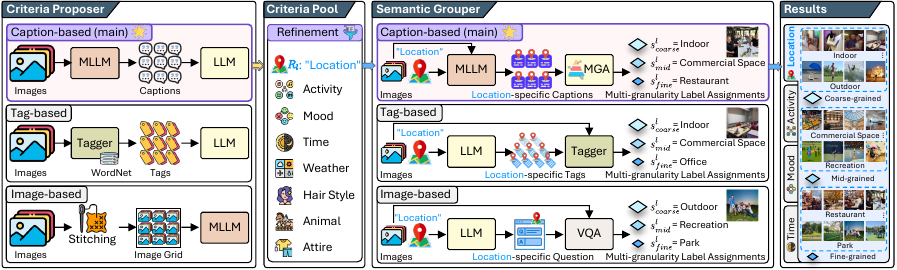};
\caption{
\textbf{All three variants of the proposed \textbf{$\bm{\mathcal{X}}$-Cluster} framework.}
We explore different design choices for both the Criteria Proposer (\textbf{left}) and the Semantic Grouper (\textbf{right}), and designate the best-performing Caption-based system (marked with \includegraphics[height=0.85em]{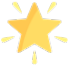}) as the main \methodshort configuration in our experiments.
}
    \lblfig{system_full}
\end{figure*}

Further, we present a system overview illustrating all three variants of our Criteria Proposer and Semantic Grouper, namely the caption based (main), tag based, and image based versions, in \reffig{system_full}.

\subsection{Further Details of Criteria Proposer}
\lblsec{app_prompt_proposer}
\myparagraph{Image-based Proposer:}
\input{tables/prompts/prompt_proposer_img_mllm}
In \reftab{prompt_proposer_img_mllm}, we present the exact prompt used in the image-based proposer for querying the \mllm{} \multillava{-7B}~\citep{li2024llava}. Given a target image set, we first randomly shuffle the images and divide them into disjoint subsets, each containing 64 images. Each subset is then stitched into an $8\times8$ image grid, treated as a single image, and fed into the \mllm{}. For each subset, the \mllm{} is prompted to propose 5 distinct grouping criteria for organizing the images within that subset, using the prompt shown in \reftab{prompt_proposer_img_mllm}. After iterating through all subsets, we take the union of the criteria proposed for each subset as the discovered criteria for the target image set. Finally, we deduplicate the discovered criteria and accumulate them into the criteria pool.

\myparagraph{Tag-based Proposer:}
\input{tables/prompts/prompt_proposer_tag_llm}
In \reftab{prompt_proposer_tag_llm}, we present the exact prompt used in the tag-based proposer for querying the \llm{} \llamathreeone{-8B}~\citep{meta2024llama31}. For a given target image set, we first utilize an open-vocabulary tagger, CLIP ViT-L/14~\citep{radford2021learning}, to assign 10 related natural language tags to each image. These tags are selected from the WordNet~\citep{miller1995wordnet} vocabulary, which contains 118k English synsets, and represent the semantic content of the images. We employ the standard prompt ``\texttt{A photo of \{concept\}}'' provided by CLIP for image tagging. Next, we embed the assigned tags into the prompt shown in \reftab{prompt_proposer_tag_llm} to carry the semantics of the entire image set and query the \llm{} to propose grouping criteria. The criteria proposed by the \llm{} are then added to the criteria pool. Note that in this case, we embed the tags for the entire dataset into a single prompt for criteria proposal, without reaching the \llm{} context length limits (\eg, 128k for \llamathreeone{-8B}) for the datasets used in our experiments. However, for larger datasets, it may be necessary to split the dataset into subsets, prompt the \llm{} for each subset, and use the union of the proposed criteria as the final output.

\myparagraph{Caption-based Proposer:}
\input{tables/prompts/prompt_proposer_cap_mllm}
\input{tables/prompts/prompt_proposer_cap_llm}
We present the prompt used in the caption-based proposer for the \mllm{} \llava{-7B}~\citep{liu2024llava} in \reftab{prompt_proposer_cap_mllm}, and the prompt for the \llm{} \llamathreeone{-8B}~\citep{meta2024llama31} in \reftab{prompt_proposer_cap_llm}. Specifically, we first use the \mllm{} with a general prompt to generate detailed descriptions for each image in the target dataset, effectively translating the visual information into natural language. The generated captions are then \textit{randomly shuffled} and split into disjoint subsets, each containing 400 captions. Next, we embed the captions from each subset into the prompt shown in \reftab{prompt_proposer_cap_llm} and use it to query the \llm{} to propose grouping criteria for the images represented by the captions. After iterating through all subsets, we take the union of the proposed criteria across subsets as the discovered criteria for the target image set. Finally, we deduplicate these criteria and add them to the criteria pool. Due to the context window limitations of \llm{s}, embedding all captions into a single prompt is infeasible. To address this, we limit each subset to 400 captions, which results in approximately 115k tokens per subset. This strategy allows us to remain within the context length limits of modern \llm{s} (\eg, 128k tokens for both \llamathreeone{} and \gptfouromni{}) while maximizing the number of samples per query to effectively propose clustering criteria.

\myparagraph{Criteria Pool Refinement:}
\input{tables/prompts/prompt_pool_refine_llm}
In \reftab{prompt_pool_refine_llm}, we present the exact prompt used for criteria pool refinement when querying the \llm{} \llamathreeone{-8B}~\citep{meta2024llama31}. Since the accumulated criteria pool $\rulespool$ may contain highly similar or noisy clustering criteria, we embed the criteria from the pool into the prompt shown in \reftab{prompt_pool_refine_llm} and ask the \llm{} to merge similar criteria and rephrase their names to enhance clarity. This process yields a refined set of grouping criteria, which is then passed to the next stage for image grouping.

\subsection{Further Details of Semantic Grouper}
\lblsec{app_prompt_grouper}
\myparagraph{Image-based Grouper:}
\input{tables/prompts/prompt_grouper_img_llm}
In \reftab{prompt_grouper_img_llm}, we present the prompt used to query the \llm{} \llamathreeone{-8B}~\citep{meta2024llama31} for automatically generating criterion-specific \vqa{} questions for the image-based grouper. The objective at this stage is to condition the \vqa{} model \blip{-2 Flan-T5$_{\text{XXL}}$}~\citep{li2023blip} to label each image across three different semantic granularity levels based on a specific criterion. To guide the \vqa{} model effectively, a criterion-specific question is required.

Rather than manually creating these questions, we embed the target criterion into the prompt shown in \reftab{prompt_grouper_img_llm} and query the \llm{} to automatically generate high-quality, criterion-specific questions. These questions are then used to direct the \vqa{} model, enabling it to accurately label each image according to the visual content relevant to the target criterion.

\myparagraph{Tag-based Grouper:}
\input{tables/prompts/prompt_grouper_tag_llm_mid}
\input{tables/prompts/prompt_grouper_tag_llm_coarse}
\input{tables/prompts/prompt_grouper_tag_llm_fine}
We present the prompts used in the tag-based grouper for querying the \llm{} \llamathreeone{-8B}. The prompt for generating criterion-specific tags is shown in \reftab{prompt_grouper_tag_llm_mid}, while the prompts for generating coarse-grained and fine-grained tags are shown in \reftab{prompt_grouper_tag_llm_coarse} and \reftab{prompt_grouper_tag_llm_fine}, respectively. 

In the tag-based grouper, we begin by embedding the target criterion into the prompt from \reftab{prompt_grouper_tag_llm_mid} to generate criterion-specific tags at a middle granularity. To enhance the diversity and coverage of the tags, we query the \llm{} 10 times and take the union of the generated tags after deduplication as candidates~\citep{liu2025test}. Following the SHiNe framework~\citep{liu2024shine}, for each middle-grained tag, we further embed it into the prompts from \reftab{prompt_grouper_tag_llm_coarse} and \reftab{prompt_grouper_tag_llm_fine} to generate 3 super-categories (coarse-grained) and 10 sub-categories (fine-grained) for each tag.

After generating coarse and fine-grained categories for all middle-grained tags, we take the union of the super-categories as the coarse-grained tag candidates and the union of the sub-categories as the fine-grained tag candidates. Lastly, we use the open-vocabulary tagger CLIP ViT-L/14 to assign the most relevant tags to each image based on cosine similarity, using candidates from each granularity level. After tagging all the images, we group those sharing the same tag into clusters, yielding the clustering result. Note that we do not utilize lexical databases such as WordNet~\citep{miller1995wordnet} or ConceptNet~\citep{speer2017conceptnet} for tag generation, as they do not support free-form input and may not capture certain discovered criteria.

\myparagraph{Caption-based Grouper:}
\input{tables/prompts/prompt_grouper_cap_mllm}
\input{tables/prompts/prompt_grouper_cap_llm_initial}
\input{tables/prompts/prompt_grouper_cap_llm_refine}
\input{tables/prompts/prompt_grouper_cap_llm_final}
We first present the \mllm{} prompt used for \llava{-7B}~\citep{liu2024llava} to generate criterion-specific captions in \reftab{prompt_grouper_cap_mllm}. Following this, we present the \llm{} \llamathreeone{-8B} prompts used in the caption-based grouper for the \textit{Initial Naming} step in \reftab{prompt_grouper_cap_llm_initial}, the \textit{Multi-granularity Cluster Refinement} step in \reftab{prompt_grouper_cap_llm_refine}, and the \textit{Final Assignment} step in \reftab{prompt_grouper_cap_llm_final}.

Specifically, we begin by generating criterion-specific captions for each image using \llava{-7B} with the prompt shown in \reftab{prompt_grouper_cap_mllm}. For each image, we then embed its criterion-specific caption and the relevant criterion into the \llm{} prompt shown in \reftab{prompt_grouper_cap_llm_initial}, querying the \llm{} to assign an initial name based on the target criterion. Once the initial names for all images in the dataset are obtained, we embed these names along with the target criterion into the prompt in \reftab{prompt_grouper_cap_llm_refine} to query the \llm{} for cluster name refinement across three semantic granularity levels: coarse, middle, and fine.

Finally, for each image, we embed the target criterion, its criterion-specific caption, and cluster candidates from each granularity level into the prompt shown in \reftab{prompt_grouper_cap_llm_final}, and use this to query the \llm{} for final cluster assignment at each granularity level.

\section{Further Details of the Compared Methods}
\lblsec{app_implementation_compared}
In this section, we provide the implementation details of the compared methods, \ictc~\citep{kwon2023image} and \mmap~\citep{yao2024multi}.

\myparagraph{Implementation details of \ictc}~\citep{kwon2023image}: 
In the original implementation of \ictc, LLaVA-1.5~\citep{liu2024visual} was used as the \mllm{}, and GPT-4-2023-03-15-preview~\citep{openai2023gpt4} as the \llm{}. However, since the GPT-4-2023-03-15-preview API has been deprecated, we re-implemented \ictc using the state-of-the-art \mllm{} \llava{-7B}~\citep{liu2024llava} and the latest version of GPT-turbo-2024-04-09 as the \llm{}, while strictly adhering to the original \ictc{} prompt design in our experiments to ensure a fair comparison.

\myparagraph{Implementation Details of \ssdllm}~\citep{luo2024llm}:  
Following similar setup of \ictc, we reproduced and compared with \ssdllm using GPT-turbo-2024-04-09 as the \llm{} and \llava{-7B} as the \mllm{}. Since \ssdllm requires a primary class name for each benchmark, we provided the ground-truth primary class in its prompt: ``Food'' for \ourfood, ``Object'' for \clevr, ``Person'' for \action, ``Playing card'' for \card, and ``Fruit'' for \fruit. For \ourcoco, which consists of everyday life scenes and lacks a consistent primary class, we used ``Object'' as a neutral placeholder in \ssdllm's prompt.

\myparagraph{Implementation details of \mmap and \msub}~\citep{yao2024multi, yao2025customized}: 
We closely followed the training configuration outlined in the original \mmap and \msub paper. Specifically, GPT-turbo-2024-04-09 was used as the \llm{} to generate reference words for each dataset. We then prompt-tuned CLIP-ViT/B32 using Adam with a momentum of 0.9, training the model for 1,000 epochs for each criterion across all datasets. Hyperparameters were optimized according to the loss score of \mmap, with the learning rate searched in \{0.1, 0.05, 0.01, 0.005, 0.001, 0.0005\}, weight decay in \{0.0005, 0.0001, 0.00005, 0.00001, 0\}, $\alpha$ and $\beta$ in \{0.0, 0.1, 0.2, ..., 1.0\}, and $\lambda$ fixed at 1 for all experiments. After training, KMeans, with the ground-truth number of clusters, was applied for each criterion and dataset to perform clustering.

\section{Further Quantitative Experimental Results}
\lblsec{app_expt_details}
In this section, we present additional numerical experiment results to supplement the figures in the main paper. In \refsec{app_expt_expanded_proposer}, we provide supplementary results for the evaluation of the Criteria Proposer in our framework. In \refsec{app_expt_expanded_grouper_study}, we present additional results for the evaluation of the Semantic Grouper across various criteria on the six tested benchmarks. Furthermore, we include expanded results comparing our framework to prior criteria-conditioned clustering methods. Lastly, we present detailed results from the ablation study of the multi-granularity refinement component in \refsec{app_expt_expanded_grouper_ablation}.

\subsection{Further Results for Criteria Proposer Study}
\lblsec{app_expt_expanded_proposer}
We provide detailed numerical results corresponding to \reffig{study_proposer_tpr} in \reftab{recall_main} and \reffig{study_proposer_datascales} in \reftab{recall_study_caption} for the six tested benchmarks.

Although captions generated by the \mllm{} may exhibit some information loss (\eg, ignoring small objects or attributes)~\citep{he2024incorporating} and hallucinations (\eg, introducing objects not present in the images)~\cite{liu2024survey}, these issues generally occur at the object or fine-grained attribute level. However, when reasoning about grouping criteria for \taskshort task, the focus is on identifying general thematic elements shared across the image set. As a result, these minor inconsistencies in the captions do not hinder the \llm{} in our framework from effectively reasoning about grouping criteria, helping the Caption-based Proposer to achieve the best performance among all the studied design choices. 
\input{tables/results_proposer/recall_main}

\input{tables/results_proposer/recall_study_caption}

\subsection{Further Results for Semantic Grouper Study}
\lblsec{app_expt_expanded_grouper_study}
In this section, we present the expanded numerical results comparing different semantic groupers to supplement the summary in \reffig{study_grouper}. Specifically, we provide detailed results for the evaluation of the six tested datasets as follows:

\begin{itemize}
    \item \ourcoco (\reffig{study_grouper}(a)) in \reftab{grouper_main_coco}
    \item \card (\reffig{study_grouper}(b)) in \reftab{grouper_main_card}
    \item \action (\reffig{study_grouper}(c)) in \reftab{grouper_main_action}
    \item \ourfood (\reffig{study_grouper}(d)) in \reftab{grouper_main_food}
    \item \fruit (\reffig{study_grouper}(e)) in \reftab{grouper_main_fruit}
    \item \clevr (\reffig{study_grouper}(f)) in \reftab{grouper_main_clevr}
\end{itemize}

In addition, we present the statistics of the predicted clusters at each granularity level in \reftab{stats_num_clusters}.

\input{tables/results_grouper/grouper_main_coco}
\input{tables/results_grouper/grouper_main_card}
\input{tables/results_grouper/grouper_main_action}
\input{tables/results_grouper/grouper_main_food}
\input{tables/results_grouper/grouper_main_fruit}
\input{tables/results_grouper/grouper_main_clevr}

\input{tables/rebuttal/stats_num_clusters}

\subsection{Further Comparative Results with \tcmcshort Methods}
\lblsec{app_expt_expanded_grouper_sota}
We provide expanded results in \reftab{grouper_sota} for each criterion and benchmark, detailing the comparison of criteria-conditioned clustering methods presented in \reftab{comp_sota} in the main paper.

\input{tables/results_grouper/grouper_sota}

\subsection{Further Results for Multi-granularity Clustering Study.}
\lblsec{app_expt_expanded_grouper_ablation}
We present expanded results in \reftab{grouper_ablation} for the ablation study on multi-granularity refinement, providing a detailed breakdown of the summary shown in \reffig{study_granularity} in the main paper.
\input{tables/results_grouper/grouper_ablation}

\section{Qualitative Analysis}
\lblsec{app_viz}
In this section, we visualize the grouping results predicted by the best configuration of our proposed framework (Caption-based Proposer and Caption-based Grouper). Specifically, we present example clustering results across different criteria for \ourcoco in \reffig{clusters_coco}, \ourfood in \reffig{clusters_food}, \action in \reffig{clusters_action}, \clevr in \reffig{clusters_clevr}, and \card in \reffig{clusters_card}. Additionally, we showcase example clustering results at different predicted granularity levels for \ourcoco in \reffig{clusters_coco_levels}.
\begin{figure*}[!th]
    \centering
    \includegraphics[width=\textwidth]{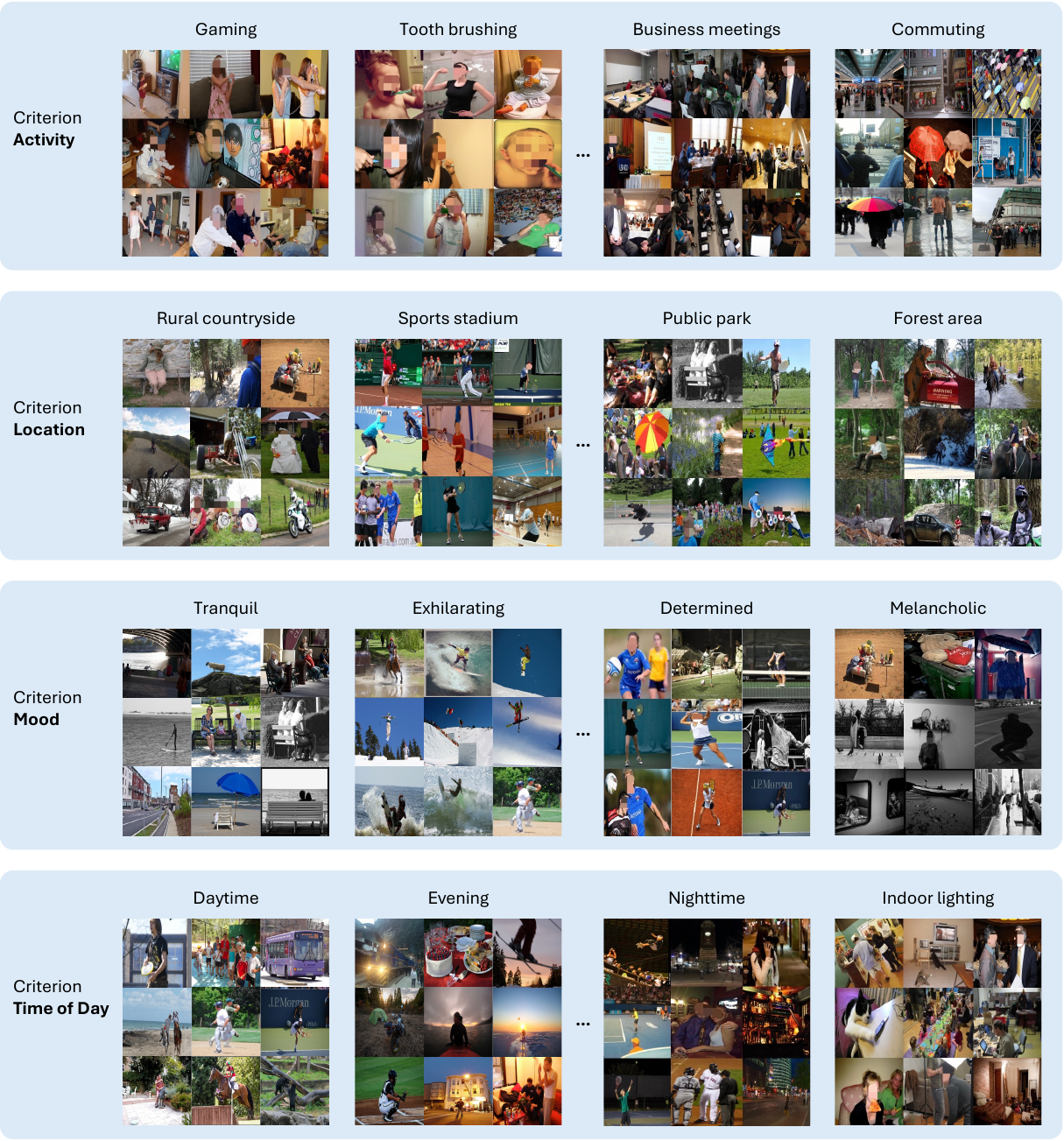}
    \caption{
    \textbf{Example predicted clusters of \ourcoco.}
    }
    \lblfig{clusters_coco}
\end{figure*}

\begin{figure*}[!th]
    \centering
    \includegraphics[width=\textwidth]{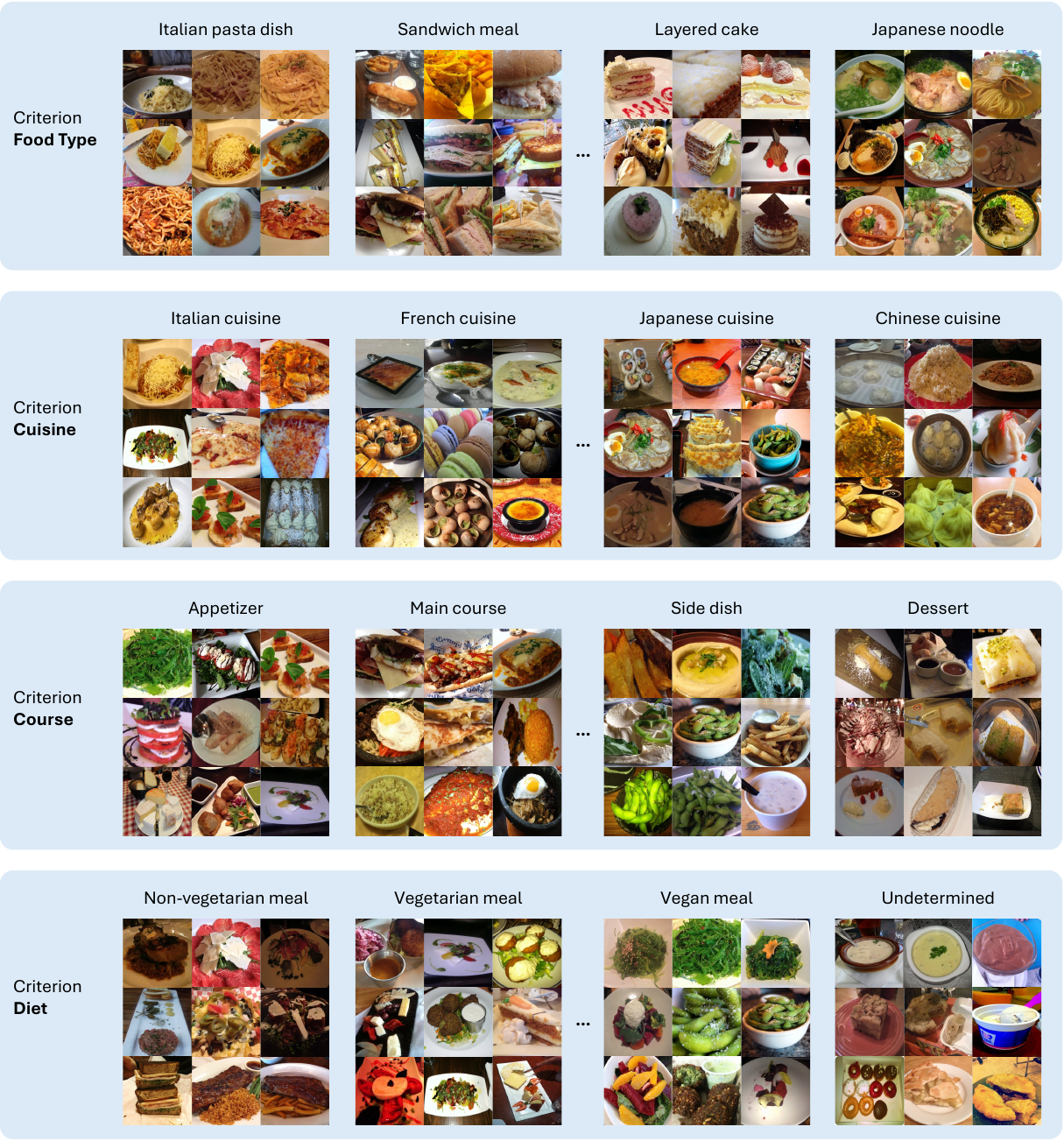}
    \caption{
    \textbf{Example predicted clusters of \ourfood.}
    }
    \lblfig{clusters_food}
\end{figure*}

\begin{figure*}[!th]
    \centering
    \includegraphics[width=\textwidth]{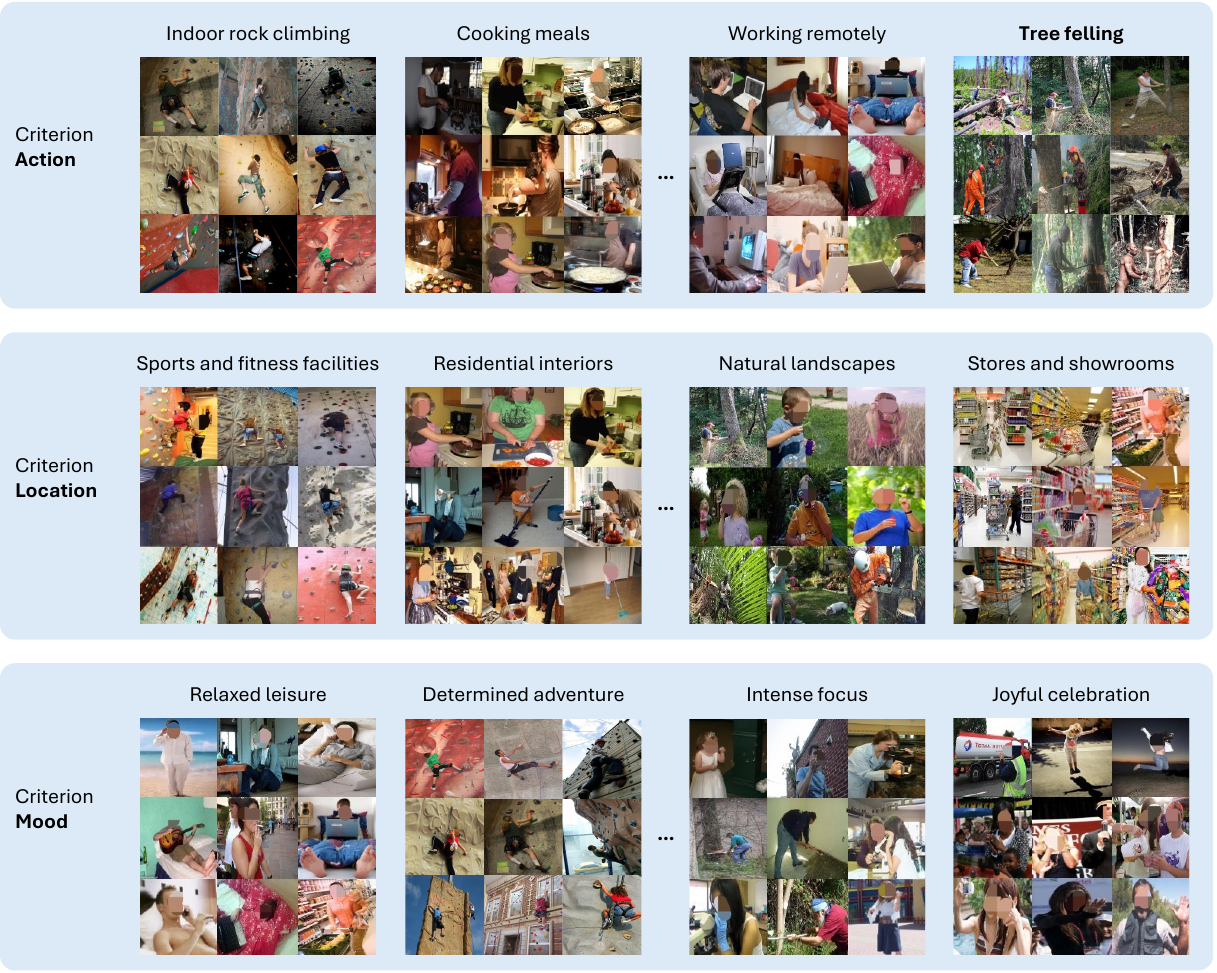}
    \caption{
    \textbf{Example predicted clusters of \action.}
    }
    \lblfig{clusters_action}
\end{figure*}

\begin{figure*}[!th]
    \centering
    \includegraphics[width=\textwidth]{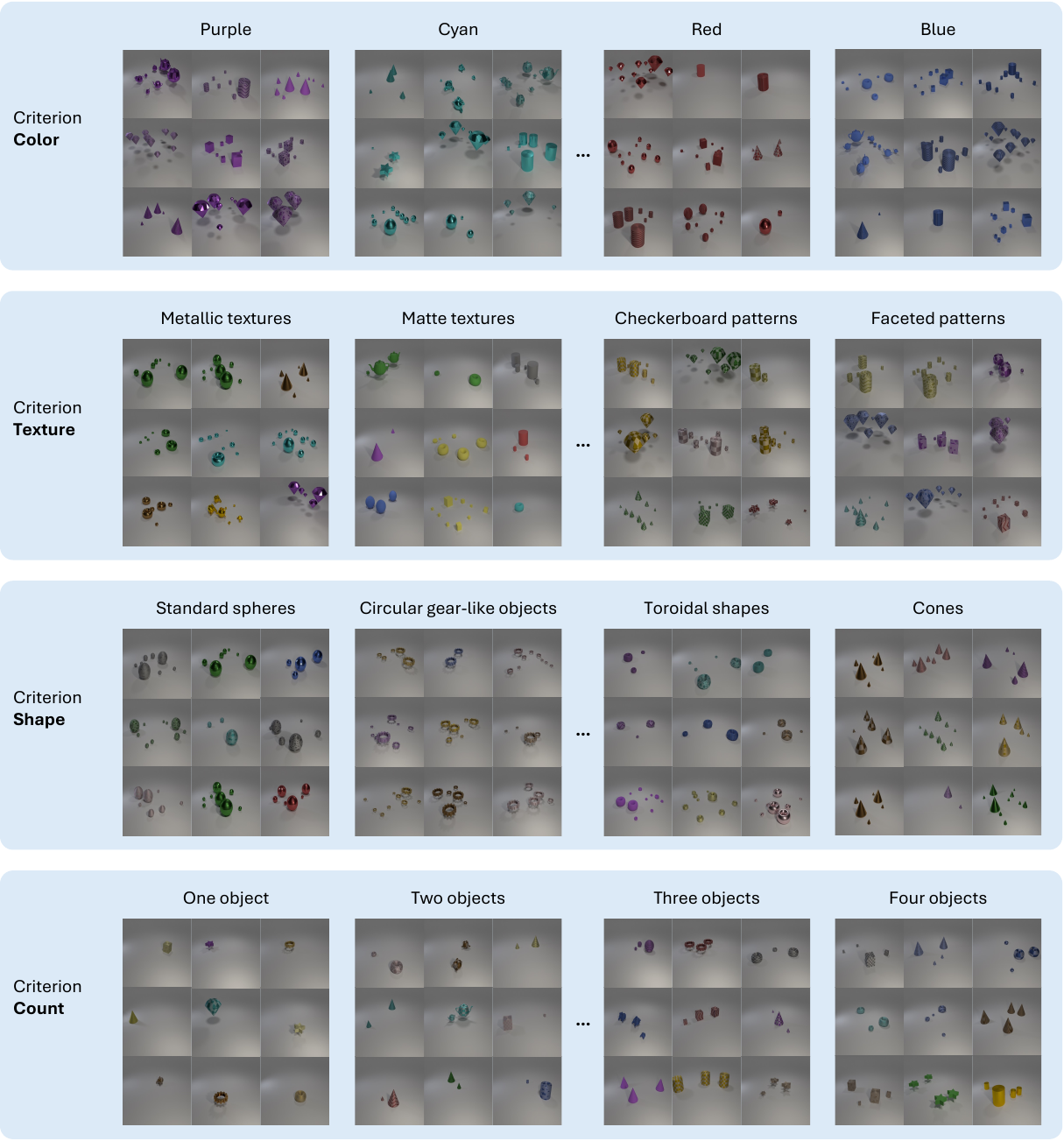}
    \caption{
    \textbf{Example predicted clusters of \clevr.}
    }
    \lblfig{clusters_clevr}
\end{figure*}

\begin{figure*}[!th]
    \centering
    \includegraphics[width=\textwidth]{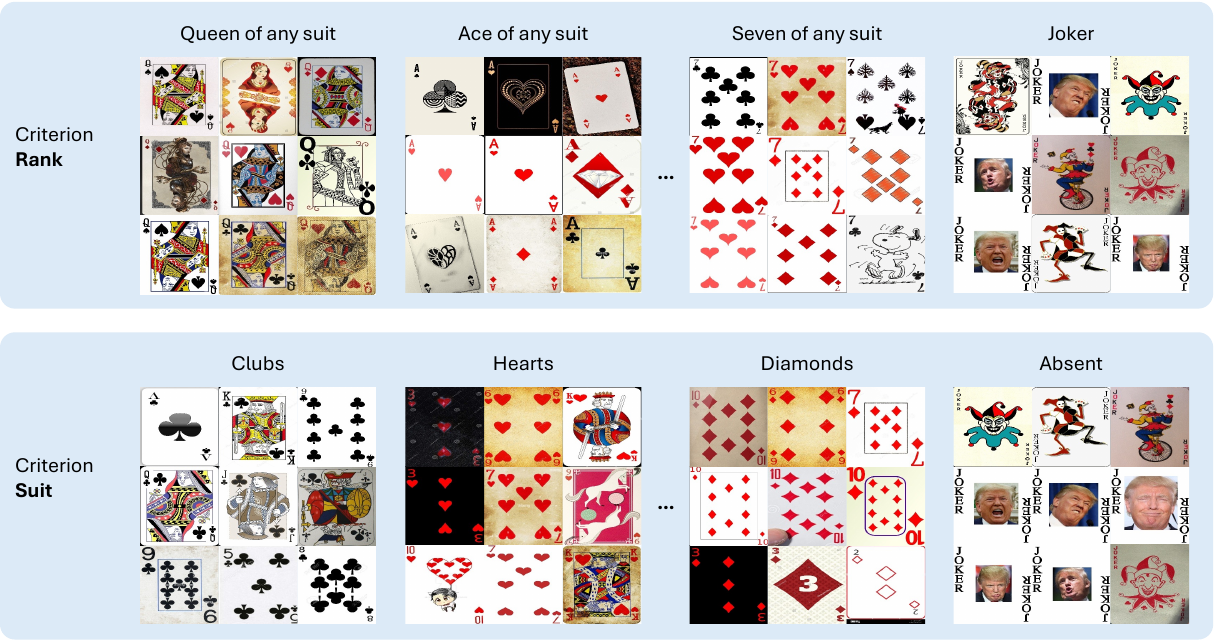}
    \caption{
    \textbf{Example predicted clusters of \card.}
    }
    \lblfig{clusters_card}
\end{figure*}

\begin{figure*}[!th]
    \centering
    \includegraphics[width=\textwidth]{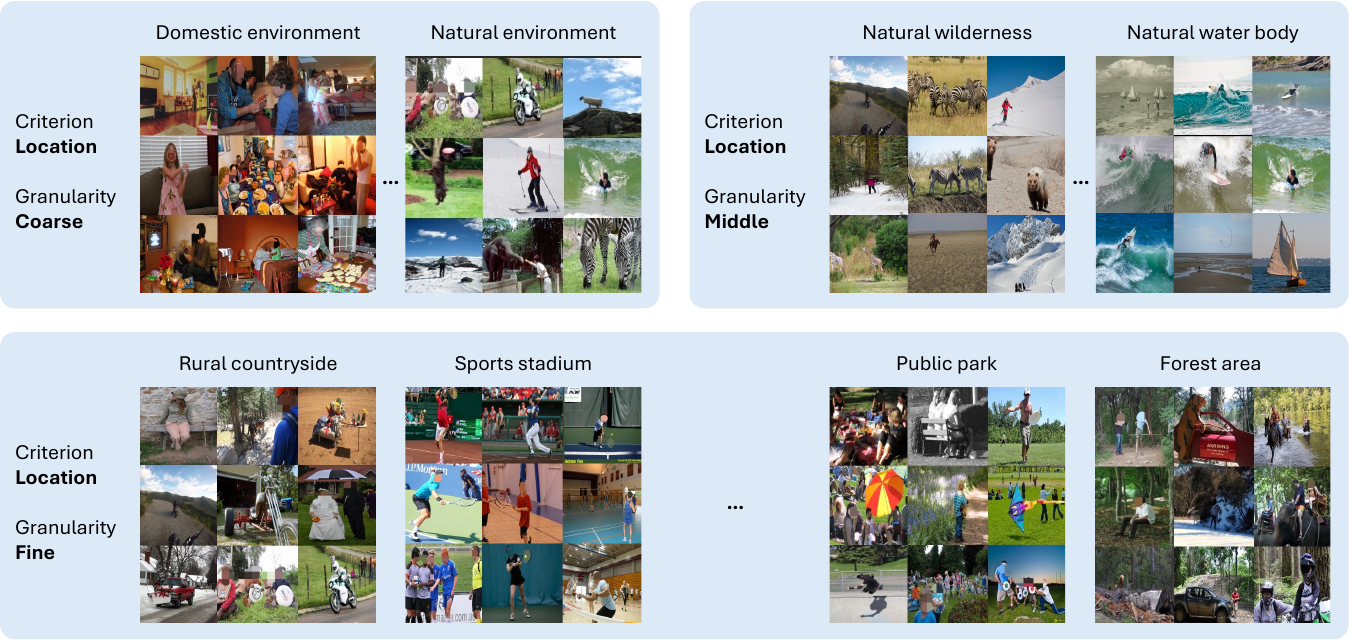}
    \caption{
    \textbf{Example predicted clusters of \ourcoco at different granularities.}
    }
    \lblfig{clusters_coco_levels}
\end{figure*}

\section{Failure Case Analysis}
\lblsec{app_failure}
In \reffig{clusters_failure}, we present several failure cases from the best configuration of our proposed framework (Caption-based Proposer and Caption-based Grouper). As observed, our method frequently mis-assigns ``Surfing'' to the ``Kayaking'' cluster under the \texttt{Activity} criterion. Upon examining the intermediate criterion captions generated by the \mllm{}, we found that this error is largely due to the \mllm{} incorrectly describing a ``Surfboard'' as a ``Kayak''. This highlights the importance of the \mllm{}'s ability to accurately describe images, as it is critical for the performance of our system. Potential improvements could include majority voting or model ensembling using different \mllm{} models.

Another issue arises in crowded scenes. When multiple people are present in an image, the model consistently assigns the \texttt{Mood} label ``Communal'' to the images. We speculate that this occurs because, in the presence of multiple people, the model struggles to accurately determine the mood of one key individual. 

Finally, we observed that our method sometimes fails to distinguish subtle, fine-grained differences between images, leading to incorrect labels. For example, as shown in \reffig{clusters_failure}, ``Edamame'' or ``Pho'' are typical dishes from China, Vietnam, and Japan, but they may be presented differently depending on the cuisine. The ``Edamame'' shown in \reffig{clusters_failure} is presented in a traditional Japanese style, yet our model incorrectly predicted it as Chinese cuisine. This oversight of fine-grained details could be improved by employing a more advanced prompting strategy~\citep{liu2024democratizing}.

\begin{figure*}[!th]
    \centering
    \includegraphics[width=\textwidth]{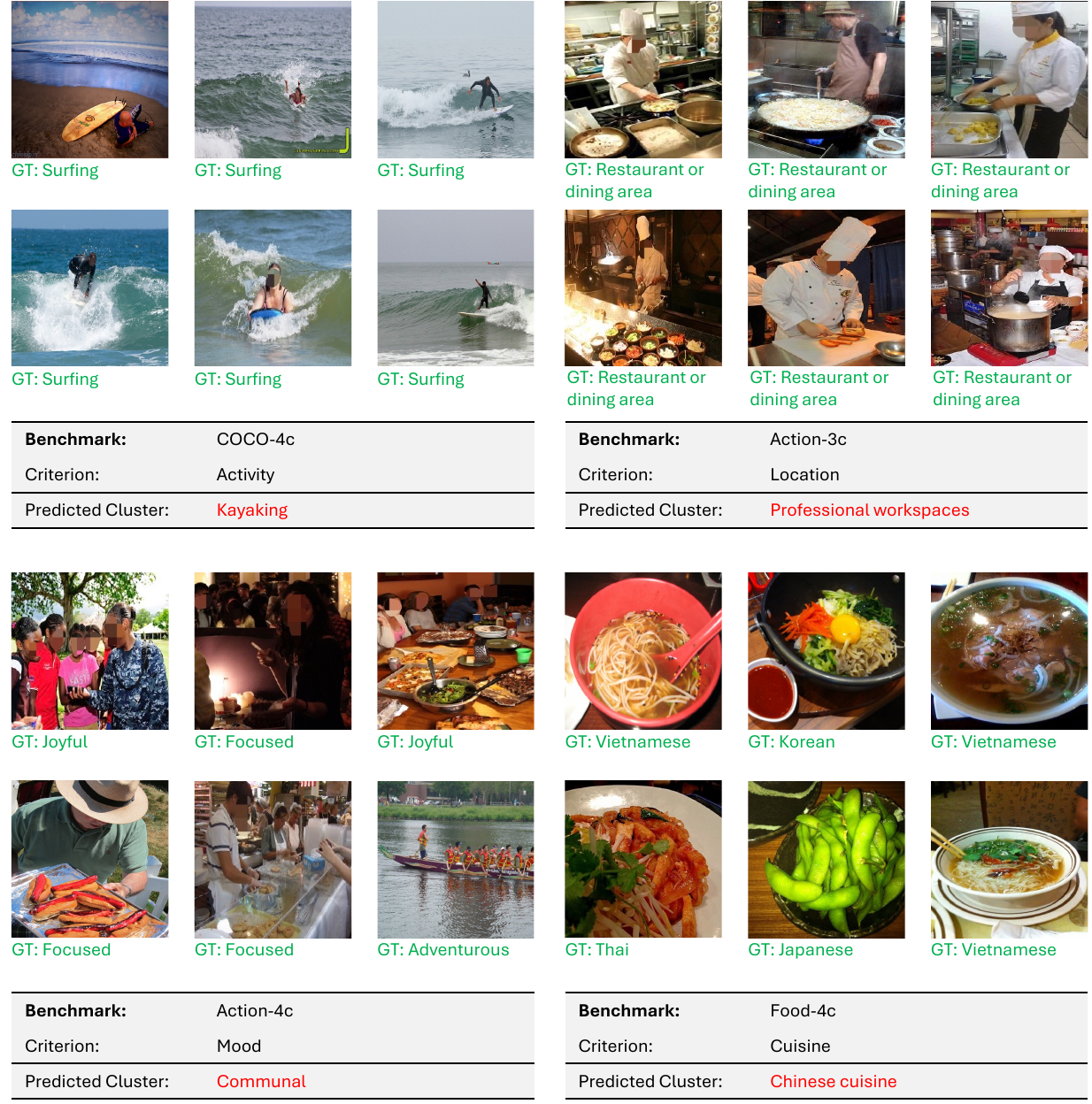}
    \caption{
    \textbf{Failure case analysis.} We show wrongly predicted images with their ground-truth label for four clusters.
    }
    \lblfig{clusters_failure}
\end{figure*}

\section{Further Study on Multi-granularity Clustering}
\lblsec{app_multi_granularity_clustering}
In this section, we provide a detailed study on how different levels of multi-granularity output from our \methodshort framework impact grouping results. Specifically, for the \action dataset, we employed human annotators to label two additional granularity levels for the criteria \texttt{Action} and \texttt{Location}. For the \texttt{Action} criterion, we consider the original annotation as fine-grained (L3) and tasked annotators to name the action in the image using more abstract and general coarse-grained (L1) and middle-grained (L2) labels. For the \texttt{Location} criterion, we consider the original annotation as middle-grained (L2) and tasked annotators to provide both more abstract coarse-grained (L1) labels and more specific fine-grained (L3) labels. This process resulted in expanded ground-truth annotations at three distinct semantic granularity levels for both the \texttt{Action} and \texttt{Location} criteria of the \action dataset.

\begin{figure}[!t]
\centering
\lesspace
    \includegraphics[width=\linewidth]{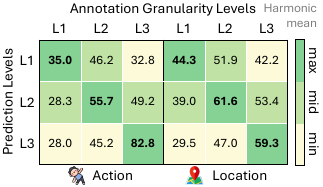}
    \caption{
    {
    \textbf{Further study on the influence of multi-granularity clustering output.}  
    We evaluate the \cacc and \sacc of the multi-granularity grouping results at each predicted clustering granularity level against each ground-truth annotation granularity level for the \texttt{Action} and \texttt{Location} criteria of the \action dataset. The Harmonic Mean of \cacc and \sacc is reported for each granularity pair. L1, L2, and L3 represent the coarse-grained, middle-grained, and fine-grained levels, respectively, for both predictions and annotations.
    }
    }
    \lblfig{further_study_granularity}
\lesspace
\end{figure}

Next, we quantitatively evaluated the multi-granularity grouping results at each predicted clustering granularity level against each ground-truth annotation granularity level by measuring clustering accuracy (CAcc) and semantic accuracy (SAcc). The main caption-based \methodshort framework was used for this experiment.  In \reffig{further_study_granularity}, we report the Harmonic Mean of CAcc and SAcc for the \texttt{Action} and \texttt{Location} criteria of \action, across each predicted clustering granularity level evaluated against each ground-truth annotation level. As clearly shown, the highest grouping performance consistently appears along the diagonal. This indicates that the best grouping performance is achieved when the predicted granularity \textit{matches} the annotation granularity.

These experimental results not only highlight the importance of the multi-granularity output of our framework but also validate the effectiveness of our multi-granularity design in aligning with user-preferred granularities that is reflected by the annotations in these experiments.

\section{Study on Handling Invalid Criteria}
\lblsec{app_discussion_fps}

At the criteria refinement step, \textit{invalid} grouping criteria (False Positives) may be proposed due to hallucinations from large language models (LLMs). While we did not observe hallucinated criteria being introduced during our experiments across six datasets and three application studies, it is important to further investigate the potential impact of such invalid criteria on the proposed \methodshort system.

To this end, we design and conduct a control experiment using the Fruit-2c dataset~\citep{muresan2018fruit}, where we \textit{artificially} introduced two ``hallucinated'' invalid grouping criteria (False Positives), \texttt{Action} and \texttt{Clothing Style}, into the refined criteria pool. These invalid criteria were then used in the subsequent grouping process to evaluate their effect on our system. We apply the main Caption-based Grouper to group fruit images based on these ``hallucinated'' criteria.

The grouping results for the two invalid criteria are presented in \reftab{false_positive_study}. As observed, when processing invalid ``hallucinated'' criteria, nearly all images are assigned to a cluster named ``Not visible'' by our framework. This occurs because, in the absence of relevant visual content in the images, the MLLM-generated captions do not include descriptors corresponding to the invalid criteria. Consequently, the LLM creates a ``Not visible'' cluster and assigns the images to it. Since the system provides interpretable outputs, users can easily identify and disregard such invalid groupings. This control experiment highlights the robustness of our system against hallucination in practical scenarios.

\begin{table*}[!ht]
    \centering
    \caption{
    \textbf{Study of the Influence of Invalid Grouping Criteria (False Positives) on the Fruit-2c Dataset.} We report the distribution of predicted groupings under the two ``hallucinated'' invalid grouping criteria. The main Caption-based Semantic Grouper is used for this experiment. \dag: For simplicity, all other minority clusters are grouped as ``Others''.
    }
    \lbltab{false_positive_study}
    \begin{tabular}{lrr}
    \toprule
        Predicted Clusters & Action (\%) & Clothing Style (\%) \\
     \toprule
        Not visible & 98.3 & 96.7 \\
        Others\dag & 1.7 & 3.3 \\
    \bottomrule
    \end{tabular}
\end{table*}

\section{Study on Model Hallucination and Bias}
\lblsec{app_limitations}
\myparagraph{Model hallucination.}  
LLM hallucination~\citep{wang2024factuality} typically occurs when LLMs are tasked with complex queries requiring world knowledge or factual information—for instance, answering a question like "Who was the 70th president of the United States?" might lead to a fabricated response. However, in our system, the use of LLMs is fully grounded in the visual descriptions (tags or captions) of the images. Consequently, the LLM output is strongly constrained to analyzing these visual descriptions, significantly reducing the likelihood of hallucination. That said, LLM hallucination can still have mild effects on clustering results. For example, as discussed in the failure case analysis in \refsec{app_failure}, the LLM incorrectly grouped ``Korean bibimbap'' and ``Vietnamese rice noodles'' under ``Chinese cuisine'' (see \reffig{clusters_failure}). MLLMs also play a crucial role in our system, as they are responsible for translating images into text for subsequent processing steps. MLLM hallucination~\citep{wang2024factuality} typically involves incorrectly identifying the existence of objects, attributes, or spatial relationships within an image. However, since our proposed system operates at the \textit{dataset level} rather than on a per-image basis, it is largely insensitive to such hallucinations, especially at the fine-grained visual detail level. Moreover, as our system is training-free, it can be further enhanced with LLM or MLLM hallucination mitigation techniques, such as the Visual Fact Checker~\citep{ge2024visual}, which we leave as a direction for future work.

\myparagraph{Model Bias.}  
Foundation models such as LLMs and MLLMs are known to inherit biases from their training data~\citep{bommasani2021opportunities}. In our system, we addressed potential biases using Hard Positive Prompting techniques: \textit{i) MLLM Bias Mitigation:} The MLLM is further prompted to generate criterion-specific captions that focus solely on describing the criterion-related content in each image. This approach constrains the MLLM from generating irrelevant content influenced by inherent biases; \textit{ii) LLM Bias Mitigation:} Similarly, when prompting the LLM to assign image captions to clusters, we condition it to concentrate exclusively on the Criterion depicted in each image (see \reftab{prompt_grouper_cap_llm_initial}).

To validate the effectiveness of these bias mitigation techniques, we conducted a fair clustering experiment. Specifically, following \citet{kwon2023image}, we sampled images for four occupations (Craftsman, Laborer, Dancer, and Gardener) from the FACET~\citep{gustafson2023facet} dataset, which contains images from 52 occupations. For each occupation, we selected 10 images of men and 10 images of women, totaling 80 images, ensuring a ground-truth gender proportion disparity of 0\% for each occupation. Using our main \methodshort system, we grouped these images based on the criterion \texttt{Occupation} using three bias mitigation strategies: \textit{i) No mitigation:} using general descriptions from the MLLM for LLM grouping; \textit{ii) Our default hard positive prompting strategy:} using criterion-specific captions from the MLLM for LLM grouping; and \textit{iii) Our default strategy with additional negative prompt:} adding a simple negative prompt, ``Do not consider gender,'' to both the MLLM captioning and LLM grouping prompts.

In this experiment, non-biased result is defined as achieving equal gender proportions within each cluster. \reftab{fair_clustering} presents the average gender ratios of the clustering results for each method across the four occupations. As observed, without bias mitigation, \methodshort exhibits noticeable gender bias in the studied occupations, with a gender disparity of 19.4\%. However, our default bias mitigation techniques effectively reduce this disparity to 4.9\%, achieving performance comparable to the addition of a manual negative prompt. This experiment demonstrates the effectiveness of our bias mitigation strategy and highlights the potential for further reducing model bias in our framework using more advanced techniques.

\begin{table*}[!ht]
    \centering
    \caption{
    \textbf{Average gender ratio and disparity} across the four studied occupations (Craftsman, Laborer, Dancer, and Gardener) from the FACET dataset. Images sampled from each occupation have an equal proportion of genders. Results from different bias mitigation strategies are reported.
    }
    \lbltab{fair_clustering}
    \begin{tabular}{lrrr}
    \toprule
        Bias Mitigation Strategy & Male (\%) & Female (\%) & Gender Disparity (\%) \\
        \toprule
        \rowcolor{verygrey}  Ground-truth & 50.0 & 50.0 & 0.0 \\
        \hline
         No mitigation & 40.3 & 59.7 & 19.4 \\
         Ours (default) & 47.6 & 52.5 & 4.9 \\
         Ours w. Negative prompt & 48.4 & 51.6 & 3.2 \\
    \bottomrule
    \end{tabular}
\end{table*}

\section{Computational Cost Analysis}
\lblsec{app_computational_cost}
The proposed \methodshort framework is training-free, requiring only inference processes. Specifically, our main framework (Caption-based) requires up to 31 GB of GPU memory to operate. All experiments reported in the paper were conducted on 4 Nvidia A100 40GB GPUs. In \reftab{computational_cost_analysis}, we provide a detailed analysis of the computational efficiency of our main \methodshort framework (Caption-based Proposer and Caption-based Grouper) on the COCO-4c benchmark (5,000 images with four criteria) across various hardware configurations. For these experiments, we used LLaVA-NeXT-7B~\citep{liu2024llava} as the MLLM and Llama-3.1-8B~\citep{meta2024llama31} as the LLM.

As shown in \reftab{computational_cost_analysis}, organizing 5,000 images based on all four discovered criteria can be completed by \methodshort in 29.1 hours on a single A100 GPU or 16.7 hours on a single H100 GPU. More importantly, most steps in our framework, such as per-image captioning and per-caption cluster assignment, are parallelizable across multiple GPUs, significantly accelerating the process. Therefore, when parallelizing the framework on 4 A100 or H100 GPUs, we achieve approximately a 4$\times$ speedup, reducing computational time to 7.6 hours on 4 A100 GPUs and 4.3 hours on 4 H100 GPUs.

\begin{table*}[!ht]
    \centering
    \caption{
    \textbf{Computational cost analysis on the COCO-4c benchmark (5,000 images with four criteria).} We report the average and total time costs on various machines. The time costs were calculated for organizing all 5,000 images according to all the 4 criteria. Our main caption-based \methodshort framework is used in this experiment.
    }
    \lbltab{computational_cost_analysis}
    \begin{tabular}{ccrr}
    \toprule
        Method & Hardware & Average time cost (sec/img) $\downarrow$ & Total time cost (hrs) $\downarrow$ \\
    \toprule
        \multirow{4}{*}{\methodshort} & 1 Nvidia A100-40GB & 20.9 & 29.1\\
         & 4 Nvidia A100-40GB & 5.5 & 7.6\\
         & 1 Nvidia H100-80GB & 12.0 & 16.7 \\
         & 4 Nvidia H100-80GB & 3.1 & 4.3\\
    \bottomrule
    \end{tabular}
\end{table*}

\section{System Sensitivity Analysis of Various \mllm{s} and \llm{s}}
\lblsec{app_expt_sensitivity}

\begin{figure*}[!ht]
    \centering
    \includegraphics[width=\textwidth]{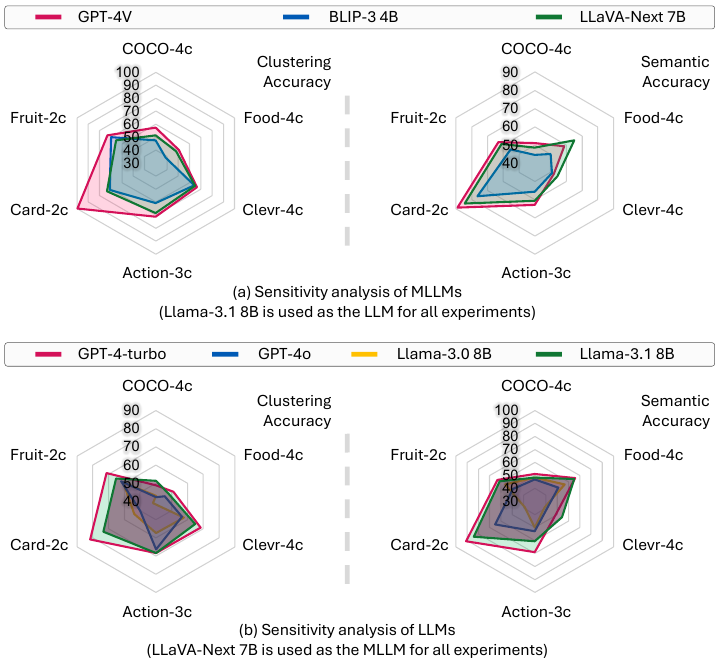}
    \caption{
    \textbf{Sensitivity analysis of different \mllm{s} and \llm{s} on the six \taskshort benchmarks.}
    \textbf{Top (a):} We fix the \llm{} to \llamathreeone{-8B} and study the impact of different \mllm{s}.
    \textbf{Bottom (b):} We fix the \mllm{} to \llava{-7B} and study the impact of different \llm{s}.
    The average \cacclong{}(\%) across different criteria is reported on the \textbf{left}, while the average \sacclong{}(\%) is reported on the \textbf{right}.
    }
    \lblfig{sensitivity_models}
\end{figure*}

In \reffig{sensitivity_models}, we perform a system-level sensitivity analysis using our default system configuration (caption-based proposer and caption-based grouper) to examine the impact of different \mllm{s} and \llm{s} on the system performance. Since all variants successfully propose the basic criteria in each benchmark, we report the average \cacclong (\cacc) and \sacclong (\sacc) across various criteria for comparative analysis.

Specifically, in \reffig{sensitivity_models}{(a)}, we first fix the \llm{} in our system to \llamathreeone{-8B}~\citep{meta2024llama31} and assess the influence of various \mllm{s}: \gptfourv~\citep{openai2023gpt4}, \blip{-3-4B}~\citep{xue2024xgen}, and \llava{-7B}~\citep{liu2024llava}. Next, in \reffig{sensitivity_models}{(b)}, we set the \mllm{} to \llava{-7B} and evaluate different \llm{s}: \gptfour{}~\citep{openai2023gpt4}, \gptfouromni{}~\citep{openai2024gpt4o}, \llamathree{-8B}~\citep{meta2024llama3}, and \llamathreeone{-8B}.

Findings in \reffig{sensitivity_models}{(a)} indicate a direct correlation between the size of the \mllm{} and the ability of our system to uncover substructures, highlighting the significant role of \mllm{s} in translating visual information into natural language. On the other hand, this scalability demonstrates that our system can enhance performance with more robust \mllm{s}, thanks to its training-free design, which ensures compatibility with any \mllm{}. Despite this, we use \llava{-7B} as our default \mllm{} due to its \textit{reproducibility}, being open-source and unaffected by API changes, and its capacity for local deployment, which \textit{upholds privacy} by not exposing sensitive image data to external entities.

As for the \llm{s}, as depicted in \reffig{sensitivity_models}(b), despite \gptfour{} showing marginally superior performance, the open-source \llamathreeone{-8B} achieves similar results across benchmarks, making it our default \llm{}. Notably, except for the \card dataset, system performance remains largely consistent regardless of the power of the \llm{}. This consistency suggests that the reasoning task for \taskshort, given the capabilities of modern \llm{s} to tackle complex problems~\citep{street2024llms}, is relatively straightforward.

\section{Study on Fine-grained Image Collections}
\lblsec{app_fine_grained}
Image collections may include fine-grained grouping criteria, such as \texttt{Bird species} in bird photography. Fine-grained criteria pose unique challenges for substructure discovery due to small inter-class differences and large intra-class variations~\citep{zhang2014part, vedaldi2014understanding, he2017fine}. This requires the model to detect subtle visual distinctions to accurately infer cluster names and guide the grouping process. The straightforward captioning process in our current framework may not fully capture these subtle visual nuances. However, the modular design of our framework allows for seamless integration of advanced cross-modal chain-of-thought (CoT) prompting strategies to address this issue.

We demonstrate this by enhancing our Caption-based Grouper with FineR~\citep{liu2024democratizing}, a cross-modal CoT prompt method specifically designed for fine-grained visual recognition. When the proposer identifies fine-grained criteria, such as \texttt{Bird species}, the framework switches to a FineR-enhanced captioning strategy that provides more detailed attribute descriptions, such as ``Wing color: Blue-grey,'' to enrich the captions and capture per-attribute visual characteristics to better support the subsequent substructure uncovering process.
\input{tables/comp_finer}

We evaluate this on two image collections containing fine-grained criteria: CUB200~\citep{wah2011caltech} and Stanford Cars196~\citep{khosla2011novel}. Our framework successfully discovers the fine-grained criteria \texttt{Bird species} for CUB200 and \texttt{Car model} for Cars196. As shown in \reftab{comp_finer}, when uncovering fine-grained substructures, integrating the FineR prompting strategy significantly improves performance by up to +15.0\% \cacc and +12.2\% \sacc, achieving results comparable to FineR itself. This demonstrates the flexibility of our system, allowing future adaptations to specific application needs, such as fine-grained image collections.

\section{Further Details of the Application Study}
\lblsec{app_application}

In this section, we present additional implementation details, evaluation results, and findings for the application study discussed in \refsec{applications} of the main paper. Specifically, \refsec{app_dataset_bias} offers further evaluation results and implementation details on using our predicted distribution to train a debiased model with GroupDRO~\citep{sagawa2019distributionally}. \refsec{app_occupation} outlines the implementation of the user study that assesses the alignment between predicted biases and human judgments, along with comprehensive findings for all studied occupations and identified criteria. Finally, \refsec{app_popularity} provides additional insights from the analysis of social media image popularity.

\subsection{Further Details on Discovering Novel Bias in Text-to-Image Diffusion Models}
\lblsec{app_occupation}
\begin{figure*}[!t]
    \centering
    \includegraphics[width=\textwidth]{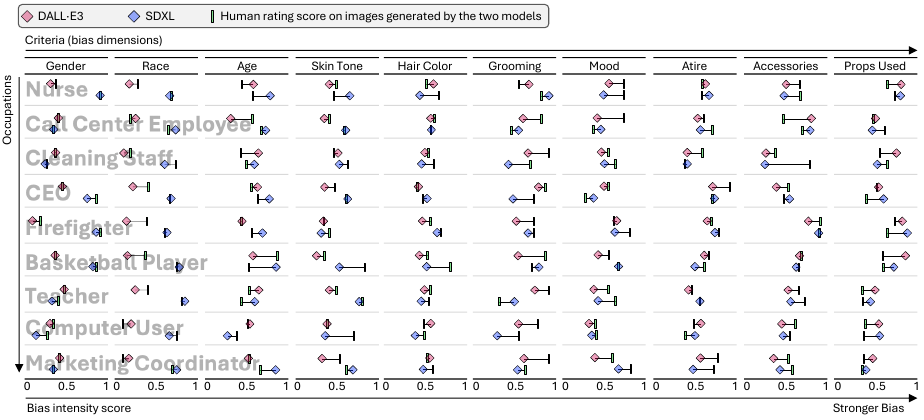}
    \caption{
        \textbf{Bias quantification results and human evaluation} for each occupation and criterion across the two studied T2I models, \dalle and SDXL. The bias intensity score is reported.
    }
    \lblfig{full_bias_evaluation}
\end{figure*}
\myparagraph{Image Generation for the Subject Occupation:}
Following prior studies~\citep{bianchi2023easily, bolukbasi2016man}, we selected nine occupations for our study: three stereotypically biased towards females (Nurse, Cleaning staff, Call center employee), three biased towards males (CEO, Firefighter, Basketball player), and three considered gender-neutral (Teacher, Computer user, Marketing coordinator). We used two state-of-the-art T2I diffusion model, \dalle~\citep{betker2023improving} and Stable Diffusion (SDXL)~\citep{podell2023sdxl} to generate 100 images for each occupation for our study. This resulted in a total of 1,800 images. For each occupation, we provide some examples of images generated by \dalle in \reffig{gallery_dalle}, while provide some examples of images generated by SDXL in \reffig{gallery_sdxl}. We only used the simple prompt ``A portrait photo of a $<$OCCUPATION$>$'' for image generation for all occupations and did not include any potential biases in the prompt. 

\myparagraph{Bias Discovery and Quantification:}
We applied our method to 1,800 generated images and automatically identified 10 grouping criteria (bias dimensions) along with their predicted distributions for each occupation image set. For this study, we utilized the mid-granularity output of our system. To evaluate the biases, we first identified the dominant cluster for each criterion—the cluster containing the largest number of images—as the \textit{bias direction}. We then calculated the normalized entropy of the distribution for each criterion of the occupation's images to determine the \textit{bias intensity} score, following the method proposed by \citet{d2024openbias}:
\begin{equation}
    \mathcal{H}_{bias}^{l} = 1 + \frac{\sum_{c^{l}\in\mathcal{C}^{l}} \log(p(c^{l}\vert\mathcal{C}^{l}, \data_{\text{Occupation}}))}{\log(\vert\mathcal{C}^{l}\vert)}
    \label{eq:bias_score}
\end{equation}
where $\data_{\text{Occupation}}$ represents the generated images for each occupation, $\mathcal{C}^{l}$ denotes the clusters discovered under the $l$-th criterion, and $p(c^{l}\vert\mathcal{C}^{l})$ is the probability of each cluster under the current distribution. The resulting score is bounded between $\mathcal{H}_{bias}^{l}\in [0, 1]$, where 0 indicates no bias towards a specific cluster (concept) under the evaluated criterion, and 1 indicates that the images are completely biased towards a particular cluster (concept) (\eg, ``Grey'' hair color) within the current bias dimension (e.g., \texttt{Hair color}). We used the score defined in \refeq{eq:bias_score} to quantify the biases for each occupation across the 10 discovered grouping criteria. We report the bias intensity score for each occupation and each model across the 10 discovered grouping criteria in \reffig{full_bias_evaluation}.

\myparagraph{Human Evaluation Study Details:}
To assess the alignment between our method's predictions and human judgments on bias detection, we conducted a user study to gather human evaluation results for the generated images. As shown in the questionnaire example in \reffig{template_survey}, participants were presented with images generated by DALL-E3 and SDXL for each occupation and were asked to identify the bias direction (dominant class) for each of the 10 discovered criteria and rate the bias intensity on a scale from 0 to 10. We collected responses from 54 anonymous participants, resulting in 6 human evaluations for each occupation and each criterion. 

The Absolute Mean Error (AME) between the bias intensity scores predicted by our system and those rated by humans (scaled to 0 to 1) was 0.1396. Additionally, our system's predicted bias directions aligned with human evaluations 72.3\% of the time, with most discrepancies occurring in the criteria of ``Age group,'' ``Skin tone,'' and ``Accessories worn.'' These findings indicate a strong correlation between our system's predictions and human judgments, validating the effectiveness of our approach. Detailed user study results are provided in \refsec{app_occupation}. We believe the discrepancies in certain criteria may be due to the influence of personal subjective cognition on respondents' answers. In \reffig{full_bias_evaluation}, we present the human evaluation results, averaged across all participants for each model, occupation, and criterion, with the human ratings scaled from 0 to 1.

\myparagraph{Complete Results and Additional Findings:}
In \reffig{full_bias_evaluation}, we present the detailed bias detection results for each model, occupation, and criterion, alongside human evaluation scores for reference. A particularly interesting phenomenon emerges: \textit{While \dalle significantly outperforms SDXL on the \textbf{well-known} bias dimensions} (\eg, \texttt{Gender}, \texttt{Race}, \texttt{Age}, and \texttt{Skin tone}), \textit{both \dalle and SDXL exhibit moderate to strong biases along the \textbf{novel} bias dimensions} (\eg, \texttt{Hair color}, \texttt{Mood}, \texttt{Attire}, and \texttt{Accessories}).

We speculate that \dalle's superior performance in mitigating well-known biases may be attributed to its ``guardrails''~\citep{dalle_guards}, designed as part of its industrial deployment to avoid amplifying social biases via its easily accessible APIs. However, these guardrails do not prevent it from exhibiting biases along the novel dimensions discovered by our method, as these dimensions remain understudied. This observation highlights the importance of studying novel biases that could potentially exist in widely used T2I generative models to prevent further bias amplification.

\subsection{Further Details on Analyzing Social Media Image Popularity}
\lblsec{app_popularity}
\begin{figure*}[!ht]
    \centering
    \includegraphics[width=\textwidth]{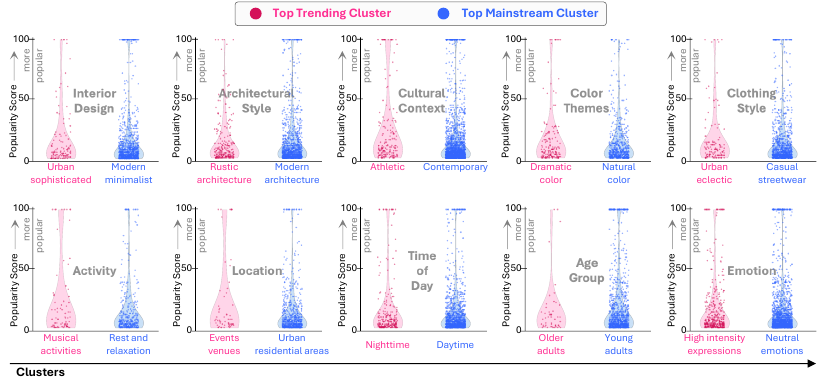}
    \caption{
        \textbf{Complete analysis of social media photo popularity on the SPID dataset.} We display the \textit{\color{trendingPink} Top Trending} and \textit{\color{trendingBlue} Top Mainstream} clusters, along with the popularity distribution of data points within these clusters across all \textit{ten} discovered criteria (in \textbf{\color{criteriagrey}Grey}).
    }
    \lblfig{app_popularity_full}
\end{figure*}

With the rise of image-centric content on social media platforms like Instagram, Flickr, and TikTok, understanding what makes an image popular has become crucial for applications such as marketing, content curation, and recommendation systems. Traditional research often approaches image popularity as a regression problem~\citep{ortis2019prediction, cheng2024retrieval}, utilizing metadata like hashtags, titles, or follower counts. However, the specific semantic visual elements that contribute to an image's popularity remain largely unexplored. In this study, we applied our proposed method to automatically categorize social media images based on semantic visual elements across different dimensions (criteria). By analyzing these interpretable results alongside image popularity metrics (\eg, number of views), we gained insights into the factors contributing to virality and identified common visual traits among popular images. These insights can provide valuable guidance for content creators and advertisers, enhancing productivity and informing strategic decision-making.

To expand on the discussion in \refsec{applications} of the main paper, we present the complete findings across all ten discovered criteria in \reffig{app_popularity_full}. Notably, we consistently observed a sharp semantic contrast between the visual elements in top trending images and those in the mainstream images across all ten criteria. For instance, there is a contrast between \textit{Urban sophisticated} and \textit{Modern minimalist} under \texttt{Interior Design}, \textit{Rustic architecture} and \textit{Modern architecture} under \texttt{Architecture Style}, and \textit{Event venues} versus \textit{Urban residential areas} under \texttt{Location}.

This recurring observation reinforces the idea that viral (or trending) content tends to capture more attention, likely because it features novel, surprising, or striking visual elements. Humans are inherently attracted to stimuli that deviate from the norm~\citep{priester2004and, bruni2012role, palmer2014theory}. On the other hand, widely uploaded yet ``neutral'' content is shared more often due to its familiarity and broad appeal, though it is less likely to provoke the strong emotional responses that fuel virality. We believe the insights generated by our method could offer valuable guidance to social media platform practitioners, helping them tailor their content more effectively to target audiences and gain a deeper understanding of social media image trends from various perspectives.

\subsection{Confirming and Mitigating Dataset Bias}
\lblsec{app_dataset_bias}
\myparagraph{Confirming and Mitigating Dataset Bias:} Given an image collection that contains \textit{spurious correlations}~\citep{geirhos2020shortcut}, we are curious whether we can proactively find this issue caused by data bias directly from the training images \textit{without} relying on either the annotations~\citep{sagawa2019distributionally} or \textit{post hoc} misclassified images~\citep{kim2024discovering}. As a case study, we applied the proposed \methodshort framework to the 162k training images of the CelebA~\citep{liu2015deep} dataset—a binary hair color classification dataset where the target label ``Blond'' is spuriously correlated with the demographic attribute ``Female'' in its training split. 

\begin{figure}[!t]
\centering
    \includegraphics[width=0.8\linewidth]{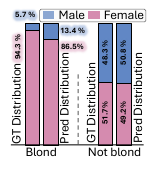}
    \caption{
    \textbf{Results of dataset bias discovery and mitigation.} Worst group and average accuracies(\%) are reported.
    }
    \lblfig{app_celebA}
\end{figure}

\input{tables/app_celebA}

\myparagraph{Findings:} As expected, our method identified the grouping criteria \texttt{Hair color} and \texttt{Gender}.
Next, we analyzed the predicted gender distributions within the ``Blond'' and ``Not Blond'' (all other colors) clusters. As shown in \reffig{app_celebA}, we observed that the gender distribution within the ``Blond'' cluster is highly skewed, with 86.5\% of the images representing females, closely matching the ground-truth distribution (94.3\%).
Such an imbalance confirms the potential issue of spurious correlations between ``Blond'' and ``Female''.
To further validate this observation, following B2T~\citep{kim2024discovering}, we used our predicted distributions to train a debiased model with GroupDRO~\citep{sagawa2019distributionally} and compared it with other unsupervised bias mitigation methods, including JTT~\citep{liu2021just}, CNC~\citep{zhang2022correct}, B2T, and GroupDRO trained with ground-truth labels.
As shown in \reftab{app_celebA_mitigation}, our debiased model achieved robust performance, comparable to that of B2T, demonstrating the reliability of its discovered distributions.

\myparagraph{Additional Evaluation:}
To further evaluate the prediction quality of our method for hair color and gender, we used the ground-truth labels from the CelebA dataset~\citep{liu2015deep} to assess the classification accuracy of them. Our method achieved an impressive classification accuracy of 99.1\% for gender and 87.4\% for hair color on the 162,770 training images, demonstrating its effectiveness for uncovering gender and hair color substructures within the training set.

In addition, we quantified the \textit{spurious correlation} between hair color and gender using the metric proposed by \citet{yang2023change}. Specifically, given the correlated gender attribute distribution $A$ and the target hair color distribution $Y$, we computed the normalized mutual information between $A$ and $Y$ to quantify the spurious correlation as:
\begin{equation}
    I(A;Y) = \frac{2I(A;Y)}{H(A) + H(Y)}
    \label{eq:correlation}
\end{equation}
where $H(A)$ and $H(Y)$ represent the normalized entropy of the gender and hair color distributions, respectively. A value of $H(A)$ or $H(Y)$ equal to 1 indicates a uniform distribution (\ie, no class imbalance). We then used the ground-truth distribution from the dataset's labels and our predicted distribution to estimate the spurious correlation intensity using the score from \refeq{eq:correlation}. For gender and hair color, our method's predictions yielded a score of $I_{Pred} = 0.10$, which is nearly identical to the ground-truth score of $I_{GT} = 0.11$. This demonstrates that our method effectively identifies and confirms the bias directly from the training set.

\myparagraph{Implementation Details of Training GroupDRO:}
To conduct debiased training using GroupDRO~\citep{sagawa2019distributionally}, we first used our predicted distribution to define four distinct training groups, rather than relying on the ground-truth distribution. We closely followed the training protocol outlined in B2T~\citep{kim2024discovering} and GroupDRO~\citep{sagawa2019distributionally}. Specifically, we fine-tuned a ResNet-50~\citep{he2016deep} model pre-trained on ImageNet~\citep{deng2009imagenet}, using the training split of the CelebA dataset~\citep{liu2015deep}. The training was performed using the SGD optimizer~\citep{ruder2016overview} with a momentum of 0.9, a batch size of 64, and a learning rate of $1 \times 10^{-5}$. We applied a weight decay of 0.1 and set the group adjustment parameter to zero. The model was trained over 50 epochs. For evaluation, we reported both the average and worst-group test accuracies, selecting the model from the epoch that achieves the highest worst-group accuracy on the validation set. The final evaluation and comparison results are provided in \reftab{app_celebA_mitigation}.

\section{{Why LLMs Improve Image Clustering?}}
\lblsec{app_discussion_system_design}

The most compelling aspect of this work lies in our \methodshort framework's ability to transform large volumes of unstructured images into natural language and leverage the advanced text understanding and summarization capabilities of LLMs to tackle the challenging \tasktitle (\taskshort) task. This approach draws inspiration from the use of LLMs in the Topic Discovery task within the NLP domain~\citep{eklund2022topic}. Our core motivation is: ``If LLMs can discover topics from documents and organize them, then by converting images into text, we can similarly use LLMs to organize unstructured images.''

Traditional clustering methods~\citep{estivill2002so, caron2018deep, van2020scan, li2023image, yu2024multiple} often depend on pre-defined criteria, pre-determined numbers of clusters, fixed feature representations (which require training), and are typically not interpretable. These limitations hinder their applicability to diverse datasets in open-world scenarios, as they demand significant human priors and retraining for each new dataset.

In contrast, LLMs~\citep{chatgpt, openai2023gpt4, touvron2023llama, meta2024llama3, meta2024llama31} excel at understanding, summarizing, and reasoning over high-level semantics expressed in natural language across diverse domains (\eg, everyday content, cultural knowledge, or medical content). Operating in a zero-shot~\citep{kojima2022large}, interpretable manner, LLMs are uniquely suited to the SMC task, which aims to discover meaningful and interpretable clustering criteria without requiring prior knowledge or training. By integrating LLMs with MLLMs~\citep{liu2024llava} into the carefully designed \methodshort framework, we enable the discovery and refinement of clustering criteria directly from the dataset's content, followed by automatic grouping of the dataset. This design allows our framework to overcome the rigid assumptions of traditional clustering methods, making it automatic, generalizable, and training-free. Our approach provides a novel perspective, demonstrating how clustering tasks can evolve beyond traditional paradigms.

\myparagraph{Challenges of employing LLMs to facilitate the SMC task.}
The main challenge of employing LLMs for the SMC task lies in accurately translating visual content from images into natural language that \llm{s} can effectively reason with. This is evident from the sensitivity analysis results in \refapp{app_expt_sensitivity}: \methodshort's performance improves with larger or more powerful \mllm{s} (see \reffig{sensitivity_models} (a)), while it remains relatively insensitive to the specific choice of \llm{} (see \reffig{sensitivity_models} (b)). In other words, the quality of image captions generated by \mllm{s} is critical for the effective use of \llm{s} in the SMC task. Specifically, in the first stage of \methodshort (criteria proposal), captions need to be as comprehensive as possible to provide \textit{rich} information for \llm{s} to discover grouping criteria. In the second stage (semantic grouping), criterion-specific captions should precisely capture relevant visual content to provide \textit{accurate} information for assigning images to clusters.

To enhance caption quality, techniques such as \mllm{} model ensembling, prompt ensembling~\citep{liu2024visual}, or stronger models like GPT-4V~\citep{openai2023gpt4} can improve comprehensiveness. For better precision, advanced prompting methods like CoT~\citep{wei2022chain} or FineR~\citep{liu2024democratizing} can capture nuanced details, while hallucination mitigation tools like Visual Fact Checker~\citep{ge2024visual} can reduce noise caused by hallucinations. However, these techniques increase computational costs and framework complexity. In this work, we choose to keep \methodshort simple yet effective, and we outline these potential improvements for future practitioners.

\section{{Future Work}}
\lblsec{app_future_work}

\myparagraph{Closed-Loop Optimization.}  
In this work, we designed our prompts following the Iterative Prompt Engineering methodology~\citep{deeplearningai_2024} introduced by Isa Fulford and Andrew Ng. In \refapp{app_prompt}, we provide the exact LLM and MLLM prompts used in our framework and break down each prompt to explain the objectives and purposes behind each design choice. These explanations cover elements such as system prompts, input formatting, task and sub-task instructions, and output instructions. Our focus in this work is on creating a highly generalizable framework, \methodshort, and we do not perform any closed-loop, dataset-specific prompt optimizations. However, in future work or application scenarios where a labeled training/validation dataset is available, practitioners could build upon our design objectives. By leveraging our proposed evaluation metrics (see \refsec{method_assessment}) for each step, it would be possible to develop a \tasktitle (\taskshort) system with a closed-loop optimization pipeline to achieve improved performance.

\myparagraph{\methodshortbold on Other Data Types.}  
The core idea of our proposed framework, \methodshort, is to \textit{use text as a proxy (or medium)} for reasoning over large volumes of unstructured data, generating human-interpretable insights at scale. As such, \methodshort can be directly applied to textual data (\eg, documents). Moreover, since natural language is a highly versatile and widely-used medium of representation, \methodshort can be extended to other data types by converting these data into text (by replacing the captioning module with suitable tools) in future work, such as:
 
\begin{itemize}
    \item \textbf{Audio Data:} Speech-to-Text models like Whisper~\citep{radford2023robust} can convert audio data into text, enabling subsequent analysis with \methodshort.
    \item \textbf{Tabular Data:} Table-to-Text models, such as TabT5~\citep{andrejczuk2022table}, can translate tabular data into text, making it compatible with \methodshort. For tables containing figures, modern MLLMs like LLaVA-Next, which support both OCR and image-to-text capabilities, can handle these elements to create a unified textual representation for \methodshort.
    \item \textbf{Protein Structures:} Protein structure-to-text models, such as ProtChatGPT~\citep{wang2024protchatgpt}, can convert protein sequences into textual descriptions for analysis with \methodshort.
    \item \textbf{Point Cloud Data:} 3D captioning models, like Cap3D~\citep{luo2024scalable}, can transform point cloud data or rendered 3D models into text, enabling their analysis using \methodshort.
\end{itemize}

We believe the versatile nature of \methodshort has the potential to open up a broad range of applications across diverse data modalities, fostering new directions in future research.

\begin{figure*}[!tt]
    \centering
    \includegraphics[width=\textwidth]{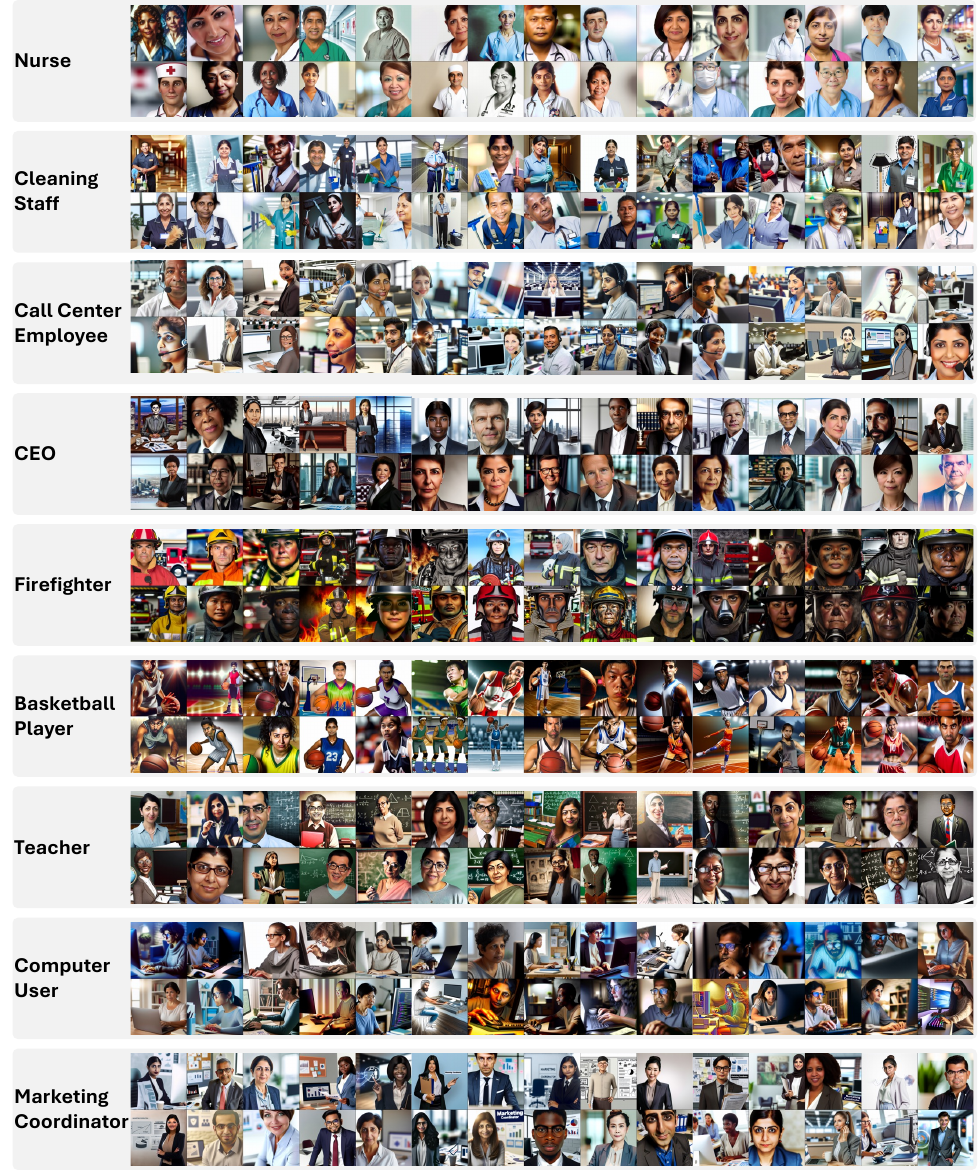}
    \caption{
    \textbf{Samples of \dalle generated images.}
    For each occupation, the simple prompt ``A portrait photo of a $<$OCCUPATION$>$'', that does not reference any potential bias dimensions such as gender, race or hair color, is fed to \dalle to generate 100 images. We present a random sample of 30 generated images.
    }
    \lblfig{gallery_dalle}
\end{figure*}

\begin{figure*}[!t]
    \centering
    \includegraphics[width=\textwidth]{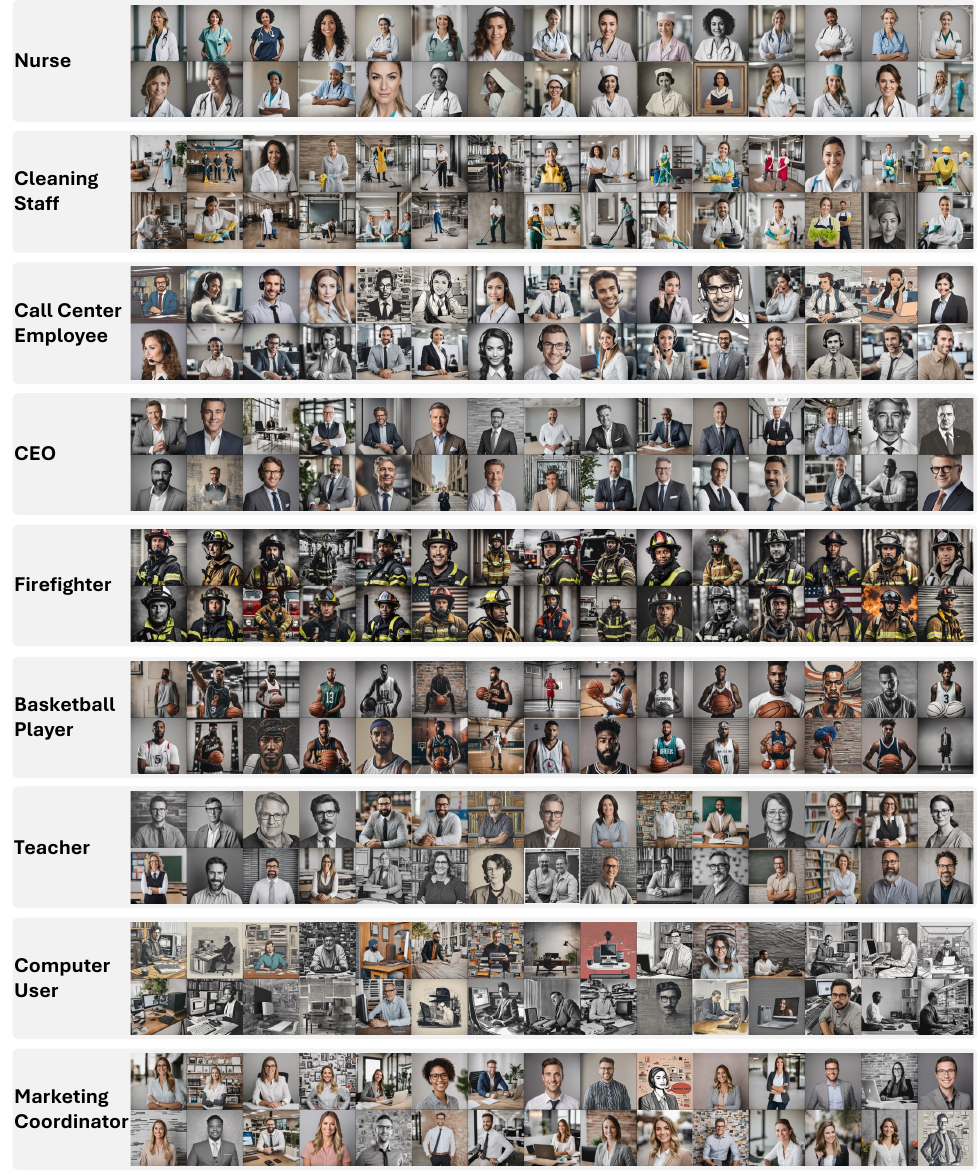}
    \caption{
    \textbf{Samples of SDXL generated images.}
    For each occupation, the simple prompt ``A portrait photo of a $<$OCCUPATION$>$'', that does not reference any potential bias dimensions such as gender, race or hair color, is fed to SDXL to generate 100 images. We present a random sample of 30 generated images.
    }
    \lblfig{gallery_sdxl}
\end{figure*}

\begin{figure*}[!t]
    \centering
    \includegraphics[height=\textheight]{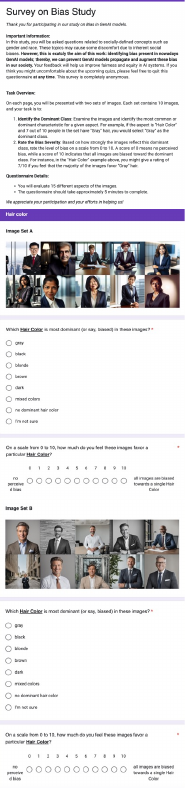}
    \caption{
    \textbf{Example of the questionnaire for human evaluation study.}
    }
    \lblfig{template_survey}
\end{figure*}

%% file: tables/benchmarks/benchmark_summary_basic.tex
\begin{table*}[!ht]
\centering

\caption{
\textbf{Summary of number of classes for the basic criteria annotation across the six benchmarks.}
}
\lbltab{benchmark_summary_basic}

\begin{tabular}{p{0.20\linewidth}p{0.20\linewidth}p{0.20\linewidth}p{0.20\linewidth}}

\toprule

\textbf{Dataset}
& {\textbf{Number of Images}}
& \textbf{Basic Criterion}
& \textbf{Number of Classes}
\\

\toprule

\multirow{4}{*}{\ourcoco}
& \multirow{4}{*}{{5,000}}
& Activity
& 64
\\
& 
& Location
& 19
\\
& 
& Mood
& 20
\\
& 
& Time of Day
& 6
\\

\hline

\multirow{4}{*}{\ourfood}
& \multirow{4}{*}{{25,250}}
& Food Type
& 101
\\
& 
& Cuisine
& 15
\\
& 
& Course
& 5
\\
& 
& Diet
& 4
\\

\hline

\multirow{3}{*}{\action}
& \multirow{3}{*}{{1,000}}
& Action
& 40
\\
& 
& Location
& 10
\\
& 
& Mood
& 4
\\

\hline

\multirow{4}{*}{\clevr}
& \multirow{4}{*}{{10,000}}
& Color
& 10
\\
& 
& Texture
& 10
\\
& 
& Shape
& 10
\\
& 
& Count
& 10
\\

\hline

\multirow{2}{*}{\card}
& \multirow{2}{*}{{8,029}}
& Rank
& 14
\\
& 
& Suit
& 5
\\

\hline

\multirow{2}{*}{\fruit}
& \multirow{2}{*}{{103}}
& Species
& 34
\\
& 
& Color
& 15
\\
\bottomrule
\end{tabular}
\end{table*}

%% file: tables/benchmarks/benchmark_names_coco4c.tex
\begin{table*}[!ht]
\centering
\tiny

\caption{
\textbf{Full class names for \ourcoco across the four basic criteria.}
}
\lbltab{benchmark_names_coco4c}

\renewcommand\tabcolsep{10pt}
\renewcommand\arraystretch{1.3}
\footnotesize 

\begin{tabular}{p{0.15\linewidth}|p{0.71\linewidth}} 
\toprule

\textbf{Criterion} & \textbf{\ourcoco} \\

\toprule

Activity 
&

``repairing a toilet'',
``playing volleyball'',
``playing guitar'',
``haircutting'',
``cutting a cigar'',
``kayaking'',
``applauding'',
``tying a tie'',
``playing basketball'',
``washing dishes'',
``gardening'',
``texting messages'',
``repairing a car'',
``peeing'',
``cleaning the floor'',
``writing on a book'',
``feeding a horse'',
``singing'',
``baking'',
``hiking'',
``smoking'',
``riding an elephant'',
``pouring liquid'',
``waving hands'',
``swimming'',
``meditating'',
``fixing a bike'',
``cutting vegetables'',
``walking a dog'',
``reading a book'',
``celebrating'',
``queuing'',
``cutting a cake'',
``brushing teeth'',
``playing soccer'',
``jumping'',
``snowboarding'',
``playing'',
``touching animals'',
``pushing a cart'',
``watching tv'',
``rowing a boat'',
``taking photos'',
``running'',
``flying a kite'',
``riding a horse'',
``playing video games'',
``holding up an umbrella'',
``throwing a frisbee'',
``lying down'',
``riding a bike'',
``drinking'',
``cooking'',
``phoning'',
``chatting'',
``skiing'',
``driving'',
``surfing'',
``skateboarding'',
``playing baseball'',
``playing tennis'',
``using a computer'',
``posing'',
``eating''
\\
\hline

Location 
&
``amusement or theme park'',
``healthcare facility'',
``virtual or digital space'',
``educational institution'',
``industrial area'',
``historical landmark'',
``public event or gathering'',
``store or market'',
``underground or enclosed space'',
``transportation hub'',
``zoo'',
``water body'',
``office or workplace'',
``park or recreational area'',
``restaurant or dining area'',
``sports facility'',
``natural environment'',
``urban area or city street'',
``residential area''
\\
\hline

Mood 
&
``anxious'',
``sombre'',
``contemplative'',
``suspenseful'',
``serene'',
``nostalgic'',
``inspired'',
``whimsical'',
``romantic'',
``mysterious'',
``melancholic'',
``chaotic'',
``humorous'',
``vibrant'',
``peaceful'',
``energetic'',
``focused'',
``joyful'',
``relaxed'',
``adventurous''
\\
\hline

Time of Day 
&
``evening'',
``afternoon'',
``night'',
``morning'',
``indoor lighting'',
``midday''
\\

\bottomrule
\end{tabular}
\end{table*}

%% file: tables/benchmarks/benchmark_names_food4c.tex
\begin{table*}[!ht]
\centering
\tiny

\caption{
\textbf{Full class names for \ourfood across the four basic criteria.}
}
\lbltab{benchmark_names_food4c}

\renewcommand\tabcolsep{10pt}
\renewcommand\arraystretch{1.3}
\footnotesize 

\begin{tabular}{p{0.15\linewidth}|p{0.71\linewidth}} 
\toprule

\textbf{Criterion} & \textbf{\ourfood} \\

\toprule

Food Type 
&
``apple pie'',
``baby back ribs'',
``baklava'',
``beef carpaccio'',
``beef tartare'',
``beet salad'',
``beignets'',
``bibimbap'',
``bread pudding'',
``breakfast burrito'',
``bruschetta'',
``caesar salad'',
``cannoli'',
``caprese salad'',
``carrot cake'',
``ceviche'',
``cheesecake'',
``cheese plate'',
``chicken curry'',
``chicken quesadilla'',
``chicken wings'',
``chocolate cake'',
``chocolate mousse'',
``churros'',
``clam chowder'',
``club sandwich'',
``crab cakes'',
``creme brulee'',
``croque madame'',
``cup cakes'',
``deviled eggs'',
``donuts'',
``dumplings'',
``edamame'',
``eggs benedict'',
``escargots'',
``falafel'',
``filet mignon'',
``fish and chips'',
``foie gras'',
``french fries'',
``french onion soup'',
``french toast'',
``fried calamari'',
``fried rice'',
``frozen yogurt'',
``garlic bread'',
``gnocchi'',
``greek salad'',
``grilled cheese sandwich'',
``grilled salmon'',
``guacamole'',
``gyoza'',
``hamburger'',
``hot and sour soup'',
``hot dog'',
``huevos rancheros'',
``hummus'',
``ice cream'',
``lasagna'',
``lobster bisque'',
``lobster roll sandwich'',
``macaroni and cheese'',
``macarons'',
``miso soup'',
``mussels'',
``nachos'',
``omelette'',
``onion rings'',
``oysters'',
``pad thai'',
``paella'',
``pancakes'',
``panna cotta'',
``peking duck'',
``pho'',
``pizza'',
``pork chop'',
``poutine'',
``prime rib'',
``pulled pork sandwich'',
``ramen'',
``ravioli'',
``red velvet cake'',
``risotto'',
``samosa'',
``sashimi'',
``scallops'',
``seaweed salad'',
``shrimp and grits'',
``spaghetti bolognese'',
``spaghetti carbonara'',
``spring rolls'',
``steak'',
``strawberry shortcake'',
``sushi'',
``tacos'',
``takoyaki'',
``tiramisu'',
``tuna tartare'',
``waffles''
\\

\hline

Cuisine 
&
``japanese'',
``indian'',
``american'',
``greek'',
``spanish'',
``mexican'',
``italian'',
``vietnamese'',
``canadian'',
``korean'',
``chinese'',
``middle eastern'',
``french'',
``thai'',
``general''
\\

\hline

Course 
&
``appetizer'',  ``main course'', ``side dish'', ``dessert'', ``breakfast''
\\

\hline

Diet 
&
``omnivore'', ``vegan'', ``vegetarian'', ``gluten free''
\\

\bottomrule
\end{tabular}
\end{table*}

%% file: tables/benchmarks/benchmark_criteria.tex
\begin{table*}[!ht]
\centering
\tiny

\caption{
\textbf{Annotated criteria for the six benchmarks.}
The basic criteria are annotated on per-image level for each benchmark, while the hard criteria (those not in the basic criteria) are further exhaustively annotated by human annotators for further evaluating the performance of the rule proposer in \taskshort task.
}
\lbltab{benchmark_criteria}

\begin{tabular}{p{0.135\linewidth}p{0.135\linewidth}|p{0.135\linewidth}p{0.135\linewidth}|p{0.135\linewidth}p{0.135\linewidth}}

\toprule

\multicolumn{2}{c|}{\textbf{\ourcoco}}
& \multicolumn{2}{c|}{\textbf{\ourfood}}
& \multicolumn{2}{c}{\textbf{\action}}
\\

\textbf{Basic criteria} & \textbf{Hard criteria} 
& \textbf{Basic criteria} & \textbf{Hard criteria}
& \textbf{Basic criteria} & \textbf{Hard criteria}
\\

\bf Total: 4 & \bf Total: 17
& \bf Total: 4 & \bf Total: 11
& \bf Total: 3 & \bf Total: 11
\\

\toprule

Activity & Activity
& Food Type & Food Type 
& Action & Action 
\\

Location & Location 
& Cuisine & Cuisine 
& Mood & Mood 
\\

Mood & Mood 
& Course & Course 
& Location & Location 
\\

Time of Day & Time of Day 
& Diet & Diet 
&  & Clothing Style
\\

  & Interaction  
&  & Tableware Type 
&  & Number of People Present 
\\

  & Number of People Present  
&  & Presentation Style 
&  & Age or Age Composition 
\\

  & Group Dynamics  
&  & Color Palette 
&  & Race or Race Composition 
\\

  & Clothing Style  
&  & Setting/Theme 
&  & Occasion or Event Type 
\\

  & Occasion or Event Type  
&  & Primary Taste 
&  & Group Dynamics 
\\

  & Photo Style  
&  & Primary Ingredient 
&  & Lighting Condition 
\\

  & Type of Animal Present  
&  & Cooking Method 
&  & Gender or Gender Composition 
\\

  & Weather  
&  &  
&  &  
\\

  & Type of Primary Object  
&  &  
&  &  
\\

  & Continent  
&  &  
&  &  
\\

  & Age or Age Composition  
&  &  
&  &  
\\

  & Race or Race Composition  
&  &  
&  &  
\\

  & Gender or Gender Composition  
&  &  
&  &  
\\

\Xhline{1pt}

\multicolumn{2}{c|}{\textbf{\clevr}}
& \multicolumn{2}{c|}{\textbf{\card}}
& \multicolumn{2}{c}{\textbf{\fruit}}
\\

\textbf{Basic criteria} & \textbf{Hard criteria} 
& \textbf{Basic criteria} & \textbf{Hard criteria}
& \textbf{Basic criteria} & \textbf{Hard criteria}
\\

\bf Total: 4 & \bf Total: 7
& \bf Total: 2 & \bf Total: 4
& \bf Total: 2 & \bf Total: 8
\\

\Xhline{1pt}
Color & Color 
& Rank & Rank 
& Species & Species 
\\

Texture & Texture 
& Suit & Suit 
& Color & Color 
\\

Shape & Shape 
&  & Color 
&  & Size 
\\

Count & Count 
&  & Illustration Style 
&  & Seasonality 
\\

  & Spatial Positioning 
&  &  
&  & Primary Taste 
\\

  & Count of Surface 
&  &  
&  & Texture 
\\

  & Complexity of Geometry 
&  &  
&  & Ripeness 
\\

  &  
&  &  
&  & Fruit Quantity and Arrangement 
\\


\bottomrule
\end{tabular}
\end{table*}

%% file: tables/prompts/prompt_proposer_img_mllm.tex
\begin{table*}[t!]
  \centering
  \caption{
  \textbf{Prompts for the \mllm{} in the image-based proposer for criteria proposing.}
  }
  \lbltab{prompt_proposer_img_mllm}
  
  \renewcommand\tabcolsep{10pt}
  \renewcommand\arraystretch{1.3}
  \footnotesize 
  
  \begin{tabular}{|p{0.2\linewidth}|p{0.66\linewidth}|} 
    \Xhline{1.2pt}
    
    \rowcolor{champagne} 
    \textbf{Prompt purpose} & \textbf{Prompt} \\
    
    \Xhline{1.2pt}
    
    System Prompt 
    & You are a helpful AI assistant
    \\
    
    \hline
    Input Explanation 
    &  This image contains 64 individual images arranged in 8 columns and 8 rows.
    \\
    
    \hline
    Goal Explanation 
    &  I am a machine learning researcher trying to identify all the possible clustering criteria or rules that could be used to group these images so I can better understand my data.
    \\
    
    \hline
    Task Instruction 
    &  Your job is to carefully analyze the entire set of the 64 images, and identify five distinct clustering criteria or rules that could be used to cluster or group these images. Please consider different characteristics. 
    \\

    \hline
    Output Instruction
    & Please write a list of the 5 identified clustering criteria or rules (separated by bullet points ``*'').
    \\
    
    \hline
    Task Reinforcement
    &  Again, I want to identify all the possible clustering criteria or rules that could be used to group these images. List the 5 distinct clustering criteria or rules that you identified from the 64 images. Answer with a list (separated by bullet points ``*'').
    
    Your response:
    \\

    \Xhline{1.2pt}
  \end{tabular}
\end{table*}

%% file: tables/prompts/prompt_proposer_tag_llm.tex
\begin{table*}[t!]
  \centering
  \caption{
  \textbf{Prompts for the \llm{} used in the tag-based proposer for criteria proposing.} We embed the exact image captions by replacing the placeholders \texttt{"\{TAGS\}"} in the prompt.
  }
  \lbltab{prompt_proposer_tag_llm}
  
  \renewcommand\tabcolsep{10pt}
  \renewcommand\arraystretch{1.3}
  \footnotesize 
  
  \begin{tabular}{|p{0.2\linewidth}|p{0.66\linewidth}|} 
    \Xhline{1.2pt}
    
    \rowcolor{champagne} 
    \textbf{Prompt purpose} & \textbf{Prompt} \\
    
    \Xhline{1.2pt}
    
    System Prompt 
    & You are a helpful assistant.
    \\
    
    \hline
    Input Explanation 
    & The following are the tagging results of a set of images in the format of ``Image ID: tag 1, tag 2, ..., tag 10''. These assigned tags reflect the visible semantic content of each image:
    \\
    
    \hline
    Tag Embedding 
    & 
    Image 1: \texttt{"\{TAGS\}"}
    
    Image 2: \texttt{"\{TAGS\}"}

    ...

    Image N: \texttt{"\{TAGS\}"}
    \\
    
    \hline
    Goal Explanation 
    & I am a machine learning researcher trying to figure out the potential clustering or grouping criteria that exist in these images. So I can better understand my data and group them into different clusters based on different criteria.
    
    \\
    
    \hline
    Task Instruction 
    & Please analyze these images by using their assigned tags. Come up with an array of distinct clustering criteria that exist in this set of images.
    \\

    \hline
    Output Instruction
    & Please write a list of clustering criteria (separated by bullet points ``*''). 
    \\
    
    \hline
    Task Reinforcement
    & Again, I want to figure out what are the potential clustering or grouping criteria that I can use to group these images into different clusters.
    List an array of clustering or grouping criteria that often exist in this set of images based on the tagging results. Answer with a list (separated by bullet points ``*'').
    
    Your response:
    \\

    \Xhline{1.2pt}
  \end{tabular}
\end{table*}

%% file: tables/prompts/prompt_proposer_cap_mllm.tex
\begin{table*}[t!]
  \centering
  \caption{
  \textbf{Prompts for the \mllm{} in the caption-based proposer for generating detailed descriptions of the images.}
  }
  \lbltab{prompt_proposer_cap_mllm}
  
  \renewcommand\tabcolsep{10pt}
  \renewcommand\arraystretch{1.3}
  \footnotesize 
  
  \begin{tabular}{|p{0.2\linewidth}|p{0.66\linewidth}|} 
    \Xhline{1.2pt}
    
    \rowcolor{champagne} 
    \textbf{Prompt purpose} & \textbf{Prompt} \\
    
    \Xhline{1.2pt}
    
    System Prompt 
    & You are a helpful AI assistant
    \\

    \hline
    Task Instruction 
    & Describe the following image in detail.
    \\

    \Xhline{1.2pt}
  \end{tabular}
\end{table*}

%% file: tables/prompts/prompt_proposer_cap_llm.tex
\begin{table*}[t!]
  \centering
  \caption{
  \textbf{Prompts for the \llm{} used in the caption-based proposer for criteria proposing.} We embed the exact image captions by replacing the placeholders \texttt{"\{CAPTION\}"} in the prompt.
  }
  \lbltab{prompt_proposer_cap_llm}
  
  \renewcommand\tabcolsep{10pt}
  \renewcommand\arraystretch{1.3}
  \footnotesize 
  
  \begin{tabular}{|p{0.2\linewidth}|p{0.66\linewidth}|} 
    \Xhline{1.2pt}
    
    \rowcolor{champagne} 
    \textbf{Prompt purpose} & \textbf{Prompt} \\
    
    \Xhline{1.2pt}
    
    System Prompt 
    & You are a helpful assistant.
    \\
    
    \hline
    Input Explanation 
    & The following are the result of captioning a set of images:
    \\
    
    \hline
    Caption Embedding 
    & 
    Image 1: \texttt{"\{CAPTION\}"}
    
    Image 2: \texttt{"\{CAPTION\}"}

    ...

    Image N: \texttt{"\{CAPTION\}"}
    \\
    
    \hline
    Goal Explanation 
    & I am a machine learning researcher trying to figure out the potential clustering or grouping criteria that exist in these images. So I can better understand my data and group them into different clusters based on different criteria.
    \\
    
    \hline
    Task Instruction 
    & Come up with ten distinct clustering criteria that exist in this set of images.
    \\

    \hline
    Output Instruction
    &  Please write a list of clustering criteria (separated by bullet points ``*''). 
    \\
    
    \hline
    Task Reinforcement
    & Again I want to figure out what are the potential clustering/grouping criteria that I can use to group these images into different clusters. List ten clustering or grouping criteria that often exist in this set of images based on the captioning results. Answer with a list (separated by bullet points ``*'').
    
    Your response:
    \\

    \Xhline{1.2pt}
  \end{tabular}
\end{table*}

%% file: tables/prompts/prompt_pool_refine_llm.tex
\begin{table*}[t!]
  \centering
  \caption{
  \textbf{Prompts for the \llm{} used in Proposed Criteria Refinement step} We embed the exact initially discovered criteria by replacing the placeholders \texttt{"\{CRITERION\}"} in the prompt.
  }
  \lbltab{prompt_pool_refine_llm}
  
  \renewcommand\tabcolsep{10pt}
  \renewcommand\arraystretch{1.3}
  \footnotesize 
  
  \begin{tabular}{|p{0.2\linewidth}|p{0.66\linewidth}|} 
    \Xhline{1.2pt}
    
    \rowcolor{champagne} 
    \textbf{Prompt purpose} & \textbf{Prompt} \\
    
    \Xhline{1.2pt}
    
    System Prompt 
    & You are a helpful assistant.
    \\

    \hline
    Input Explanation 
    & I am a machine learning researcher working with a set of images. I aim to cluster this set of images based on the various clustering criteria present within them. Below is a preliminary list of clustering criteria that I've discovered to group these images:
    \\

    \hline
    Criteria Embedding: 
    & * Criterion 1: \texttt{"\{CRITERION\}"}

    * Criterion 2: \texttt{"\{CRITERION\}"}

    ...
    
    * Criterion L: \texttt{"\{CRITERION\}"}
    \\

    \hline
    Goal Explanation 
    & My goal is to refine this list by merging similar criteria and rephrasing them using more precise and informative terms. This will help create a set of distinct, optimized clustering criteria.
    \\

    \hline
    Task Instruction 
    & Your task is to first review and understand the initial list of clustering criteria provided. Then, assist me in refining this list by:
    
    * Merging similar criteria.
    
    * Expressing each criterion more clearly and informatively.
    \\

    \hline
    Output Instruction
    & Please respond with the cleaned and optimized list of clustering criteria, formatted as bullet points (using ``*'').
    
    Your response:
    \\

    \Xhline{1.2pt}
  \end{tabular}
\end{table*}

%% file: tables/prompts/prompt_grouper_img_llm.tex
\begin{table*}[t!]
  \centering
  \caption{
  \textbf{Prompts for the \llm{} used in the image-based grouper for automatic criterion-specific \vqa{} question generation.} We embed the exact discovered criterion by replacing the placeholder \texttt{"\{CRITERION\}"} in the prompt.
  }
  \lbltab{prompt_grouper_img_llm}
  
  \renewcommand\tabcolsep{10pt}
  \renewcommand\arraystretch{1.3}
  \footnotesize 
  
  \begin{tabular}{|p{0.2\linewidth}|p{0.66\linewidth}|} 
    \Xhline{1.2pt}
    
    \rowcolor{champagne} 
    \textbf{Prompt purpose} & \textbf{Prompt} \\
    
    \Xhline{1.2pt}
    
    System Prompt 
    & You are a helpful assistant.
    \\

    \hline
    Goal Explanation 
    & Hello! I am a machine learning researcher focusing on image categorization based on the aspect of \texttt{"\{CRITERION\}"} depicted in images. 
    \\
    
    \hline
    Task Instruction 
    & Therefore, I need your assistance in designing a prompt for the Visual Question Answering (VQA) model to help it identify the \texttt{"\{CRITERION\}"} category in a given image at three different granularity. Please help me design and generate this prompt using the following template: "Question: [Generated VQA Prompt Question] Answer (reply with an abstract, a common, and a specific category name, respectively):". The generated prompt should be simple and straightforward.
    \\

    \hline
    Output Instruction
    &  Please respond with only the generated prompt using the following format ``* Answer *''.
    
    Your response:
    \\

    \Xhline{1.2pt}
  \end{tabular}
    \lesspace

\end{table*}

%% file: tables/prompts/prompt_grouper_tag_llm_mid.tex
\begin{table*}[t!]
  \centering
  \caption{
  \textbf{Prompts for the \llm{} used in the tag-based grouper for generating middle-grained criterion-specific tags.} We embed the exact discovered criterion by replacing the placeholder \texttt{"\{CRITERION\}"} in the prompt.
  }
  \lbltab{prompt_grouper_tag_llm_mid}
  
  \renewcommand\tabcolsep{10pt}
  \renewcommand\arraystretch{1.3}
  \footnotesize 
  
  \begin{tabular}{|p{0.2\linewidth}|p{0.66\linewidth}|} 
    \Xhline{1.2pt}
    
    \rowcolor{champagne} 
    \textbf{Prompt purpose} & \textbf{Prompt} \\
    
    \Xhline{1.2pt}
    
    System Prompt 
    & You are a helpful assistant.
    \\

    \hline
    Goal Explanation 
    & Hello! I am a machine learning researcher focusing on image categorization of a certain aspect. I'm interested in generating a list of tags specifically for categorizing the types of \texttt{"\{CRITERION\}"} depicted in images.
    \\
    
    \hline
    Task Instruction 
    & Please provide a list of potential \texttt{"\{CRITERION\}"} category names. Please generate diverse category names. Do not include too general or specific category names such as ``Sports''. 
    \\

    \hline
    Output Instruction
    & Please respond with the list of category names. Each category should be formatted as follows: ``* Category Name''.
    
    Your response:
    \\

    \Xhline{1.2pt}
  \end{tabular}
\end{table*}

%% file: tables/prompts/prompt_grouper_tag_llm_coarse.tex
\begin{table*}[t!]
  \centering
  \caption{
  \textbf{Prompts for the \llm{} used in the tag-based grouper for generating coarse-grained criterion-specific tags.} We embed the exact discovered criterion and middle-grained category by replacing the placeholder \texttt{"\{CRITERION\}"} and \texttt{"\{MIDDLE-GRAINED CATEGORY NAME\}"} in the prompt, respectively.
  }
  \lbltab{prompt_grouper_tag_llm_coarse}
  
  \renewcommand\tabcolsep{10pt}
  \renewcommand\arraystretch{1.3}
  \footnotesize 
  
  \begin{tabular}{|p{0.2\linewidth}|p{0.66\linewidth}|} 
    \Xhline{1.2pt}
    
    \rowcolor{champagne} 
    \textbf{Prompt purpose} & \textbf{Prompt} \\
    
    \Xhline{1.2pt}
    
    System Prompt 
    & You are a helpful assistant.
    \\

    \hline
    Task Instruction 
    & Generate a list of three more abstract or general \texttt{"\{CRITERION\}"} super-categories that the following \texttt{"\{CRITERION\}"} category belongs to and output the list separated by ``\&'' (without numbers): \texttt{"\{MIDDLE-GRAINED CATEGORY NAME\}"}
    \\

    \hline
    Output Instruction
    & Your response:
    \\

    \Xhline{1.2pt}
  \end{tabular}
\end{table*}

%% file: tables/prompts/prompt_grouper_tag_llm_fine.tex
\begin{table*}[t!]
  \centering
  \caption{
  \textbf{Prompts for the \llm{} used in the tag-based grouper for generating fine-grained criterion-specific tags.} We embed the exact discovered criterion and middle-grained category by replacing the placeholder \texttt{"\{CRITERION\}"} and \texttt{"\{MIDDLE-GRAINED CATEGORY NAME\}"} in the prompt, respectively.
  }
  \lbltab{prompt_grouper_tag_llm_fine}
  
  \renewcommand\tabcolsep{10pt}
  \renewcommand\arraystretch{1.3}
  \footnotesize 
  
  \begin{tabular}{|p{0.2\linewidth}|p{0.66\linewidth}|} 
    \Xhline{1.2pt}
    
    \rowcolor{champagne} 
    \textbf{Prompt purpose} & \textbf{Prompt} \\
    
    \Xhline{1.2pt}
    
    System Prompt 
    & You are a helpful assistant.
    \\

    \hline
    Task Instruction 
    & 
    Generate a list of ten more detailed or specific \texttt{"\{CRITERION\}"} sub-categories of the following \texttt{"\{CRITERION\}"} category and output the list separated by ``\&'' (without numbers): \texttt{"\{MIDDLE-GRAINED CATEGORY NAME\}"}
    \\

    \hline
    Output Instruction
    & Your response:
    \\

    \Xhline{1.2pt}
  \end{tabular}
\end{table*}

%% file: tables/prompts/prompt_grouper_cap_mllm.tex
\begin{table*}[t!]
  \centering
  \caption{
  \textbf{Prompts for the \mllm{} used in the caption-based grouper for generating criterion-specific captions.} We embed the exact discovered criterion by replacing the placeholder \texttt{"\{CRITERION\}"} in the prompt.
  }
  \lbltab{prompt_grouper_cap_mllm}
  
  \renewcommand\tabcolsep{10pt}
  \renewcommand\arraystretch{1.3}
  \footnotesize 
  
  \begin{tabular}{|p{0.2\linewidth}|p{0.66\linewidth}|} 
    \Xhline{1.2pt}
    
    \rowcolor{champagne} 
    \textbf{Prompt purpose} & \textbf{Prompt} \\
    
    \Xhline{1.2pt}
    
    System Prompt 
    & You are a helpful AI assistant.
    \\

    \hline
    Task Instruction 
    & 
    Analyze the image focusing specifically on the \texttt{"\{CRITERION\}"}. Provide a detailed description of the \texttt{"\{CRITERION\}"} depicted in the image. Highlight key elements and interactions relevant to the \texttt{"\{CRITERION\}"} that enhance the understanding of the scene.
    \\

    \hline
    Output Instruction
    & Your response:
    \\

    \Xhline{1.2pt}
  \end{tabular}
\lesspace

\end{table*}

%% file: tables/prompts/prompt_grouper_cap_llm_initial.tex
\begin{table*}[t!]
  \centering
  \caption{
  \textbf{Prompts for the \llm{} used in the caption-based grouper at the \textit{Initial Naming} step for initially assigning a criterion-based category name to the image based on its criterion-specific caption.} 
  We embed the exact discovered criterion and the corresponding criterion-specific caption by replacing the placeholder \texttt{"\{CRITERION\}"} and \texttt{"\{CRITERION-SPECIFIC CAPTION\}"} in the prompt, respectively.
  }
  \lbltab{prompt_grouper_cap_llm_initial}
  
  \renewcommand\tabcolsep{10pt}
  \renewcommand\arraystretch{1.3}
  \footnotesize 
  
  \begin{tabular}{|p{0.2\linewidth}|p{0.66\linewidth}|} 
    \Xhline{1.2pt}
    
    \rowcolor{champagne} 
    \textbf{Prompt purpose} & \textbf{Prompt} \\
    
    \Xhline{1.2pt}
    
    System Prompt 
    & You are a helpful assistant.
    \\
    
    \hline
    Input Explanation 
    & The following is the description about the \texttt{"\{CRITERION\}"} of an image:
    \\
    
    \hline
    Caption Embedding 
    & 
    \texttt{"\{CRITERION-SPECIFIC CAPTION\}"}
    \\
    
    \hline
    Goal Explanation 
    & I am a machine learning researcher trying to assign a label to this image based on what is the \texttt{"\{CRITERION\}"} depicted in this image.

    \\
    
    \hline
    Task Instruction 
    & Understand the provided description carefully and assign a label to this image based on what is the \texttt{"\{CRITERION\}"} depicted in this image. 
    \\

    \hline
    Output Instruction
    & Please respond in the following format within five words: "*Answer*". Do not talk about the description and do not respond long sentences. The answer should be within five words.
    \\

    \hline
    Task Reinforcement
    & Again, your job is to understand the description and assign a label to this image based on what is the \texttt{"\{CRITERION\}"} shown in this image.
    
    Your response:
    \\

    \Xhline{1.2pt}
  \end{tabular}
\end{table*}

%% file: tables/prompts/prompt_grouper_cap_llm_refine.tex
\begin{table*}[t!]
  \centering
  \caption{
  \textbf{Prompts for the \llm{} used in the caption-based grouper at the \textit{Multi-granularity Cluster Generation} step for refining the initially assigned names to a structured three granularity levels.}
  We embed the exact discovered criterion and the initially assigned name categories by replacing the placeholder \texttt{"\{CRITERION\}"} and \texttt{"\{MIDDLE-GRAINED CATEGORY NAME\}"} in the prompt, respectively.
  }
  \lbltab{prompt_grouper_cap_llm_refine}
  
  \renewcommand\tabcolsep{10pt}
  \renewcommand\arraystretch{1.3}
  \footnotesize 
  
  \begin{tabular}{|p{0.2\linewidth}|p{0.66\linewidth}|} 
    \Xhline{1.2pt}
    
    \rowcolor{champagne} 
    \textbf{Prompt purpose} & \textbf{Prompt} \\
    
    \Xhline{1.2pt}
    
    System Prompt 
    & You are a helpful assistant.
    \\
    
    \hline
    Input Explanation 
    & The following is an initial list of \texttt{"\{CRITERION\}"} categories. These categories might not be at the same semantic granularity level. For example, category 1 could be ``cutting vegetables'', while category 2 is simply ``cutting''. In this case, category 1 is more specific than category 2.
    \\
    
    \hline
    Category Embedding 
    & 
    * \texttt{"\{MIDDLE-GRAINED CATEGORY NAME\}"}
    
    * \texttt{"\{MIDDLE-GRAINED CATEGORY NAME\}"}

    ...

    * \texttt{"\{MIDDLE-GRAINED CATEGORY NAME\}"}
    \\
    
    \hline
    Task Instruction 
    & 
    These categories might not be at the same semantic granularity level. For example, category 1 could be ``cutting vegetables'', while category 2 is simply ``cutting''. In this case, category 1 is more specific than category 2.
    Your job is to generate a three-level class hierarchy (class taxonomy, where the first level contains more abstract or general coarse-grained classes, the third level contains more specific fine-grained classes, and the second level contains intermediate mid-grained classes) of \texttt{"\{CRITERION\}"} based on the provided list of \texttt{"\{CRITERION\}"} categories. Follow these steps to generate the hierarchy.
    \\
    
    \hline
    Sub-task Instruction 
    & Follow these steps to generate the hierarchy:

    Step 1 - Understand the provided initial list of \texttt{"\{CRITERION\}"} categories. The following three-level class hierarchy generation steps are all based on the provided initial list.

    Step 2 - Generate a list of abstract or general \texttt{"\{CRITERION\}"} categories as the first level of the class hierarchy, covering all the concepts present in the initial list.
    
    Step 3 - Generate a list of middle-grained \texttt{"\{CRITERION\}"} categories as the second level of the class hierarchy, in which the middle-grained categories are the subcategories of the categories in the first level. The categories in the second-level are more specific than the first level but should still cover and reflect all the concepts present in the initial list.
    
    Step 4 - Generate a list of more specific fine-grained \texttt{"\{CRITERION\}"} categories as the third level of the class hierarchy, in which the categories should reflect more specific \texttt{"\{CRITERION\}"} concepts that you can infer from the initial list. The categories in the third-level are subcategories of the second-level.
    
    Step 5 - Output the generated three-level class hierarchy as a JSON object where the keys are the level numbers and the values are a flat list of generated categories at each level, structured like:
    
    \{
    
    ``level 1": [``categories''],
    
    ``level 2": [``categories''],
    
    ``level 3": [``categories'']
    
    \}
    \\

    \hline
    Output Instruction
    & Please only output the JSON object in your response and simply use a flat list to store the generated categories at each level.
    
    Your response:
    \\

    \Xhline{1.2pt}
  \end{tabular}
\end{table*}

%% file: tables/prompts/prompt_grouper_cap_llm_final.tex
\begin{table*}[t!]
  \centering
  \caption{
  \textbf{Prompts for the \llm{} used in the caption-based grouper at the \textit{Final Assignment} step.}
  We embed the exact discovered criterion and the refined category names from each granularity level, by replacing the placeholder \texttt{"\{CRITERION\}"} and \texttt{"\{CANDIDATE CATEGORY NAME\}"} in the prompt, respectively.
  }
  \lbltab{prompt_grouper_cap_llm_final}
  
  \renewcommand\tabcolsep{10pt}
  \renewcommand\arraystretch{1.3}
  \footnotesize 
  
  \begin{tabular}{|p{0.2\linewidth}|p{0.66\linewidth}|} 
    \Xhline{1.2pt}
    
    \rowcolor{champagne} 
    \textbf{Prompt purpose} & \textbf{Prompt} \\
    
    \Xhline{1.2pt}
    
    System Prompt 
    & You are a helpful assistant.
    \\
    
    \hline
    Input Explanation 
    & The following is a detailed description about the \texttt{"\{CRITERION\}"} of an image.
    \\
    
    \hline
    Caption Embedding 
    & 
    \texttt{"\{CRITERION-SPECIFIC CAPTION\}"}
    \\
    
    \hline
    Task Instruction 
    & Based on the content and details provided in the description, classify the image into one of the specified \texttt{"\{CRITERION\}"} categories listed below:
    \\
    
    \hline
    Candidate Category
    
    Embedding 
    &  \texttt{"\{CRITERION\}"} categories:

    * \texttt{"\{CANDIDATE CATEGORY NAME\}"}
    
    * \texttt{"\{CANDIDATE CATEGORY NAME\}"}

    ...

    * \texttt{"\{CANDIDATE CATEGORY NAME\}"}
    \\

    \hline
    Output Instruction
    &  Ensure that your classification adheres to the details mentioned in the image description. Respond with the classification result in the following format: ``*category name*''.
    
    Your response:
    \\

    \Xhline{1.2pt}
  \end{tabular}
  \lesspace
\end{table*}

%% file: tables/results_proposer/recall_main.tex
\begin{table*}[!ht]
\centering
\caption{
\textbf{Comparison of True Positive Rate (\tpr) (\%) for criteria proposers across the six \taskshort benchmarks}. \tpr performance is reported for both Basic and Hard ground-truth criteria. The best performance is highlighted in bold.
}
\lbltab{recall_main}
\tablestyle{2.4pt}{1.2}

\begin{tabular}{lrrrrrrrrrrrrrr}
\toprule

& \multicolumn{2}{c}{{\ourcoco}}
& \multicolumn{2}{c}{{\ourfood}}
& \multicolumn{2}{c}{{\action}}
& \multicolumn{2}{c}{{\clevr}}
& \multicolumn{2}{c}{{\card}}
& \multicolumn{2}{c}{{\fruit}}
& \multicolumn{2}{c}{{Average}}

\\

& \multicolumn{1}{c}{Basic} & \multicolumn{1}{c}{Hard}
& \multicolumn{1}{c}{Basic} & \multicolumn{1}{c}{Hard}
& \multicolumn{1}{c}{Basic} & \multicolumn{1}{c}{Hard}
& \multicolumn{1}{c}{Basic} & \multicolumn{1}{c}{Hard}
& \multicolumn{1}{c}{Basic} & \multicolumn{1}{c}{Hard}
& \multicolumn{1}{c}{Basic} & \multicolumn{1}{c}{Hard}
& \multicolumn{1}{c}{Basic} & \multicolumn{1}{c}{Hard}
\\
\toprule

{Image-based}
&  \color{codegreen}\bf100.0&  52.9
&  25.0	&  36.4
&  66.7	&  54.6
&  50.0	&  28.6
&  50.0	&  25.0
&  50.0	&  20.0
&  56.9	&  36.2
\\

{Tag-based}
&  50.0&  	35.3
&  \color{codegreen}\bf100.0&  72.7
&  66.7&  	36.4
&  75.0	&  42.9
&  50.0	&  50.0
&  50.0	&  20.0
&  65.3	&  42.9
\\

{Caption-based}
&  \color{codegreen}\bf100.0&  \color{codegreen}\bf64.7
&  \color{codegreen}\bf100.0&  \color{codegreen}\bf81.8
&  \color{codegreen}\bf100.0&  \color{codegreen}\bf72.7
&  \color{codegreen}\bf100.0&  \color{codegreen}\bf71.4
&  \color{codegreen}\bf100.0&  \color{codegreen}\bf100.0
&  \color{codegreen}\bf100.0&  \color{codegreen}\bf60.0
&  \color{codegreen}\bf 100.0 &  \color{codegreen}\bf 75.1
\\

\bottomrule
\end{tabular}
\end{table*}

%% file: tables/results_proposer/recall_study_caption.tex
\begin{table*}[!ht]
\centering
\small

\caption{
\textbf{Study of the impact of data scale on criteria discovery.} The Caption-based Proposer is used for criteria discovery, and \tpr performance (\%) is reported on the \textit{Hard} ground-truth criteria sets across the six \taskshort benchmarks for different data scales. The best performance is highlighted in bold.
}

\lbltab{recall_study_caption}

\begin{tabular}{lrrrrrrr}
\toprule

\multicolumn{1}{c}{{Data scales}}
& \multicolumn{1}{c}{{\ourcoco}}
& \multicolumn{1}{c}{{\ourfood}}
& \multicolumn{1}{c}{{\action}}
& \multicolumn{1}{c}{{\clevr}}
& \multicolumn{1}{c}{{\card}}
& \multicolumn{1}{c}{{\fruit}}
& \multicolumn{1}{c}{{Average}}

\\

\toprule

{100\%}
& \color{codegreen}\bf{64.7} & \color{codegreen}\bf{81.8} & \color{codegreen}\bf72.7 & 71.4 & \color{codegreen}\bf 100.0 & \color{codegreen}\bf{60.0} & \color{codegreen}\bf 75.1
\\

{80\%}
& 47.1 & 72.7 & 54.6 & 71.4 & 75.0 & 30.0 & 58.5
\\

{60\%}
& 52.9 & 63.6 & 54.6 & 71.4 & \color{codegreen}\bf 100.0 & 50.0 & 65.4
\\

{40\%}
& 41.2 & 45.5 & 45.5 & \color{codegreen}\bf{85.7} & \color{codegreen}\bf 100.0 & 40.0 & 59.6
\\

{20\%}
& 35.3 & 45.5 & 36.4 & 42.9 & \color{codegreen}\bf 100.0 & 40.0 & 50.0
\\

{1 img}
& 23.5 & 36.4 & 27.3 & 57.1 & 75.0 & 50.0 & 44.9
\\

\bottomrule
\end{tabular}
\end{table*}

%% file: tables/results_grouper/grouper_main_coco.tex
\begin{table*}[!t]
\centering

\caption{
\textbf{Comparison of Semantic Groupers on \ourcoco.} We report \cacctitle (\cacc), \sacctitle (\sacc), and their Harmonic Mean (HM) in percentages (\%). These results are plotted in \reffig{study_grouper}(a).
}

\lbltab{grouper_main_coco}
\tablestyle{3.8pt}{1.2}

\begin{tabular}{l|rrr|rrr|rrr|rrr}
\toprule

\multirow{2}{*}{Methods}
& \multicolumn{3}{c|}{Activity}
& \multicolumn{3}{c|}{Location}
& \multicolumn{3}{c|}{Mood}
& \multicolumn{3}{c}{Time of Day}
\\

& \cacc & \sacc & HM
& \cacc & \sacc & HM
& \cacc & \sacc & HM
& \cacc & \sacc & HM
\\

\toprule

\rowcolor{verygrey} 
CLIP Zero-shot
& 62.6 & 73.5 & 67.6 
& 34.3 & 51.5 & 41.1
& 22.4 & 43.3 & 29.5
& 40.6 & 74.1 & 52.4
\\

\hline

KMeans CLIP
& 34.4 & - & -
& 32.7 & - & -
& 18.9 & - & -
& 38.6 & - & -
\\

KMeans DINOv1
& 34.8 & - & -
& 37.5 & - & -
& 17.9 & - & -
& 36.5 & - & -
\\

KMeans DINOv2
& 38.2 & - & -
& 37.9 & - & -
& 22.5 & - & -
& 43.8& - & -
\\

\hline

Img-based BLIP-2
& 48.7 & 64.1 & \color{codegreen}\bf55.3
& 39.6 & 48.0 & 43.4
& 30.2 & 37.5 & 33.4
& 40.7 & 60.3 & 48.6
\\

Img-based LLaVA
& 46.5 & 61.8 & 53.1
& 34.0 & 46.3 & 39.2
& 28.0 & 24.7 & 26.3
& 39.4 & 51.7 & 44.7
\\

Tag-based
& 43.2 & 51.5 & 47.0
& 28.6 & 46.6 & 35.5
& 13.0 & 25.6 & 17.2
& 19.3 & 48.8 & 27.7
\\

Caption-based
& 44.1 & 48.9 & 46.4
& 55.2 & 55.6 & \color{codegreen}\bf55.4
& 38.1 & 32.6 & \color{codegreen}\bf35.2
& 67.6 & 56.7 & \color{codegreen}\bf61.7
\\

\bottomrule
\end{tabular}
\end{table*}

%% file: tables/results_grouper/grouper_main_card.tex
\begin{table*}[!t]
\centering

\caption{
\textbf{Comparison of Semantic Groupers on \card.} We report \cacctitle (\cacc), \sacctitle (\sacc), and their Harmonic Mean (HM) in percentages (\%). These results are plotted in \reffig{study_grouper}(b).
}
\lbltab{grouper_main_card}
\tablestyle{3.8pt}{1.2}

\begin{tabular}{l|rrr|rrr}
\toprule

\multirow{2}{*}{Methods}
& \multicolumn{3}{c|}{Suit}
& \multicolumn{3}{c}{Rank}
\\

& \cacc & \sacc & HM
& \cacc & \sacc & HM
\\

\toprule
	
\rowcolor{verygrey} 
CLIP Zero-shot
& 47.9 & 69.5 & 56.7
& 35.0 & 64.2 & 45.3
\\

\hline

KMeans CLIP
& 45.0 & - & -
& 28.6 & - & -
\\

KMeans DINOv1
& 38.5 & - & -
& 20.7 & - & -
\\

KMeans DINOv2
& 36.7 & - & -
& 22.3 & - & -
\\

\hline

Img-based BLIP-2
& 66.7 & 77.7 & \color{codegreen}\bf71.8
& 47.5 & 54.4 & 50.7
\\

Img-based LLaVA
& 36.8 & 65.8 & 47.2
& 24.6 & 49.8 & 32.9
\\

Tag-based
& 39.2 & 32.9 & 35.8
& 22.3 & 39.1 & 28.4
\\

Caption-based
& 54.5 & 73.6 & 62.6
& 92.1 & 95.1 & \color{codegreen}\bf93.6
\\

\bottomrule
\end{tabular}
\end{table*}

%% file: tables/results_grouper/grouper_main_action.tex
\begin{table*}[!t]
\centering

\caption{
\textbf{Comparison of Semantic Groupers on \action.} We report \cacctitle (\cacc), \sacctitle (\sacc), and their Harmonic Mean (HM) in percentages (\%). These results are plotted in \reffig{study_grouper}(c).
}
\lbltab{grouper_main_action}
\tablestyle{3.8pt}{1.2}

\begin{tabular}{l|rrr|rrr|rrr}
\toprule

\multirow{2}{*}{Methods}
& \multicolumn{3}{c|}{Action}
& \multicolumn{3}{c|}{Location}
& \multicolumn{3}{c}{Mood}
\\

& \cacc & \sacc & HM
& \cacc & \sacc & HM
& \cacc & \sacc & HM
\\

\toprule

\rowcolor{verygrey} 
CLIP Zero-shot
& 97.1 & 99.2 & 98.1
& 66.7 & 67.1 & 66.9
& 75.5 & 80.7 & 78.0
\\

\hline

KMeans CLIP
& 62.3 & - & -
& 58.3 & - & -
&  & - & -
\\

KMeans DINOv1
& 49.3 & - & -
& 61.4 & - & -
&  & - & -
\\

KMeans DINOv2
& 75.7 & - & -
& 67.6 & - & -
&  & - & -
\\

\hline

Img-based BLIP-2
& 79.7 & 80.9 & 80.3
& 43.3 & 42.4 & 42.8
& 43.1 & 43.8 & 43.4
\\

Img-based LLaVA
& 70.1 & 60.5 & 65.0
& 45.8 & 42.8 & 44.2
& 32.0 & 38.0 & 34.7
\\

Tag-based
& 70.2 & 55.0 & 61.6
& 36.8 & 48.1 & 41.7
& 50.7 & 47.6 & 49.1
\\

Caption-based
& 82.8 & 82.8 & \color{codegreen}\bf82.8
& 69.8 & 55.2 & \color{codegreen}\bf61.6
& 52.3 & 50.2 & \color{codegreen}\bf51.2
\\

\bottomrule
\end{tabular}
\end{table*}

%% file: tables/results_grouper/grouper_main_food.tex
\begin{table*}[!t]
\centering

\caption{
\textbf{Comparison of Semantic Groupers on \ourfood.} We report \cacctitle (\cacc), \sacctitle (\sacc), and their Harmonic Mean (HM) in percentages (\%). These results are plotted in \reffig{study_grouper}(d).
}
\lbltab{grouper_main_food}
\tablestyle{3.8pt}{1.2}

\begin{tabular}{l|rrr|rrr|rrr|rrr}
\toprule

\multirow{2}{*}{Methods}
& \multicolumn{3}{c|}{Food Type}
& \multicolumn{3}{c|}{Cuisine}
& \multicolumn{3}{c|}{Course}
& \multicolumn{3}{c}{Diet}
\\

& \cacc & \sacc & HM
& \cacc & \sacc & HM
& \cacc & \sacc & HM
& \cacc & \sacc & HM
\\

\toprule

\rowcolor{verygrey} 
CLIP Zero-shot
& 90.6 & 94.6 & 92.6
& 54.9 & 81.4 & 65.6
& 63.5 & 84.7 & 72.6
& 47.6 & 59.9 & 53.0
\\

\hline

KMeans CLIP
& 66.1 & - & -
& 29.8 & - & -
& 49.5 & - & -
& 36.9 & - & -
\\

KMeans DINOv1
& 33.6 & - & -
& 15.3 & - & -
& 38.1 & - & -
& 41.4 & - & -
\\

KMeans DINOv2
& 72.7 & - & -
& 22.5 & - & -
& 47.6 & - & -
& 43.4 & - & -
\\

\hline

Img-based BLIP-2
& 54.2 & 71.4 & \color{codegreen}\bf61.6
& 54.8 & 73.3 & \color{codegreen}\bf62.7
& 42.3 & 71.0 & 53.0
& 34.2 & 53.8 & 41.9
\\

Img-based LLaVA
& 42.2 & 64.0 & 50.9
& 33.7 & 57.6 & 42.6
& 46.9 & 73.1 & 57.1
& 27.0 & 40.5 & 32.4
\\

Tag-based
& 45.0 & 63.3 & 52.6
& 48.8 & 42.1 & 45.2
& 42.7 & 70.1 & 53.1
& 25.2 & 34.1 & 29.0
\\

Caption-based
& 34.6 & 54.2 & 42.2
& 47.0 & 65.9 & 54.9
& 69.1 & 85.7 & \color{codegreen}\bf76.5
& 41.5 & 54.0 & \color{codegreen}\bf46.9
\\

\bottomrule
\end{tabular}
\end{table*}

%% file: tables/results_grouper/grouper_main_fruit.tex
\begin{table*}[!t]
\centering

\caption{
\textbf{Comparison of Semantic Groupers on \fruit.} We report \cacctitle (\cacc), \sacctitle (\sacc), and their Harmonic Mean (HM) in percentages (\%). These results are plotted in \reffig{study_grouper}(e).
}
\lbltab{grouper_main_fruit}
\tablestyle{3.8pt}{1.2}

\begin{tabular}{l|rrr|rrr}
\toprule

\multirow{2}{*}{Methods}
& \multicolumn{3}{c|}{Species}
& \multicolumn{3}{c}{Color}
\\

& \cacc & \sacc & HM
& \cacc & \sacc & HM
\\

\toprule

\rowcolor{verygrey} 
CLIP Zero-shot
& 84.0 & 93.1 & 88.3
& 54.8 & 83.5 & 66.1
\\

\hline

KMeans CLIP
& 67.1 & - & -
& 39.6 & - & -
\\

KMeans DINOv1
& 53.8 & - & -
& 36.0 & - & -
\\

KMeans DINOv2
& 71.2 & - & -
& 36.7 & - & -
\\

\hline

Img-based BLIP-2
& 70.7 & 68.3 & 69.5
& 40.9 & 70.6 & 51.8
\\

Img-based LLaVA
& 63.9 & 67.8 & 65.8
& 51.0 & 83.2 & \color{codegreen}\bf63.2
\\

Tag-based
& 64.0 & 67.1 & 65.5
& 54.1 & 44.1 & 48.6
\\

Caption-based
& 76.9 & 70.7 & \color{codegreen}\bf73.7
& 53.3 & 51.5 & 52.4
\\

\bottomrule
\end{tabular}
\end{table*}

%% file: tables/results_grouper/grouper_main_clevr.tex
\begin{table*}[!t]
\centering

\caption{
\textbf{Comparison of Semantic Groupers on \clevr.} We report \cacctitle (\cacc), \sacctitle (\sacc), and their Harmonic Mean (HM) in percentages (\%). These results are plotted in \reffig{study_grouper}(f).
}
\lbltab{grouper_main_clevr}
\tablestyle{3.8pt}{1.2}

\begin{tabular}{l|rrr|rrr|rrr|rrr}
\toprule

\multirow{2}{*}{Methods}
& \multicolumn{3}{c|}{Color}
& \multicolumn{3}{c|}{Texture}
& \multicolumn{3}{c|}{Count}
& \multicolumn{3}{c}{Shape}
\\

& \cacc & \sacc & HM
& \cacc & \sacc & HM
& \cacc & \sacc & HM
& \cacc & \sacc & HM
\\

\toprule

\rowcolor{verygrey} 
CLIP Zero-shot
& 77.7 & 94.0 & 85.1
& 34.1 & 41.9 & 37.6
& 43.7 & 81.5 & 56.9
& 71.1 & 72.7 & 71.9
\\

\hline

KMeans CLIP
& 48.8 & - & -
& 61.4 & - & -
& 44.2 & - & -
& 56.1 & - & -
\\

KMeans DINOv1
& 53.0 & - & -
& 58.4 & - & -
& 47.5 & - & -
& 67.0 & - & -
\\

KMeans DINOv2
& 44.1 & - & -
& 46.9 & - & -
& 52.5 & - & -
& 87.0 & - & -
\\

\hline

Img-based BLIP-2
& 69.3 & 76.5 & \color{codegreen}\bf72.7
& 57.8 & 34.4 & 43.1
& 25.7 & 55.9 & 35.2
& 69.1 & 62.6 & 65.7
\\

Img-based LLaVA
& 56.5 & 63.5 & 59.8
& 51.9 & 26.9 & 35.4
& 53.7 & 39.4 & 45.4
& 64.3 & 71.3 & \color{codegreen}\bf67.6
\\

Tag-based
& 66.6 & 55.3 & 60.4
& 57.2 & 40.2 & 47.3
& 47.4 & 8.3 & 14.1
& 62.7 & 36.5 & 46.2
\\

Caption-based
& 70.3 & 63.4 & 66.7
& 65.3 & 42.1 & \color{codegreen}\bf51.2
& 65.7 & 73.3 & \color{codegreen}\bf69.3
& 58.4 & 38.5 & 46.4
\\

\bottomrule
\end{tabular}
\end{table*}

%% file: tables/rebuttal/stats_num_clusters.tex
\begin{table*}[!ht]
\centering

\caption{
{
\textbf{Summary of cluster counts across six benchmarks for the comparison of semantic groupers.} The results yield by the main Caption-based Grouper is reported. Specifically, we report: \textit{i)} GT: the number of ground-truth clusters; \textit{ii)} Pred-Init: predicted clusters from initial names; \textit{iii)} Pred-Coarse: predicted coarse-grained clusters after multi-granularity refinement; \textit{iv)} Pred-Middle: predicted middle-grained clusters after multi-granularity refinement; and \textit{v)} Pred-Fine: predicted fine-grained clusters after multi-granularity refinement.
}
}

\lbltab{stats_num_clusters}

\begin{tabular}{ll|rrrrr}

\toprule

Dataset
& Criteria
& \multicolumn{1}{c}{GT}
& \multicolumn{1}{c}{Pred-Init}
& \multicolumn{1}{c}{Pred-Corase}
& \multicolumn{1}{c}{Pred-Middle}
& \multicolumn{1}{c}{Pred-Fine}

\\

\toprule

\multirow{4}{*}{\ourcoco}
& Activity
& 64
& 203
& 12
& 23
& 52
\\
& Location
& 19
& 145
& 7
& 14
& 28
\\
& Mood
& 20
& 122
& 15
& 25
& 30
\\
& Time of Day
& 6
& 96
& 2
& 8
& 31
\\

\hline

\multirow{4}{*}{\ourfood}
& Food Type
& 101
& 301
& 7
& 37
& 127
\\
& Cuisine
& 15
& 141
& 9
& 18
& 53
\\
& Course
& 5
& 97
& 4
& 12
& 78
\\
& Diet
& 4
& 139
& 5
& 8
& 64
\\

\hline

\multirow{3}{*}{\action}
& Action
& 40
& 71
& 8
& 15
& 51
\\
& Location
& 10
& 82
& 5
& 10
& 67
\\
& Mood
& 4
& 95
& 6
& 18
& 55
\\

\hline

\multirow{4}{*}{\clevr}
& Color
& 10
& 25
& 6
& 12
& 17
\\
& Texture
& 10
& 23
& 2
& 5
& 12
\\
& Shape
& 10
& 22
& 5
& 11
& 14
\\
& Count
& 10
& 11
& 2
& 4
& 11
\\

\hline

\multirow{2}{*}{\card}
& Rank
& 14
& 147
& 4
& 7
& 16
\\
& Suit
& 5
& 56
& 4
& 7
& 30
\\

\hline

\multirow{2}{*}{\fruit}
& Species
& 34
& 54
& 8
& 25
& 38
\\
& Color
& 15
& 66
& 5
& 15
& 39
\\
\bottomrule
\end{tabular}
\end{table*}

%% file: tables/results_grouper/grouper_sota.tex
\begin{table*}[!ht]
\centering

\caption{
\textbf{Comparison with criteria-conditioned clustering methods on the six \taskshort benchmarks.} We report \cacctitle (\cacc) and \sacctitle (\sacc)as percentages (\%). Average (\textit{Avg.}) \cacc and \sacc across different criteria on each dataset is also provided. For reference, we include the pseudo upper-bound (UB) performance of CLIP ViT-L/14 in zero-shot transfer, using ground-truth criteria and class names. Note that both \ictc and \mmap utilize ground-truth criteria and the number of clusters for clustering. These expanded results correspond to \reftab{comp_sota}.
}

\lbltab{grouper_sota}
\tablestyle{6.9pt}{1}

\begin{tabular}{llcccccccccccc}

\toprule

\multirow{2}{*}{Benchmark}
& \multirow{2}{*}{Criterion}
& \multicolumn{2}{c}{\color{gray}UB}
& \multicolumn{2}{c}{\ictc}
& \multicolumn{2}{c}{\ssdllm}
& \multicolumn{2}{c}{\mmap}
& \multicolumn{2}{c}{\msub}

& \multicolumn{2}{c}{Ours}
\\

&
& \color{gray}\cacc & \color{gray}\sacc 
& \cacc & \sacc 
& \cacc & \sacc 
& \cacc & \sacc 
& \cacc & \sacc 
& \cacc & \sacc 
\\

\cmidrule(r){1-1}
\cmidrule(r){2-2}
\cmidrule(r){3-4}
\cmidrule(r){5-6}
\cmidrule(r){7-8}
\cmidrule(r){9-10}
\cmidrule(r){11-12}
\cmidrule(r){13-14}

\multirow{5}{*}{\ourcoco}
& Activity
&\color{gray}62.6&\color{gray}73.5
&51.3&53.2
&44.0& 52.1
&33.8& -
&35.9& -
&44.1&48.9
\\

& Location
&\color{gray}34.3&\color{gray}51.5
&58.5&54.0
&51.2& 52.9
&35.3& -
&37.4& -
&55.2&55.6
\\

& Mood
&\color{gray}22.4&\color{gray}43.3
&23.2&40.4
&15.9& 39.3
&20.9& -
&23.0& -
&38.1&32.6
\\

& Time of Day
&\color{gray}40.6&\color{gray}74.1
&62.8&65.2
&55.5& 64.1
&45.7& -
&47.8& -
&67.6&56.7
\\

& $Avg.$
&\color{gray}40.1&\color{gray}60.6
&48.9&\color{codegreen}\bf53.2
&41.6&52.1
&33.9& -
&36.0& - 
&\color{codegreen}\bf51.2&48.4
\\

\cmidrule(r){1-1}
\cmidrule(r){2-2}
\cmidrule(r){3-4}
\cmidrule(r){5-6}
\cmidrule(r){7-8}
\cmidrule(r){9-10}
\cmidrule(r){11-12}
\cmidrule(r){13-14}

\multirow{5}{*}{\ourfood}
& Food Type
&\color{gray}90.6&\color{gray}94.6
&36.0&52.6
&33.1& 46.5
&48.9& -
&52.4& -
&34.6&54.2
\\

& Cuisine
&\color{gray}54.9&\color{gray}81.4
&46.8&42.4
&43.9& 36.3
&31.7& -
&35.2& -
&47.0&65.9
\\

& Course
&\color{gray}63.5&\color{gray}84.7
&70.5&89.5
&67.6& 83.4
&48.6& -
&52.1& -
&69.1&85.7
\\

& Diet
&\color{gray}47.6&\color{gray}59.9
&48.5&62.1
&45.6& 56.0
&45.9& -
&49.4& -
&41.5&54.0
\\

& $Avg.$
&\color{gray}64.1&\color{gray}80.2
&\color{codegreen}\bf50.5&61.7
&47.5& 55.5
&43.8& -
&47.3& -
&48.1&\color{codegreen}\bf64.9
\\

\cmidrule(r){1-1}
\cmidrule(r){2-2}
\cmidrule(r){3-4}
\cmidrule(r){5-6}
\cmidrule(r){7-8}
\cmidrule(r){9-10}
\cmidrule(r){11-12}
\cmidrule(r){13-14}

\multirow{5}{*}{\clevr}
& Color
&\color{gray}77.7&\color{gray}94.0
&51.2&43.2
&47.8& 44.0
&75.3& -
&84.7& - 
&70.3&63.4
\\

& Texture
&\color{gray}34.1&\color{gray}41.9
&64.9&26.4
&61.5& 27.2
&56.5& -
&65.9& -
&65.3&42.1
\\
			
& Count
&\color{gray}43.7&\color{gray}81.5
&46.9&39.0
&43.5& 39.8
&53.9& -
&63.3& -
&65.7&73.3
\\

& Shape
&\color{gray}71.1&\color{gray}72.7
&70.0&38.7
&66.6& 39.5
&65.5& -
&74.9& -
&58.4&38.5
\\

& $Avg.$
&\color{gray}56.7&\color{gray}72.5
&58.3&36.8
&54.8&37.6
&62.8& -
&72.2& -
&\color{codegreen}\bf64.9&\color{codegreen}\bf54.3
\\

\cmidrule(r){1-1}
\cmidrule(r){2-2}
\cmidrule(r){3-4}
\cmidrule(r){5-6}
\cmidrule(r){7-8}
\cmidrule(r){9-10}
\cmidrule(r){11-12}
\cmidrule(r){13-14}

\multirow{4}{*}{\action}
& Action
&\color{gray}97.1&\color{gray}99.2
&86.4&58.7
&88.1& 55.3
&51.3& -
&55.0& -
&82.8&76.3
\\

& Location
&\color{gray}66.7&\color{gray}67.1
&82.0&52.9
&83.7& 49.5
&59.4& -
&63.1& -
&69.8&55.2
\\

& Mood
&\color{gray}75.5&\color{gray}80.7
&60.8&57.4
&62.5& 54.0
&71.0& -
&74.7& -
&52.3&50.2
\\

& $Avg.$
&\color{gray}79.8&\color{gray}82.3
&76.4&56.3
&\color{codegreen}\bf78.1 & 52.9
&60.6& -
&64.3& -
&68.3&\color{codegreen}\bf60.6
\\

\cmidrule(r){1-1}
\cmidrule(r){2-2}
\cmidrule(r){3-4}
\cmidrule(r){5-6}
\cmidrule(r){7-8}
\cmidrule(r){9-10}
\cmidrule(r){11-12}
\cmidrule(r){13-14}

\multirow{3}{*}{\card}
& Suit
&\color{gray}47.9&\color{gray}69.5
&54.9&65.6
&47.5& 60.7
&41.3& -
&44.0& -
&54.5&73.6
\\
		
& Rank
&\color{gray}35.0&\color{gray}64.2
&94.6&96.8
&87.2& 91.9
&32.6& -
&35.3& -
&92.1&95.1
\\

& $Avg.$
&\color{gray}41.4&\color{gray}66.9
&\color{codegreen}\bf74.8&81.2
&67.3&76.3 
&36.9&-
&39.6&- 
&73.3&\color{codegreen}\bf84.3
\\

\cmidrule(r){1-1}
\cmidrule(r){2-2}
\cmidrule(r){3-4}
\cmidrule(r){5-6}
\cmidrule(r){7-8}
\cmidrule(r){9-10}
\cmidrule(r){11-12}
\cmidrule(r){13-14}

\multirow{3}{*}{\fruit}
& Species
&\color{gray}84.0&\color{gray}93.1
&69.3&66.9
&68.1& 58.6
&58.8& -
&62.2& -
&76.9&70.7
\\
	
& Color
&\color{gray}54.8&\color{gray}83.5
&57.2&43.3
&56.0& 35.0
&43.3& -
&46.7& -
&53.3&51.5
\\
	
& $Avg.$
&\color{gray}69.4&\color{gray}88.3
&63.3&55.1
&62.0&46.8
&51.0& -
&54.4& -
&\color{codegreen}\bf65.1&\color{codegreen}\bf61.1
\\

\bottomrule
\end{tabular}
\end{table*}

%% file: tables/results_grouper/grouper_ablation.tex
\begin{table*}[!t]
\centering

\caption{
\textbf{Ablation study of multi-granularity refinement on the six \taskshort benchmarks.}
We compare three ways of constructing cluster names: Initial Names (IN), Flat Refinement (FR), Multi-granularity Refinement (MR). We report \cacctitle (\cacc) and \sacctitle (\sacc)as percentages (\%). Average (\textit{Avg.}) \cacc and \sacc across different criteria on each dataset is also provided. These expanded results correspond to the plotting shown in \reffig{study_granularity}.
}
\lbltab{grouper_ablation}

\begin{tabular}{l l@{\ \ \ \ }r@{\ \ \ \ }r@{\ \ \ \ }r@{\ \ \ \ }r@{\ \ \ \ }r@{\ \ \ \ }r}
\toprule

\multirow{2}{*}{Benchmark}
& \multirow{2}{*}{Criterion}
& \multicolumn{2}{c}{IN}
& \multicolumn{2}{c}{FR}
& \multicolumn{2}{c}{MR}
\\

&
& \cacc & \sacc 
& \cacc & \sacc 
& \cacc & \sacc 
\\

\toprule

\multirow{5}{*}{\ourcoco}
			
& Activity
&14.1&48.5
&34.5&40.5
&44.1&48.9
\\

& Location
&30.0&51.9
&41.4&56.0
&55.2&55.6
\\

& Mood
&6.6&34.7
&21.9&32.1
&38.1&32.6
\\

& Time of Day
&24.4&50.5
&28.2&54.4
&67.6&56.7
\\

& $Avg.$
&18.8&46.4
&31.5&45.8
&\color{codegreen}\bf51.2&\color{codegreen}\bf48.4
\\

\hline

\multirow{5}{*}{\ourfood}
			
& Food Type
&33.9&52.4
&35.5&54.3
&34.6&54.2
\\

& Cuisine
&30.6&39.7
&27.6&36.5
&47.0&65.9
\\

& Course
&52.9&81.1
&62.8&83.0
&69.1&85.7
\\

& Diet
&14.0&46.6	
&36.8&58.2	
&41.5&54.0
\\

& $Avg.$
&32.9&55.0
&40.7&58.0
&\color{codegreen}\bf48.1&\color{codegreen}\bf64.9
\\

\hline

\multirow{5}{*}{\clevr}	
& Color
&56.5& 49.7
&60.9& 53.0
&70.3&63.4
\\

& Texture
&56.5& 26.0
&60.9& 33.0
&65.3&42.1
\\

& Count
&56.5& 39.6
&56.5& 40.8
&65.7&73.3
\\

& Shape
&47.8& 33.6
&47.8& 41.8
&58.4&38.5
\\

& $Avg.$
&54.3& 37.2
&56.5& 42.2
&\color{codegreen}\bf64.9&\color{codegreen}\bf54.3
\\

\hline

\multirow{4}{*}{\action}

& Action
&72.2& 63.6
&90.5& 63.0
&82.8&76.3
\\

& Location
&46.0& 50.4
&65.9& 59.3
&69.8&55.2
\\

& Mood
&20.6& 41.9	
&46.0& 51.0
&52.3&50.2
\\

& $Avg.$
&46.3& 52.0
&67.5& 57.8
&\color{codegreen}\bf68.3&\color{codegreen}\bf60.6
\\

\hline

\multirow{3}{*}{\card}

& Suit
&40.9& 50.1
&45.7& 45.7
&54.5&73.6
\\

& Rank
&43.0& 55.1
&47.7& 54.6
&92.1&95.1
\\

& $Avg.$
&42.0& 52.6
&46.7& 50.2
&\color{codegreen}\bf73.3&\color{codegreen}\bf84.3
\\

\hline

\multirow{3}{*}{\fruit}

& Species
&59.2& 68.6
&64.1& 67.0
&76.9&70.7
\\

& Color
&41.8& 56.7
&44.7& 42.3
&53.3&51.5
\\

& $Avg.$
&50.5& 62.7
&54.4& 54.7
&\color{codegreen}\bf65.1&\color{codegreen}\bf61.1
\\

\bottomrule
\end{tabular}
\end{table*}

%% file: tables/comp_finer.tex
\begin{table}[!ht]
    \centering
    \caption{
    \textbf{Study of substructure discovery for fine-grained criteria.}
    We report \cacclong{} (\cacc) and \sacclong{} (\sacc) as percentages (\%). The pseudo upper-bound (UB) performance is obtained using CLIP~\citep{radford2021learning} ViT-L/14 in a zero-shot transfer setting with the ground-truth class names. \dag: We compare with FineR~\citep{liu2024democratizing} without its post-class name refinement step to ensure a fair comparison.
    }

    \lbltab{comp_finer}
    \tablestyle{1.5pt}{1.5}

    \begin{tabular}{@{}l@{\ \ }c@{\ }c@{\ \ }c@{\ }c@{\ \ }}
    \toprule
    & \multicolumn{2}{c}{CUB200} & \multicolumn{2}{c}{Car196}\\
    & \cacc & \sacc & \cacc & \sacc\\
    
    \cmidrule(r){1-1}
    \cmidrule(r){2-3}
    \cmidrule(r){4-5}

    \color{gray} UB &
    \color{gray} 57.4 & \color{gray} 80.5 & \color{gray} 63.1 & \color{gray} 66.3
    \\
        
    \cmidrule(r){1-1}
    \cmidrule(r){2-3}
    \cmidrule(r){4-5}

    FineR\dag &
    44.8 & 64.5 & \colorbox{firstBest}{\bf 33.8} & \colorbox{firstBest}{\bf 52.9}
    \\
    
    \cmidrule(r){1-1}
    \cmidrule(r){2-3}
    \cmidrule(r){4-5}
    
    Ours &
    30.1 & 56.7 & 21.3 & 35.9
    \\
    
    Ours + FineR &
    \colorbox{firstBest}{\bf 45.1} & \colorbox{firstBest}{\bf 68.9} & 31.1 & 47.3
    \\
\bottomrule
\end{tabular}
\end{table}

%% file: tables/app_celebA.tex
\begin{table}[!h]
    \centering
    \lesspace
    \caption{
    \textbf{Debiasing Results and Comparison on CelebA.} We use the groups discovered by \methodshort to train DRO and compare it with state-of-the-art debiasing methods. Additionally, we present DRO results using the ground-truth distribution (DRO+GT) for reference.
    }
    \lbltab{app_celebA_mitigation}

    \begin{tabular}{lcc}
    \toprule
    Method
    & Worst
    & Avg.
    \\

    \hline
    
    JTT
    & 81.5
    & 88.1
    \\

    CNC
    & 88.8
    & 89.9
    \\

    DRO+B2T
    & 90.4
    & {\bf93.2}
    \\
    DRO+\methodshort
    & {\bf90.9}
    & 93.1
    
    \\
    
    \hline
    {\bf\color{annoGrey}DRO+GT} 
    & {\color{annoGrey} 89.7}
    & {\color{annoGrey} 93.6}
    \\
    
\bottomrule
\lesspace
\end{tabular}
\end{table}